\PassOptionsToPackage{pdftex}{graphicx}
\documentclass[a4paper,11pt]{article}

\usepackage{amsmath,amsfonts,bm}

\usepackage{amsmath,amssymb,amsthm,enumerate,enumitem,hyperref,algorithm,algorithmic,tikz,cancel,xcolor,array,tabularx,natbib,bbm,subcaption,cleveref,overpic,graphicx,ctable,cases,diagbox,pifont,mdframed,mathdots,wrapfig,changepage,multirow,colortbl,float,pgf,etoolbox}
\usepackage[labelfont=bf,tableposition=top]{caption}
\usepackage{mathtools}
\usepackage{esint}
\usepackage{overpic}

\definecolor{high}{HTML}{FF0000}
\definecolor{low}{HTML}{00FF00}
\newcommand*{\opacity}{50}

\definecolor{amber}{HTML}{FFBF00}

\newcommand*{\minvalscale}{1.0000}
\newcommand*{\maxvalscale}{3.2009}

\newcommand*{\minvaloffset}{1.0000}
\newcommand*{\maxvaloffset}{3.6152}

\newcommand{\gradientscale}[1]{

    \ifdimcomp{#1pt}{>}{\maxvalscale pt}{#1}{

        \ifdimcomp{#1pt}{<}{\minvalscale pt}{#1}{

            \pgfmathparse{int(round(100*(#1/(\maxvalscale-\minvalscale))-(\minvalscale*(100/(\maxvalscale-\minvalscale)))))}

            \xdef\tempa{\pgfmathresult}

            \cellcolor{high!\tempa!low!\opacity} #1

    }}

}

\newcommand{\gradientoffset}[1]{

    \ifdimcomp{#1pt}{>}{\maxvaloffset pt}{#1}{

        \ifdimcomp{#1pt}{<}{\minvaloffset pt}{#1}{

            \pgfmathparse{int(round(100*(#1/(\maxvaloffset-\minvaloffset))-(\minvaloffset*(100/(\maxvaloffset-\minvaloffset)))))}

            \xdef\tempa{\pgfmathresult}

            \cellcolor{high!\tempa!low!\opacity} #1

    }}

}

\usepackage{hyperref}
\usepackage{xcolor}
\hypersetup{
    colorlinks,
    linkcolor={red!50!black},
    citecolor={magenta},
    urlcolor={blue!90!black}
}

\usepackage{amsmath,amsfonts,bm}

\usepackage{graphicx}
\usepackage{booktabs}
\usepackage{tabularx}
\usepackage{array}
\usepackage{tikz}
\usepackage{adjustbox}
\usepackage{xcolor}
\usepackage{caption}
\usepackage{makecell}
\usepackage{calc}

\usepackage[most]{tcolorbox} 

\definecolor{amazonorange}{HTML}{FF9900}
\definecolor{darkorange}{HTML}{AA5500}

\tcbset{
  highlightbox/.style={
    colback=amazonorange!20,    
    colframe=darkorange!100,  
    boxrule=0.4pt,     
    arc=2pt,           
    outer arc=2pt,
    left=4pt, right=4pt, top=2pt, bottom=2pt, 
    fonttitle=\bfseries,
    coltitle=black,
    sharp corners,
    enhanced
  }
}

\theoremstyle{definition}
\newtheorem{thm}{Theorem}

\def\eqref#1{equation~\ref{#1}}

\def\1{\bm{1}}

\DeclareMathAlphabet{\mathsfit}{\encodingdefault}{\sfdefault}{m}{sl}
\SetMathAlphabet{\mathsfit}{bold}{\encodingdefault}{\sfdefault}{bx}{n}

\newcommand{\R}{\mathbb{R}}

\usepackage{hyperref}
\usepackage{url}

\usepackage{amsmath,amsfonts,amssymb,amsthm}

\usepackage[margin=1in]{geometry}

\usepackage{parskip}

\usepackage{hyperref}

\usepackage{xcolor}
\hypersetup{
    colorlinks,
    linkcolor={red!50!black},
    citecolor={blue!50!black},
    urlcolor={blue!80!black}
}

\usepackage{url}
\usepackage{authblk}
\usepackage{bbm}
\usepackage{comment}
\usepackage{accents}
\usepackage{adjustbox}
\usepackage{booktabs}
\usepackage{caption}
\usepackage{tikz}
\usepackage[pdftex]{graphicx}

\newif\ifvspaces
\vspacesfalse
\newcommand{\myvspace}[1]{\ifvspaces\vspace{#1}\fi}

\title{Understanding the Implicit Biases of Design Choices for \\ Time Series Foundation Models}

\author{
		Annan Yu,$^{1,}$\thanks{Work done during an internship at AWS.} \hspace{+0.3cm} 
 Danielle C. Maddix,$^{2,}$\thanks{Correspondence to: Danielle C. Maddix $<$\url{dmmaddix@amazon.com}$>$.} \hspace{+0.3cm} Boran Han,$^{2}$ \hspace{+0.3cm} Xiyuan Zhang,$^{2}$ \\
 
  \vspace{-0.1cm}
 
 Abdul Fatir Ansari,$^{2}$ \hspace{+0.3cm} Oleksandr Shchur,$^{2}$ \hspace{+0.3cm} Christos Faloutsos,$^{3}$ \\
 
 \vspace{0.25cm}
 
 Andrew Gordon Wilson,$^{4}$ \hspace{+0.3cm} Michael W. Mahoney,$^{4}$ \hspace{+0.3cm} Yuyang Wang$^{2}$ \\
	
	\vspace{0.6cm}
	
	$^1$ Center for Applied Mathematics, Cornell University, Ithaca, NY 14853, USA \\
	$^2$ Amazon Web Services (3075 Olcott St., Santa Clara, CA 95054, USA) \\
    $^3$ Amazon Selling Partner Services (501 Fairview Ave. N., Seattle, WA 98109, USA) \\
    $^4$ Amazon Supply Chain Optimization Technologies (7 West 34th St., New York, NY 10001, USA)
}

\date{\today}

\begin{document}

\maketitle

\begin{abstract}
Time series foundation models (TSFMs) are a class of potentially powerful, general-purpose tools for time series forecasting and related temporal tasks, but their behavior is strongly shaped by subtle inductive biases in their design. 
Rather than developing a new model and claiming that it is better than existing TSFMs, e.g., by winning on existing well-established benchmarks, our objective is to understand how the various ``knobs'' of the training process affect model quality. 
Using a mix of theory and controlled empirical evaluation, we identify several design choices (patch size, embedding choice, training objective, etc.) and show how they lead to implicit biases in fundamental model properties (temporal behavior, geometric structure, how aggressively or not the model regresses to the mean, etc.); and we show how these biases can be intuitive or very counterintuitive, depending on properties of the model and data. 
We also illustrate in a case study on outlier handling how multiple biases can interact in complex ways; and we discuss implications of our results for learning the bitter lesson and building TSFMs.
\end{abstract}

\section{Introduction}
\label{sec:intro}

Design decisions associated with modern machine learning models typically involve multiple trial-and-error iterations using one or more of many possible ``knobs,'' including model architectures, loss functions, and parameters and hyperparameters such as learning rate, batch size, patch size, and embedding strategy.
A common workflow is to adjust these knobs based on intuitions about the model and then use available data, typically in the form of curated, well-established benchmarks, to train and evaluate the model. 
For example, in the case of time series, intuitions include that nearby data points are similar and should be mapped close to each other, that time is continuous and that this continuity should be maintained within the model, and that regression algorithms should regress to the mean. 
This process, which is expensive (since one typically has only weak control on the precise effect of varying knobs, since various knobs interact in complex ways, and since one often does not even know what knobs are most relevant for improving a given downstream task) is iterated upon until one obtains a model that is \emph{quantitatively} ``better'' on well-established benchmarks.

Unfortunately, model improvements often do not stand the test of time---when the data and compute are \emph{much} larger, or when the data are \emph{qualitatively} different than the well-established benchmarks on which the model was trained, the same knob fiddling can have a different, and often deleterious, effect.
This leads to worse results and more model development effort in the longer term.
Understanding how various design choices affect downstream model properties is of increasing importance as the community aims to design foundation-like models that are forward-compatible with qualitatively  more and different data.

Here, we consider these issues in the context of recently-developed time series foundation models (TSFMs).
Our goal is not to develop ``yet another'' model.
Rather, we aim to understand the effects, i.e., the implicit biases, encoded by these various design decisions. 
We are particularly interested in how those effects vary with the properties of the data and the model, in particular for data that are similar to well-established benchmarks versus data that are qualitatively different. 
In time series forecasting, there are multiple well-known public benchmarks, e.g., M4~\citep{makridakis2020m4}, ETT~\citep{zhou2021informer}, and the large collection from ~\citet{godahewa2021} including Kaggle's Tourism~\citep{tourism_kaggle}, Electricity~\citep{trindade2015electricity}, Traffic~\citep{traffic}, and Dominick~\citep{dominick}. There is also a recent explosion of TSFMs, e.g., LLMTime~\citep{gruver2023large}, Chronos~\citep{ansari2024chronos}, Moirai~\citep{woo2024unified}, TimesFM~\citep{das2024decoder}, Chronos-Bolt~\citep{ChronosBolt}, WaveToken~\citep{masserano2024enhancing}, and TOTEM~\citep{talukder2024totem}.  
If a foundation model is to be easily-adapted to various tasks, with minimal additional task-specific data~\citep{liang2024foundation}, this leads to the important open question: how does one know the biases of various knobs in a TSFM on data that are qualitatively different than the benchmarks on which the TSFM was trained and evaluated?

\begin{figure}
    \centering
    \includegraphics[width=0.98\linewidth]{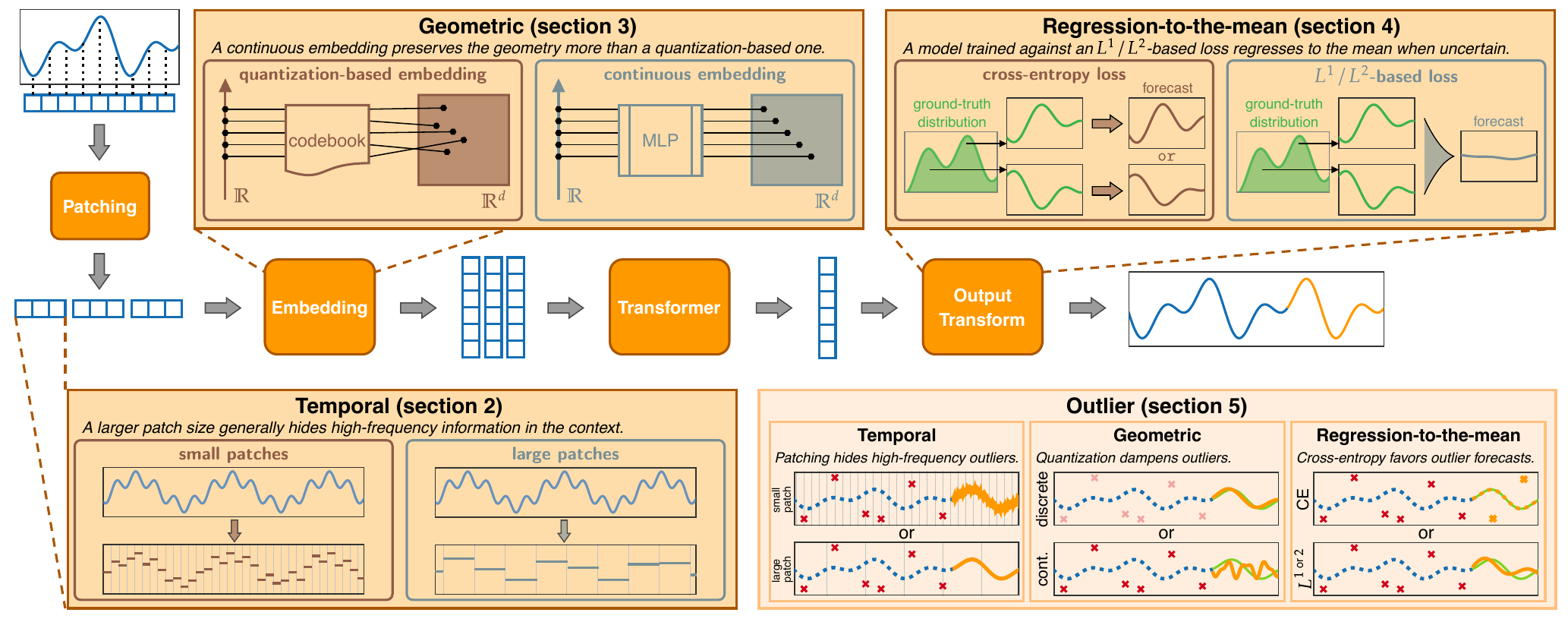}
    
    \caption{An illustration of the three broad classes of biases that we identify and investigate. The \emph{temporal bias} tends to be induced by patching, which transplants the temporal dependencies from the input sequence to a Transformer's hidden space; the \emph{geometric bias} tends to be related to the input embedding, where a discrete versus a continuous embedding preserves the input space's geometry differently; the \emph{regression-to-the-mean bias} tends to be caused by the training loss, where a continuous loss tends to average different outcomes more than a cross-entropy loss does. We also illustrate the interaction of these biases in a case study of outlier handling.}
    \label{fig:biases}
    
\end{figure}

In this paper, we identify three classes of design choices --- the ``knobs'' of patch size, whether the embedding is discrete or continuous, and the choice of loss; and, based on extensive analysis, we characterize three corresponding classes of implicit biases --- which we call the ``temporal bias,'' the ``geometric bias,'' and the ``regression-to-the-mean bias.''
We describe in detail how each design choice tends to affect the corresponding bias; and we show how, depending on the model, this effect could be intuitive or very counterintuitive.
We also provide an example of how the design choices can interact in complex ways.
Our main results, summarized in~\Cref{fig:biases}, are as~follows.

\begin{itemize}[leftmargin=*,noitemsep,nolistsep]
\item  
    \textbf{Patch Size $\rightarrow$ Temporal Bias.} 
    Given a discrete sequence 
    sampled from a time series, 
    many TSFMs group these inputs into patches of length $k$, 
    with $k=1$ as a ``trivial'' corner case,
    and map each patch into a Transformer's hidden space. 
    This design choice of the patch size parameter leads to a \textit{temporal bias}: 
    depending on its value, a TSFM may favor low-frequency information more or less aggressively.
    A larger patch size, e.g., 16 in Chronos-Bolt, favors low-frequency components in the time series, which can be challenging on chaotic (or even non-chaotic) systems that consist of high frequency and/or mixed-frequency components.
    Conversely, a smaller patch size, e.g., 1 in Chronos or 2 or 4 or 8 in Chronos-Bolt, can capture parts of these high frequency components.  
    Importantly, many time series are periodic or quasi-periodic,%
    \footnote{Periodic phenomena repeat at \emph{precise} intervals, while quasi-periodic phenomena follow complex, non-repeating patterns of \emph{roughly} regular intervals.  Examples of the former include New Year's Day and someone's birthday, which (up to leap year) occur every 365 days exactly; while examples of the latter include Mother's Day, which wanders around the calendar, and modes that hop in nonlinear chaotic dynamical systems.}
    and capturing these two related but distinct phenomena can depend on temporal bias in subtle ways.
    (See~\cref{sec:frequency}.)
    \item
    \textbf{Embedding Strategy $\rightarrow$ Geometric Bias.} 
    In TSFMs, there are two common embedding strategies: discrete; and continuous.
    A discrete embedding quantizes a time series into a finite number of regions, where each is treated as a ``word token,'' as in an LLM. 
    This is used in Chronos. 
    A continuous embedding defines a trainable continuous function, usually an MLP, that maps the time series 
    into a Transformer's hidden space.
    This is used in, e.g., Chronos-Bolt, Moirai, TimesFM and Time-MoE.
    This design choice leads to a \textit{geometric bias}: a continuous embedding naturally tends to preserve geometric relationships; while a quantization-based embedding introduces a continuous-to-discrete  ``unrounding noise,'' and then relies on training to (imperfectly) recover continuous information in the hidden space.  
    As a result, quantization-based embeddings introduce several qualitative distinctions: they emphasize local information over global context; they treat fine- and large-scale patterns with less discrimination; and they remain invariant to signals that are shifted by an offset.
    (See~\cref{sec:geometry}.)
    \item 
    \textbf{Training Loss $\rightarrow$ Regression-to-the-mean Bias.} 
    Discrete and continuous embeddings are two ways to handle the input domain; similarly, the output space can also be treated in a discrete or a continuous way. 
    A cross-entropy loss, e.g., in Chronos, computes forecasts as a discrete classification task, and is more aligned with LLMs. 
    A continuous regression-based loss is more aligned with classical statistical methods: the $L^2$-based mean squared loss is a popular choice for models that perform point forecasting, e.g., TimesFM, whereas models that predict quantiles are usually trained on an $L^1$-based quantile loss or negative log-likelihood loss, e.g., Chronos-Bolt.   
    This design choice leads to a \textit{regression-to-the-mean bias}: TSFMs that use a regression-based loss tend to regress to either the mean or median forecast, while TSFMs that use a cross-entropy loss tend to regress much less aggressively and thus, for instance, have the flexibility to hop between basins for chaotic data and to better learn quasi-periodic information. 
    (See~\cref{sec:regression}.)
\end{itemize}

In each case, we provide theory and/or quantification through controlled empirical evaluations to corroborate the existence of each bias and to demonstrate how each bias follows from the corresponding design choice.  
To demonstrate this, we mostly compare Chronos versus Chronos-Bolt because they are natural to compare:
they use the same T5 Transformer backbone, they have comparable sizes, and they are pretrained on similar datasets (while most TSFMs are trained on different datasets). 
We also show in ablations that other TSFMs relying on the same choices of knobs behave (qualitatively) similarly to Chronos-Bolt. 
Lastly, as an illustration of how to use our insights, we demonstrate how these various biases can interact and mix with each other through an outlier handling analysis.
(See~\cref{sxn:application}.) 
An important summary conclusion of our investigation is that
\emph{design choices --- e.g., a large patch size, a continuous embedding, and a continuous training loss --- which may induce intuitive implicit biases that work well on well-established time series benchmarks, may actually be deleterious to qualitatively different datasets such as chaotic systems, and for these datasets the opposite design choices may be beneficial}. 
(See~\Cref{app:goodandbad}.)
See~\Cref{app:relatedwork} for related work.

\noindent\textbf{Motivation: deep dive on ``improving'' Chronos with Chronos-Bolt.}
This work was motivated by recent developments and evaluations on Chronos and other TSFMs.
From the perspective of ``knobs,'' Chronos is peculiar relative to other TSFMs in several ways: it tokenizes time series values using scaling and quantization into a fixed vocabulary, and then it trains existing transformer-based language model architectures on these tokenized time series via a cross-entropy loss, rather than using a continuous embedding and regression-based loss.
To ``improve'' the original Chronos model, Chronos-Bolt was developed.
Chronos-Bolt replaced quantization with a continuous embedding to preserve distances better, and the discrete cross-entropy loss with the $L^1$-based quantile loss, which has been commonly used in probabilistic forecasting methods, e.g., MQ-(R)CNN/Transformer~\citep{wen2018multihorizonquantilerecurrentforecaster, pmlr-v151-park22a, eisenach2022mqtransformermultihorizonforecastscontext}.
These changes were motivated by intuitive considerations in developing forecasting methods, as outlined above, and they were ablated and validated on well-established benchmarks.
Thus, it came as a surprise when~\citet{zhang2024zero,zhang2025context} showed that Chronos, while not the top performer (relative to other TSFMs) on many standard benchmarks, achieves \emph{higher} accuracy and produces quantitatively \emph{better} forecasts than other TSFMs, including Chronos-Bolt, on the qualitatively-different time series data from nonlinear chaotic systems.

\noindent\textbf{Implications more generally: for the bitter lesson.}
Understanding how design choices affect downstream model performance requires one to ask:
What does it mean for a model to be ``better''?
If one fixes model size and varies the amount of data, or vice versa, then one might hope to observe a sweet spot as some parameter is changed, or one might hope to observe continual if diminishing improvement.
If one varies model and data together, then one might hope to observe some sort of neural scaling, where improvement continues across scales of model and data and compute, without saturating.
For a fixed amount of compute and data, the former is important.
For exponentially increasing amount of compute and data, the latter is important.
It's not obvious (and likely not true) that the best models designed for one regime will be the same as the best models designed for the other regime --- or that the implicit biases for models appropriate for one regime will be the same as the implicit biases for models appropriate for the other regime (see~\Cref{app:goodandbad}). 
Related to this is that it is not clear how the various knobs of the model development process will be affected by qualitatively more and different data and compute that will be available in the future.
Our work highlights challenges with learning this ``bitter lesson'': 
\emph{design principles for developing machine learning systems that are forward-compatible with absorbing qualitatively more/different data and compute in a graceful way are not as well understood as the design principles for developing models that work with data and compute of a given size and scale}.
We expect that our work to shed light on this.

\section{Temporal Bias: How Do TSFMs Learn Time?}
\label{sec:frequency}

In this section, we analyze the effects of various design choices on the temporal properties of a TSFM, which lead to a \textit{temporal} bias.
Time series data are fundamentally different from discrete language data, in that they are often thought of as evolving over a continuous time domain.  
As a result, notions such as frequency, periodicity, and seasonality play a central role, and these notions often determine the dominant structures that a model must capture. 
Here, we distinguish between two related temporal notions (frequency and periodicity), and we describe how they are affected by different design choices.
We outline key considerations in this section, and leave additional details to \Cref{app:temporal}.

\textbf{FREQUENCY.} 
The term \emph{frequency bias} was observed and studied in general neural networks~\citep{rahaman2019spectral,yang2019fine,xu2020frequency}, and it was later extended to sequential models such as Transformers~\citep{piao2024fredformer} and state-space models (SSMs)~\citep{yu2024tuning}.  
In general, these models show different inductive biases when learning signals of different frequencies. 
One might hypothesize that we generally want to learn low-frequency modes better, because they usually contain less noise and/or describe more dominant ``first-order'' structure.
However, in many tasks, e.g., forecasting chaotic systems or systems with quasi-periodic properties, high-frequency modes contain essential information that steers the trajectories, and their role cannot be downplayed. 
It is important to understand the cause of the frequency bias and to be able to tune it for the nature of the specific problem.

A TSFM operates on a discrete sample of a continuous time domain, in which a time series is defined.
Given a continuous time series that can be sampled, we can hope to control this bias by changing the sampling interval $\Delta t$.  
More commonly, however, time series are recorded at fixed frequency intervals, and the data is a sampled sequence with a given frequency, and we have no control over $\Delta t$.
When we talk about the ``frequency,'' we imagine the Fourier modes in a fixed sequence $\mathbf{x} = (x_1, \ldots, x_L)$, indexed by natural numbers from $1$ to $L$. 
Then, we ask: what determines the frequency content that a TSFM will learn? 
Perhaps surprisingly, a frequency bias is induced as soon as $\mathbf{x}$ is embedded into the hidden space. The reason is that an embedding function $\boldsymbol{\phi}: \R^k \rightarrow \R^d$ maps a patch of size $k$, i.e., $(x_i, \ldots, x_{i+k-1})$, into the hidden space.
The following result, which is proven in \Cref{app:proof_of_freq_bias_thm}, makes precise a sense in which patches of different frequencies are embedded differently.

\begin{thm}\label{thm.frequencybias}
    Let $k$ be a patch size, $1 < n < k$ be any positive integer, and $\omega \leq \min(k/n, n)-1$ be a bandwidth. Let $B_1, B_2, \ldots, B_n \subset [k] := \{1, \ldots, k\}$ be $n$ mutually disjoint sets whose bandwidths are below $\omega$, i.e., $|B_i| \leq \omega$ for every $1 \leq i \leq n$. Let unit vectors $\mathbf{v}_1, \ldots, \mathbf{v}_n \in \R^k$ be $n$ patches that belong to the same band, i.e., $\text{supp}(\hat{\mathbf{v}}_i) \subset B_j$ for all $1 \leq i \leq n$ and a common $j$. Let unit vectors $\mathbf{u}_1, \ldots, \mathbf{u}_n \in \R^k$ be $n$ patches that belong to all different bands, i.e., $\text{supp}(\hat{\mathbf{u}}_i) \subset B_i$ for all $1 \leq i \leq n$. Let $\mathbf{V} = \begin{bmatrix}
        \mathbf{v}_1 & \cdots & \mathbf{v}_n
    \end{bmatrix} \in \R^{k \times n}$ and $\mathbf{U} = \begin{bmatrix}
        \mathbf{u}_1 & \cdots & \mathbf{u}_n
    \end{bmatrix} \in \R^{k \times n}$, and let $\boldsymbol{\phi}(\mathbf{x}) = \mathbf{W}_2\, \text{ReLU}(\mathbf{W}_1 \mathbf{x}) + \mathbf{b}$ be a neural network embedding, with $\mathbf{W}_1 \in \R^{m \times k}$, $\mathbf{W}_2 \in \R^{d \times m}$, and $\mathbf{b} \in \R^d$. The following two statements hold:
    \myvspace{-0.5\baselineskip}
    \begin{enumerate}[leftmargin=*]
        \item (Patches of similar frequencies are embedded in a shared subspace.) Let $0 < \varepsilon \leq 1$ be given. Given any neural network width $m$ and any $\mathbf{W}_1$ and $\mathbf{W}_2$ such that $\|\boldsymbol{\phi}(\mathbf{V})\|_2 = \Omega(1) \|\mathbf{W}_2\|_2 \|\mathbf{W}_1\|_2 \|\mathbf{V}\|_2$ and any bias term $\mathbf{b}$ such that $\|\mathbf{b} \,\boldsymbol{1}_{n}^\top\|_F = \mathcal{O}(1) \|\mathbf{W}_2\, \text{ReLU}(\mathbf{W}_1 \mathbf{V})\|_F$, we have
        \begin{align*}
            \text{rank}_\varepsilon(\boldsymbol{\phi}(\mathbf{V})) &:= \left|\left\{j \;\middle|\; \sigma_j(\boldsymbol{\phi}(\mathbf{V})) > \varepsilon \sigma_1(\boldsymbol{\phi}(\mathbf{V})) \right\}\right| = \mathcal{O}(1) \varepsilon^{-2} \omega,\\
            \text{stab-rank}(\boldsymbol{\phi}(\mathbf{V})) &:= \|\boldsymbol{\phi}(\mathbf{V})\|_F / \|\boldsymbol{\phi}(\mathbf{V})\|_2 = \mathcal{O}(\omega).
        \end{align*}
        
        \myvspace{-0.5\baselineskip}
        
        \item (Patches of different frequencies are embedded in nearly orthogonal subspaces.) There exists a universal constant $\varepsilon > 0$ so that the following statement holds. Assume that $\mathbf{W}_1$ is a random Gaussian matrix whose entries follow i.i.d.\ $\mathcal{N}(0,\alpha)$ distribution and $\mathbf{W}_2$ is a random Gaussian matrix whose entries follow i.i.d.\ $\mathcal{N}(0,\beta)$ distribution, where $\alpha$ and $\beta$ are any positive numbers. Set the bias term to be $\mathbf{b} = -\mathbf{W}_2 \boldsymbol{1}_{m} / \sqrt{2\pi}$. Then, for any $0 < \delta < 1$, $m = \Omega(\log(1/\delta))$, $d = \Omega(\log(n/\delta))$, $m > n$, and $m / d = \Theta(1)$, with probability no less than $1 - \delta$, we have
        \[
            \text{rank}_\varepsilon(\boldsymbol{\phi}(\mathbf{U})) = \Omega(n), \qquad \text{stab-rank}(\boldsymbol{\phi}(\mathbf{U})) = \Omega(n).
        \]
    \end{enumerate}
    \myvspace{-0.5\baselineskip}
    Moreover, the constants in all $\mathcal{O}$ and $\Omega$-notations are universal.
\end{thm}

\begin{figure}[h]
    \centering
    \myvspace{-0.5\baselineskip}
    \includegraphics[width=1\linewidth, trim=0cm 0.2cm 0cm 0cm, clip]{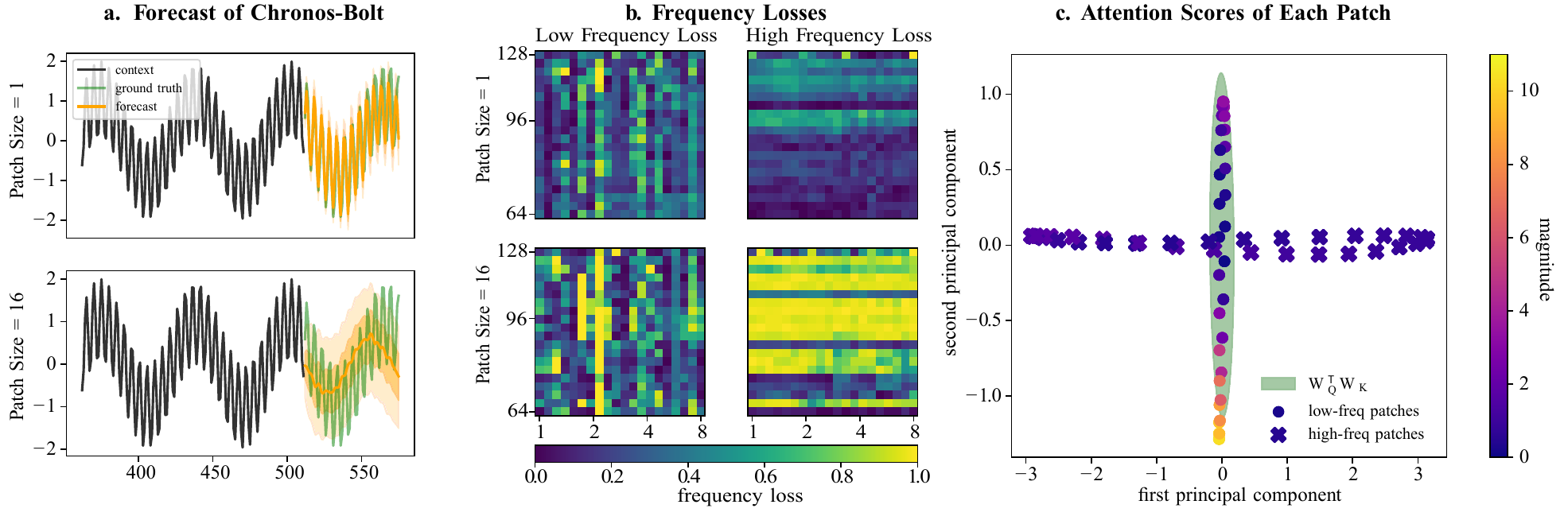}
    \caption{In (a), we show the forecasts of a signal consisting of a low-frequency and a high-frequency mode using two pretrained Chronos-Bolt models with patch size $k = 1$ and $k = 16$, respectively. In (b), we vary the frequencies of the two modes, and we evaluate the frequency loss in predicting each mode. We see that Chronos-Bolt with a patch size of $16$ fails significantly at capturing the high-frequency information.
    In (c), we project the embeddings of low-frequency patches and high-frequency patches onto the two principal components, and we show how the attention matrices align with these two principal components on average. We observe that the attention scores are heavily dominated by the low-frequency patches.
    }
    \label{fig:frequency}
    \myvspace{-0\baselineskip}
\end{figure}

The $\varepsilon$-rank and stable rank of a matrix $\boldsymbol{\phi}(\mathbf{V})$ or $\boldsymbol{\phi}(\mathbf{U})$ measure the ``dimension'' of its column space.  
In essence,~\Cref{thm.frequencybias} says that patches with similar frequencies are embedded into a low-dimensional subspace of $\mathbb{R}^d$, with the ``effective'' dimension bounded proportionally to the bandwidth~$\omega$ (see also~\citet{yu2025transformer}).  
At the same time, low-frequency and high-frequency patches are embedded in almost orthogonal subspaces. 
(See~\Cref{fig:frequency}c.) 
See~\Cref{fig:frequencybiasthm} in~\Cref{app:frequency} for a conceptual illustration of~\Cref{thm.frequencybias}.
One may wonder: what role does the patch size $k$ play in this story?  
If $k$ is small, then all patches share a similar bandlimit, because we must have $\omega \leq k$; as a result, all patches are embedded into a low-dimensional subspace of $\mathbb{R}^d$.

Now that we know that when $k$ is large, low-frequency and high-frequency patches are embedded very differently, the next question is: which frequencies does the model learn best?  
In practice, we find that most TSFMs with a large patch size~$k$ learn lower frequencies better than higher frequencies (see~\Cref{fig:frequency}a,b). 
This aligns with the conventional notion of ``frequency bias'' in other deep learning models~\citep{rahaman2019spectral,yang2019fine,yu2024tuning}. A full understanding of this phenomenon can be obtained by analyzing the training dynamics of Transformers, but even without going into that level of detail, we have an intuitive understanding: in many time series, the dominant first-order structure lies in the low-frequency components, while high-frequency content often appears more irregular or noisy.
As a result, the projection matrices $\mathbf{W}_Q$, $\mathbf{W}_K$, and $\mathbf{W}_V$ tend to align more closely with the subspace containing low-frequency information (see~\Cref{fig:frequency}c).

\begin{wrapfigure}{r}{0.6\textwidth}  
  \centering
    \includegraphics[width=\linewidth]{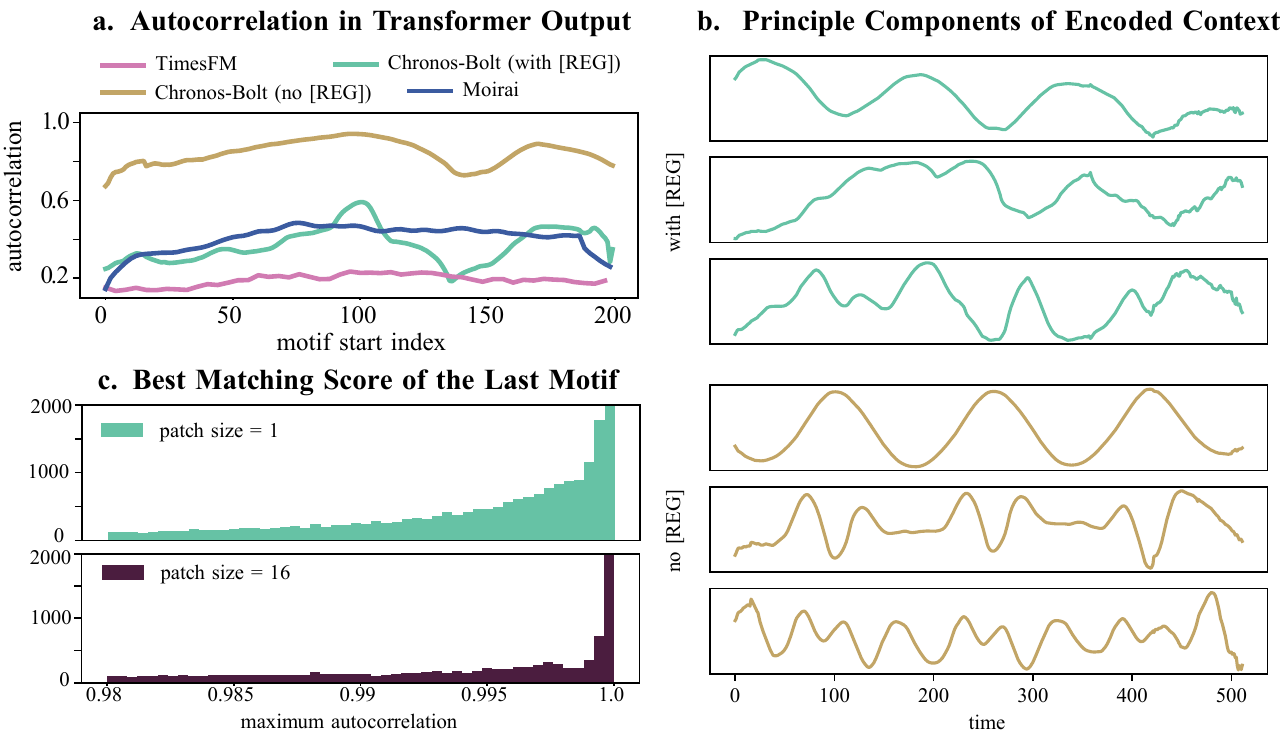}
    \caption{In (a), we show the autocorrelation in the output of a Transformer given a periodic input context, which measures the periodicity preservation of a model. In (b), we show leading principal components of the encoded context of a Chronos-Bolt model, with or without the \texttt{[REG]} token, given a periodic input. In (c) , we show the best matching score, aggregated on $10,000$ randomly sampled time series in Chronos' training dataset, with or without patching. The best matching score is computed by comparing the last motif of length $64$ to all previous motifs.}
    \label{fig:periodicity}
\end{wrapfigure}

\textbf{PERIODICITY.} 
Periodicity is a fundamental property of many time-series applications. 
By ``periodicity,'' we refer to loosely-recurring ``periods'' in a time series, not necessarily occurring \emph{precisely} at any fixed frequency. 
Maintaining strict periodic fidelity facilitates precisely-recurring pattern forecasting, but allowing more flexibility can be useful when time series are nonstationary or patterns drift. 
(See~\Cref{app:goodandbad} for cases where the periodicity bias is ``beneficial'' or ``deleterious.'') 
As with the frequency bias, patching is a knob that affects the periodicity bias. 
Given a sampled sequence from a time series with periodic patterns, if the sequence’s recurring motifs do not align with the chosen patch size, then the patching process can introduce aliasing effects. 
These effects distort the original frequency content, and this effectively changes the periodicity of the embedded sequence (see~\Cref{fig:periodicity}c).


In addition to patching, the \emph{periodicity bias} can be affected by other knobs. TSFMs that rely on the encoder-only, encoder--decoder, and decoder-only architectures differ in how well they preserve periodic structures present in the context. 
Encoders are generally more effective at preserving the periodicity since they process the entire input sequence bidirectionally to capture long-range periodic patterns (see~\Cref{fig:periodicity}a).

Surprisingly, we find that the use of a special \texttt{[REG]} token, which is commonly used in encoder--decoder models as a designated position for aggregating global information, can influence the periodicity bias (see~\Cref{fig:periodicity}b).  
The \texttt{[REG]} token is designed to serve as a fixed reference position for aggregating global information, and it can be useful for tasks such as regression or classification.  
Without masking, we observe empirically that the \texttt{[REG]} token can contaminate the preceding context, which disrupts many periodic patterns that otherwise would be preserved. 
(See~\Cref{app:periodicity} for a fine-grained analysis.)

\begin{tcolorbox}[highlightbox]
\textbf{Summary:} Frequency and periodicity biases are two connected views of how TSFMs represent time.  
\emph{Frequency bias} is controlled by the patch size $k$, which determines whether embeddings emphasize low- or high-frequency components.  
\emph{Periodicity bias} is controlled by the alignment of patch size $k$ with the underlying recurrent motifs, architectural choices, such as encoder versus decoder design, and the use of an unmasked \texttt{[REG]} token.
\end{tcolorbox}

\section{Geometric Bias: How Do TSFMs Learn Geometry?}
\label{sec:geometry}

In this section, we analyze the effects of various design choices on the geometric properties of the input embedding, which lead to a \textit{geometric} bias.  
There are two broad classes of embeddings: \emph{quantization-based} (featured, e.g., in Chronos) and \emph{continuous} (featured in most TSFMs, including Chronos-Bolt, TimesFM, and Moirai). 
A quantization-based embedding partitions $\mathbb{R}$ into regions $D_1, \ldots, D_V$ and defines $\boldsymbol{\phi}_Q: \mathbb{R} \rightarrow \mathbb{R}^d$ as a step function that is constant on each $D_i$.  
A continuous embedding instead parameterizes $\boldsymbol{\phi}_C: \mathbb{R}^k \rightarrow \mathbb{R}^d$ directly with a neural network. We use Chronos to represent the quantization-based embedding and Chronos-Bolt with a patch size of $k = 1$ to be representative of the continuous embedding.  
Note that we pretrain a new Chronos-Bolt model with $k = 1$ to eliminate the temporal bias from larger patch sizes, as discussed in \Cref{sec:frequency}, and to be more directly comparable with Chronos.

In a continuous embedding, the topology of $\mathbb{R}$ is inherently preserved by the continuity of $\boldsymbol{\phi}_C$; but in a quantization-based embedding, the ordering of the real line must be (re)learned during training.
Since each $\boldsymbol{\phi}_Q(D_i)$ is initially a random vector, training rarely recovers the full geometry of $\mathbb{R}$ in $\mathbb{R}^d$. At first glance, it may seem that it is generally good to preserve the geometry of the input space, as a continuous embedding does.
We show that while breaking the geometry does lead to certain problems, it is also surprisingly beneficial in some cases. 
Here, we focus on three related geometric notions (angles, distances, and norms), and we describe how they are affected by different design choices.
We outline key considerations in this section, and leave additional details to~\Cref{app:geomtric}.

\begin{figure}[!htb]
    \centering
    \myvspace{-0.5\baselineskip}
    \includegraphics[width=1\linewidth]{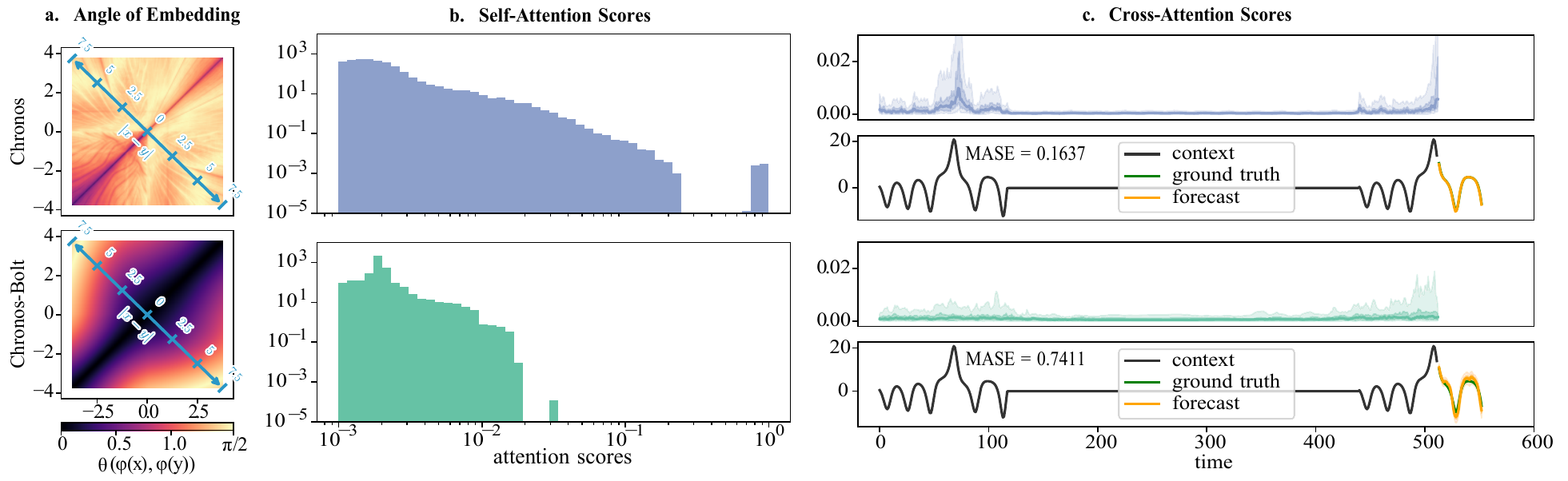}
    \vspace{-1.5\baselineskip}
    \caption{Panel (a) shows the angles between the embedded vectors $\boldsymbol{\phi}(x)$ and $\boldsymbol{\phi}(y)$ for Chronos' embedding $\boldsymbol{\phi}_Q$ and Chronos-Bolt's embedding $\boldsymbol{\phi}_C$, respectively. Panel (b) shows a histogram of all self-attention scores when inferring the two models on a corpus of contexts. Panel (c) shows the forecasts from both models on a context formed by repeating a motif of a chaotic system. Chronos achieves a significantly lower MASE than Chronos-Bolt. 
    The cross-attention scores are aggregated across all layers when generating the first prediction token. See~\Cref{app:geomtric} for details.} 
    \label{fig:locality}
\end{figure}

\textbf{ANGLES.} 
Our first measurement of the geometric bias is the angle between embedded vectors:
\[
    \theta_Q(x,y) = \arccos\left(\frac{|\boldsymbol{\phi}_Q(x) \cdot \boldsymbol{\phi}_Q(y)|}{\|\boldsymbol{\phi}_Q(x)\|_2 \|\boldsymbol{\phi}_Q(y)\|_2}\right), 
    \quad
    \theta_C(x,y) = \arccos\left(\frac{|\boldsymbol{\phi}_C(x) \cdot \boldsymbol{\phi}_C(y)|}{\|\boldsymbol{\phi}_C(x)\|_2 \|\boldsymbol{\phi}_C(y)\|_2}\right).
\]
In~\Cref{fig:locality}a, $\theta_Q(x,y)$ is significantly larger than $\theta_C(x,y)$ as a function of $|x-y|$, which makes an input appear more different from its close neighbors. 
This difference increases with the number of bins used in quantization.

The difference between the two types of embedding introduces an \emph{angular bias} in learning local information versus more global information.
This bias affects both self-attention and cross-attention in a Transformer.  
In~\Cref{fig:locality}b, the self-attention scores of Chronos are bimodal, i.e., tokens place high attention weights on their immediate neighbors, and low attention weights on distant tokens.  
In contrast, the attention scores of Chronos-Bolt are more evenly distributed, indicating a broader mixing of contextual information.  
A similar pattern appears in the right panels for cross-attention. 
Chronos focuses almost entirely on the most relevant part of the context, whereas Chronos-Bolt attends more broadly across the context.
Among other things, this makes Chronos better at ``parroting'' --- one reason why it performs better than other TSFMs on chaotic systems~\citep{zhang2025context}. This comes at the cost of less mixing across the context, which can hurt performance on tasks that require more complex reasoning (see~\Cref{app:goodandbad} for an example of this tradeoff).

\begin{figure}[!htb]
    \centering
    \myvspace{-0.5\baselineskip}
    \begin{overpic}[width=1\linewidth, trim=3cm 0cm 3cm 1cm, clip]{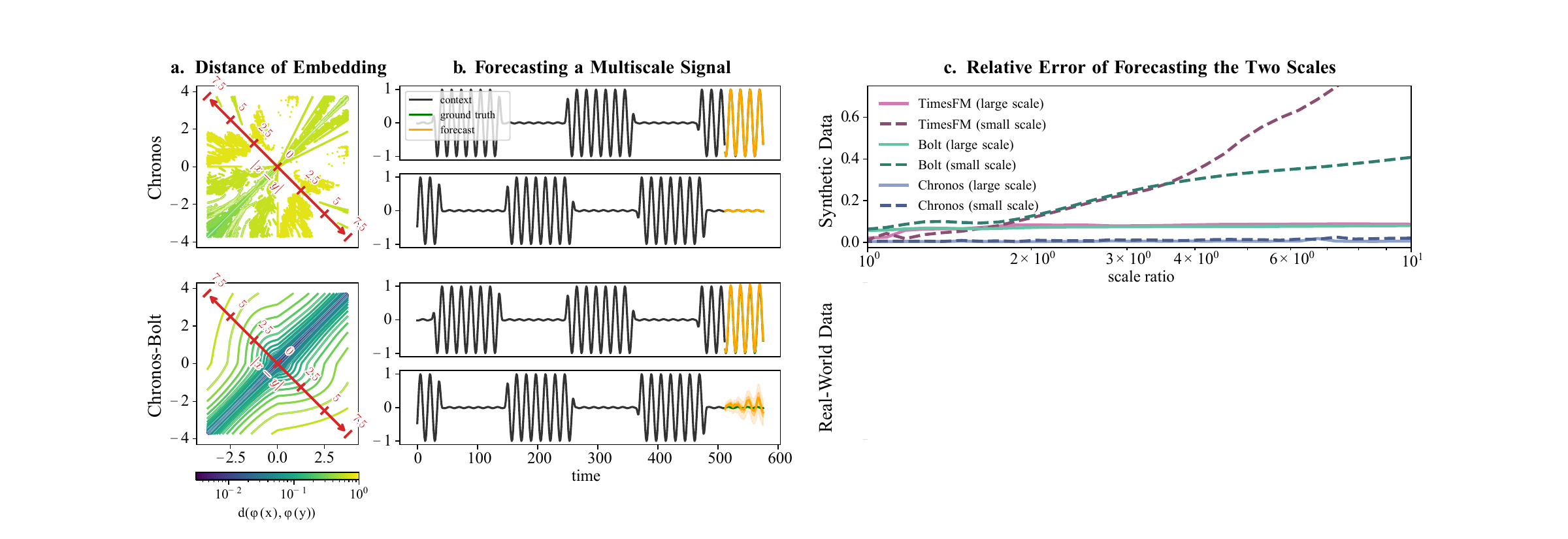}
    \put(55,12){
    \resizebox{0.42\textwidth}{!}{
\begin{tabular}{clcccccccc}
\toprule
\multicolumn{2}{c}{scale ratio} & $1$ & $4/3$ & $2$ & $4$ & $8$ & $40/3$ & $20$ & $40$ \\
\midrule
\multirow{3}{*}{\rotatebox{90}{ large scale}} &Chronos&\gradientscale{1.0000}&\gradientscale{1.0057}&\gradientscale{1.0088}&\gradientscale{1.0191}&\gradientscale{1.0215}&\gradientscale{1.0234}&\gradientscale{1.0327}&\gradientscale{1.0382} \\
&Bolt&\gradientscale{1.0000}&\gradientscale{1.0047}&\gradientscale{1.0063}&\gradientscale{1.0140}&\gradientscale{1.0218}&\gradientscale{1.0271}&\gradientscale{1.0333}&\gradientscale{1.0341} \\
&TimesFM&\gradientscale{1.0000}&\gradientscale{1.0031}&\gradientscale{1.0045}&\gradientscale{1.0091}&\gradientscale{1.0150}&\gradientscale{1.0197}&\gradientscale{1.0224}&\gradientscale{1.0266} \\
&Moirai&\gradientscale{1.0000}&\gradientscale{1.0014}&\gradientscale{1.0050}&\gradientscale{1.0101}&\gradientscale{1.0227}&\gradientscale{1.0299}&\gradientscale{1.0329}&\gradientscale{1.0401} \\
\midrule
\multirow{3}{*}{\rotatebox{90}{small scale}} &Chronos&\gradientscale{1.0000}&\gradientscale{1.0291}&\gradientscale{1.0133}&\gradientscale{1.0755}&\gradientscale{1.1882}&\gradientscale{1.2749}&\gradientscale{1.3602}&\gradientscale{1.5559} \\
&Bolt&\gradientscale{1.0000}&\gradientscale{1.0815}&\gradientscale{1.1098}&\gradientscale{1.2207}&\gradientscale{1.3687}&\gradientscale{1.5538}&\gradientscale{1.7256}&\gradientscale{2.6792} \\
&TimesFM&\gradientscale{1.0000}&\gradientscale{1.1293}&\gradientscale{1.1622}&\gradientscale{1.3805}&\gradientscale{1.5120}&\gradientscale{1.8436}&\gradientscale{2.0182}&\gradientscale{3.2009} \\
&Moirai&\gradientscale{1.0000}&\gradientscale{1.0742}&\gradientscale{1.1131}&\gradientscale{1.1922}&\gradientscale{1.3186}&\gradientscale{1.4866}&\gradientscale{1.7705}&\gradientscale{2.4818} \\
\bottomrule
\end{tabular}}
    }
    \end{overpic}
    \vspace{-2\baselineskip}
    \caption{Panel (a) shows the distance between the embedded vectors for Chronos' embedding $\boldsymbol{\phi}_Q$ and Chronos-Bolt's embedding $\boldsymbol{\phi}_C$. Panel (b) shows Chronos' and Chronos-Bolt's forecasts of a multi-scale time series. Panel (c) quantifies the forecasting quality by computing the relative MSE, as we increase the ratio between the two scales. See~\Cref{app:geomtric} for more details.
    }
    \myvspace{-\baselineskip}
    \label{fig:scale}
\end{figure}

\textbf{DISTANCES.}
Our second measurement of the geometric bias is the relative distance, defined as:
\[
    d_Q(x,y) = \frac{\|\boldsymbol{\phi}_Q(x) - \boldsymbol{\phi}_Q(y)\|_2}{\|\boldsymbol{\phi}_Q(x)\|_2 + \|\boldsymbol{\phi}_Q(y)\|_2}, 
    \quad
    d_C(x,y) = \frac{\|\boldsymbol{\phi}_C(x) - \boldsymbol{\phi}_C(y)\|_2}{\|\boldsymbol{\phi}_C(x)\|_2 + \|\boldsymbol{\phi}_C(y)\|_2}.
\]
Just as the learned quantization embedding $\boldsymbol{\phi}_Q$ bends the input space~$\mathbb{R}$ in the hidden space~$\mathbb{R}^d$, as we saw via angle measurement, it also stretches or squeezes~$\mathbb{R}$. 
While a continuous embedding is naturally expected to map nearby numbers to nearby vectors in the hidden space, a learned $\boldsymbol{\phi}_Q$ is generally less scale-preserving (see \Cref{fig:scale}a).

The mismatch in scale preservation introduces a \emph{distance bias}: a quantization-based embedding tends to magnify small scales, which makes nearby numbers appear more distinct in the hidden space. As a result, when a multi-scale structure is present, the Chronos model is more sensitive to and better at learning fine-scale patterns than a Chronos-Bolt model. \Cref{fig:scale}b well-illustrates this phenomenon: while both Chronos and Chronos-Bolt are good at learning the large-scale pattern, only Chronos nicely learns the small-scale patterns. A more systematic quantification is given in~\Cref{fig:scale}c, where we show that, as the ratio between the two scales increases, Chronos-Bolt, TimesFM, and Moirai, all of which rely on a continuous embeddings, are less capable of learning the fine-scale pattern. Whether this is an advantage or a drawback is again task-dependent (see~\Cref{app:goodandbad} for details): it can be valuable when fine-scale detail is important, e.g., when learning high-frequency information or when learning to perform context parroting; but may be unnecessary --- or even distracting --- in simpler situations when large-scale trends dominate.

\begin{figure}[!htb]
    \centering
    \myvspace{-0.5\baselineskip}
    \begin{overpic}[width=1\linewidth, trim=3cm 0cm 3cm 1cm, clip]{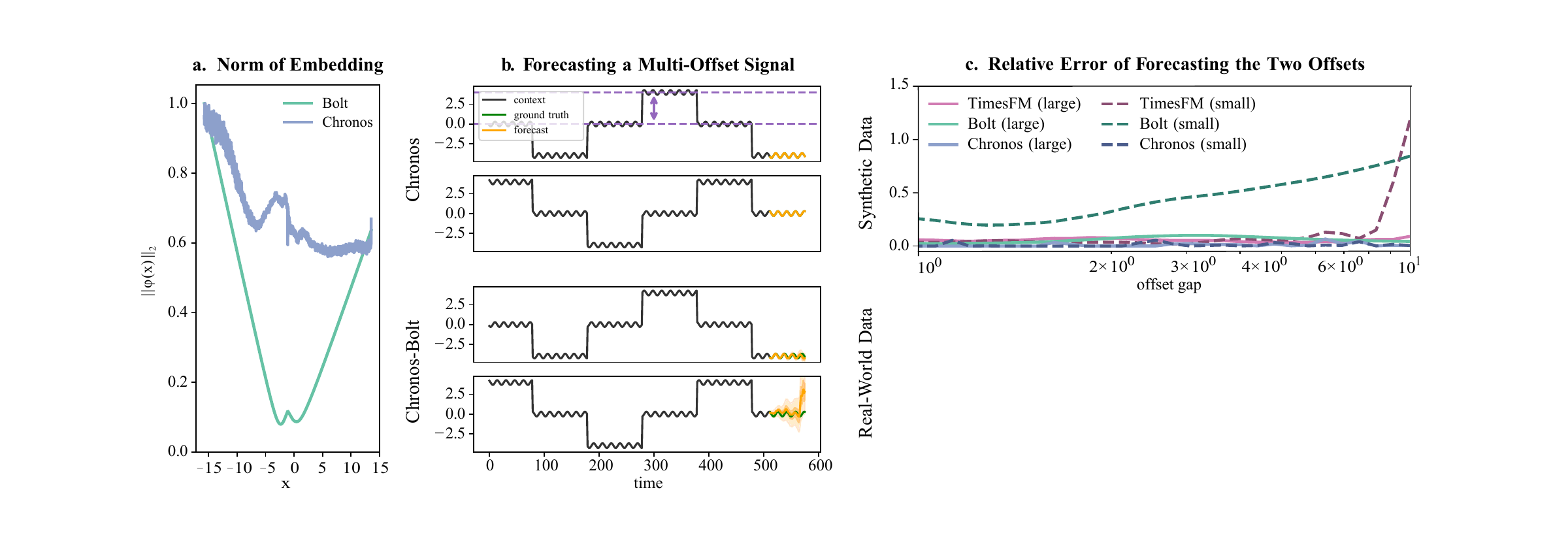}
    \put(57,12){
    \resizebox{0.42\textwidth}{!}{
\begin{tabular}{clcccccccc}
\toprule
\multicolumn{2}{c}{offset gap} & $1$ & $4/3$ & $2$ & $4$ & $8$ & $40/3$ & $20$ & $40$ \\
\midrule
\multirow{3}{*}{\rotatebox{90}{high offset}} &Chronos&\gradientoffset{1.0000}&\gradientoffset{1.0023}&\gradientoffset{1.0047}&\gradientoffset{1.0111}&\gradientoffset{1.0145}&\gradientoffset{1.0134}&\gradientoffset{1.0227}&\gradientoffset{1.0282} \\
&Bolt&\gradientoffset{1.0000}&\gradientoffset{1.0025}&\gradientoffset{1.0045}&\gradientoffset{1.0039}&\gradientoffset{1.0079}&\gradientoffset{1.0125}&\gradientoffset{1.0199}&\gradientoffset{1.0321} \\
&TimesFM&\gradientoffset{1.0000}&\gradientoffset{1.0019}&\gradientoffset{1.0008}&\gradientoffset{1.0048}&\gradientoffset{1.0090}&\gradientoffset{1.0121}&\gradientoffset{1.0179}&\gradientoffset{1.0277} \\
&Moirai&\gradientoffset{1.0000}&\gradientoffset{1.0090}&\gradientoffset{1.0059}&\gradientoffset{1.0099}&\gradientoffset{1.0177}&\gradientoffset{1.0241}&\gradientoffset{1.0250}&\gradientoffset{1.0328} \\
\midrule
\multirow{3}{*}{\rotatebox{90}{low offset}} &Chronos&\gradientoffset{1.0000}&\gradientoffset{1.1151}&\gradientoffset{1.1032}&\gradientoffset{1.0690}&\gradientoffset{1.1078}&\gradientoffset{1.1729}&\gradientoffset{1.3423}&\gradientoffset{1.6219} \\
&Bolt&\gradientoffset{1.0000}&\gradientoffset{1.1138}&\gradientoffset{1.1672}&\gradientoffset{1.2192}&\gradientoffset{1.3566}&\gradientoffset{1.6279}&\gradientoffset{2.1845}&\gradientoffset{3.1620} \\
&TimesFM&\gradientoffset{1.0000}&\gradientoffset{1.1781}&\gradientoffset{1.2995}&\gradientoffset{1.3808}&\gradientoffset{1.4780}&\gradientoffset{1.8239}&\gradientoffset{2.6884}&\gradientoffset{3.4165} \\
&Moirai&\gradientoffset{1.0000}&\gradientoffset{1.2196}&\gradientoffset{1.3142}&\gradientoffset{1.4234}&\gradientoffset{1.5517}&\gradientoffset{1.9059}&\gradientoffset{2.8262}&\gradientoffset{3.6152} \\
\bottomrule
\end{tabular}}
    }
    \end{overpic}
    \vspace{-3\baselineskip}
    \caption{Panel (a) shows the norm of the embedded vectors $\boldsymbol{\phi}_Q(x)$ and $\boldsymbol{\phi}_C(x)$. Panel (b) shows the forecasts of a multi-offset time series for Chronos and Chronos-Bolt, respectively. Panel (c) quantifies the forecasting quality by computing the relative (to each stage's amplitude) MSE of the forecasts as we increase the offset. See~\Cref{app:geomtric} for more details.}
    \label{fig:offset}
    \myvspace{-0.5\baselineskip}
\end{figure}

\textbf{NORMS.} 
Our last measurement of the geometric bias is the norm, which concerns the mapping of a single element, rather than pairs of elements, as in the other two cases. For a continuous embedding $\boldsymbol{\phi}_C$, many TSFMs use a ReLU-activated MLP.  
Because the ReLU activation is positive-homogeneous, $\boldsymbol{\phi}_C$ maps small input values to small vectors in $\mathbb{R}^d$ and large values to large vectors.
A quantization-based embedding $\boldsymbol{\phi}_Q$ behaves differently. Since the mapping of each bin, $\boldsymbol{\phi}_Q(D_i)$, is initialized from the same distribution, it has no inherent bias toward preserving norms (see~\Cref{fig:offset}a).

The tendency of $\boldsymbol{\phi}_C$ to map large numbers to large vectors leads to a \emph{norm bias}.
In~\Cref{fig:offset}b, the scale of each period is the same; the only difference is their offsets.  
When a period is far from zero, its embedded vectors occupy a larger portion of the embedded context, which makes it easier for the model to learn.  
Conversely, a period near zero is mapped to small vectors that are more easily overwhelmed by the rest of the context.  
\Cref{fig:offset}c illustrates this effect: as the offset increases (equivalently, as the period’s oscillation amplitude decreases while the offset stays fixed), Chronos-Bolt struggles to forecast the near-zero period. 
For regular signals with multiple offsets, this can introduce an undesirable imbalance; but for sparse signals, it can be advantageous, by allowing the model to focus on nonzero entries and ignoring the zeros.
(See~\Cref{app:goodandbad} for beneficial and deleterious examples.)

\begin{tcolorbox}[highlightbox]
\textbf{Summary of the Geometric Bias:} Geometric bias concerns how the geometry of the input domain is preserved by embeddings, and it is controlled by the choice between continuous and quantization-based strategies.  
From the angle perspective, quantization induces an \emph{angular bias}, making the model favor local information.  
From the distance perspective, continuous embeddings advocate a \emph{distance bias}, making fine-scale patterns less learnable.  
From the norm perspective, continuous embeddings induce a \emph{norm bias}, weighting large-magnitude elements more heavily, while quantization embeddings treat inputs more evenly.
\end{tcolorbox}

\section{Regression-to-the-Mean Bias: Do TSFMs Regress to the Mean?}
\label{sec:regression}

In this section, we identify and analyze a \emph{regression-to-the-mean} bias, which arises when a TSFM faces uncertainty, i.e., when the future is stochastic. In this case, should the model settle on a safe, compromised prediction, e.g., regress to the mean forecast, or should the model aggressively commit to one of the possible outcomes? 
Recent work on chaotic systems provides an illustration of this dilemma and the benefits of the latter approach~\citep{zhang2024zero, zhang2025context}.
Here, we analyze the effects of various design choices on how aggressively (or not) a TSFM regresses to the mean.
We outline key considerations in this section, and leave additional details to~\Cref{app:regression}.

How a model reacts to uncertainty heavily depends on how it is trained. 
Most TSFMs are trained with either an $L^2$-based loss (e.g., mean-squared loss) or an $L^1$-based loss (e.g., quantile loss). 
A notable exception is Chronos, which is trained with a cross-entropy loss, like LLMs.
If our target comes from an unknown probability distribution with large variance, then:  
\myvspace{-0.5\baselineskip}
\begin{itemize}[leftmargin=*]
    \item \textbf{$L^2$-based loss:} favors the \emph{mean} of the distribution, by averaging all possible outcomes.  

    \myvspace{-0.25\baselineskip}
    
    \item \textbf{$L^1$-based loss:} favors the \emph{median}, which gives a central split of the probability mass in half.  

    \myvspace{-0.25\baselineskip}
    
    \item \textbf{Cross-entropy loss:} models the full probability distribution~\citep{stewart2023regressionclassificationinfluencetask} and can settle on a/the ``\emph{mode}.'' This can avoid a ``compromise'' between possible outcomes, and it allows the model to represent distributions with large variances and stronger heterogeneity more faithfully.
\end{itemize}
\myvspace{-0.5\baselineskip}
A $L^2$- or $L^1$-based loss explicitly tailors the forecast to minimize pointwise error metrics such as MAE or MSE. From traditional wisdom, a low MAE or MSE is usually good, but time series forecasting is not all about pointwise fitting. For instance, in chaotic systems, the fractal dimension is a measurement of the long-term geometry of the trajectories, and regressing to the median or the mean can severely damage it~\citep{zhang2024zero}.
(See~\Cref{app:goodandbad} for cases where the regression bias is ``beneficial'' or ``deleterious.'') 
Unlike mean- or median-based forecasts, ``regressing to the mode'' lets the prediction reflect a sharp, high-probability outcomes, instead of a smoothed central tendency.

\begin{figure}[!htb]
    \centering
    \includegraphics[width=1\linewidth]{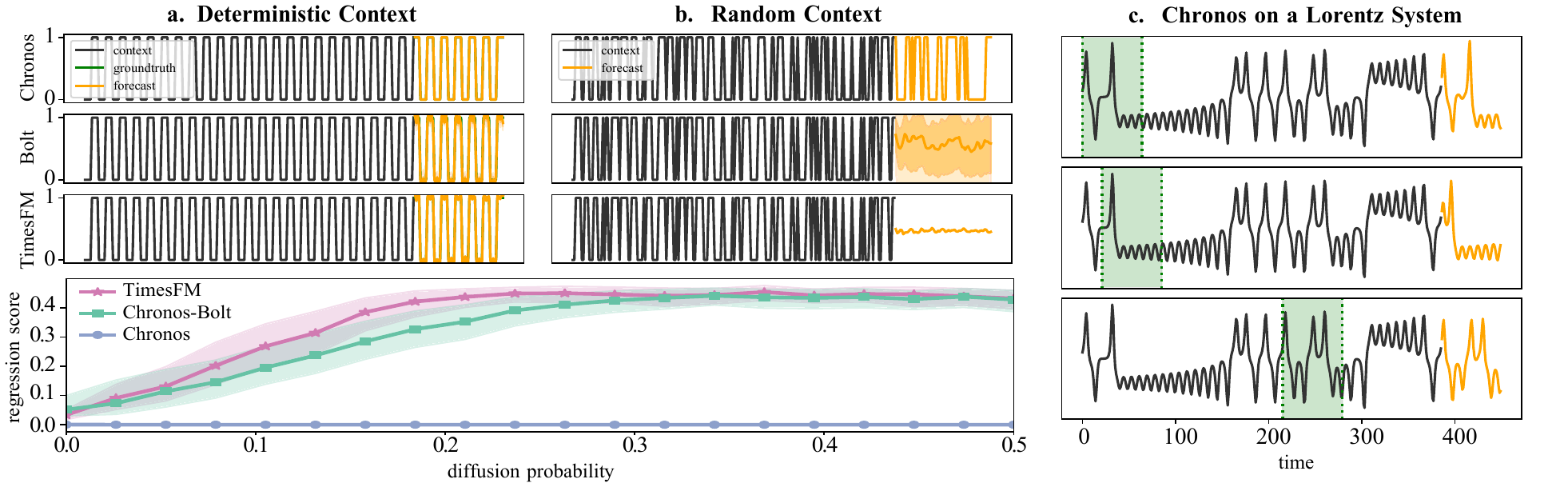}
    \caption{Panels (a) and (b) show the forecasts from three TSFMs of a purely deterministic and a purely random context, respectively. We also quantify the ``regression score'' by computing $\min(|\hat{y}|, |1-\hat{y}|)$, where $\hat{y}$ denotes the forecast from a model. As the context involves more uncertainty, TimesFM and Chronos-Bolt regress to the mean and the median, respectively, more than Chronos. Panel (c) shows three examples where Chronos ``parrots'' three distinct outcome branches from the context of a Lorentz chaotic system. See~\Cref{app:regression} for more~details.}
    \label{fig:regression}
\end{figure}

We illustrate the regression-to-the-mean bias using the examples in~\Cref{fig:regression}.  
In (a), we show a periodic walk between two branches, $0$ and $1$.  
Since the process is fully deterministic, Chronos (cross-entropy loss), Chronos-Bolt ($L^1$ loss), and TimesFM ($L^2$ loss) produce similar forecasts.  
Panel (b) uses a context generated by a random walk.  
With probability $1/2$ on each branch, the uncertainty in the context drives the forecast of TimesFM toward the mean, Chronos-Bolt toward the median (any value in $[0,1]$), and Chronos toward the mode (either $0$ or $1$).  
We can also quantify this effect by ``bridging'' the purely deterministic and purely random cases. For each $0 \leq p \leq 1/2$, we take a periodic walk and diffuse it with probability~$p$.  
As $p$ increases --- i.e., the context becomes more uncertain --- models trained with a continuous loss regress more strongly than the one trained with cross-entropy.

\begin{tcolorbox}[highlightbox]
\textbf{Summary of the Regression-to-the-Mean Bias:} 
Models trained with $L^2$- or $L^1$-based losses tend to collapse future variability into a central forecast, reflecting mean or median behavior and usually resulting in a good pointwise fitting.  
In contrast, cross-entropy-based training represents the full distribution and can preserve sharper, more extreme, outcomes, instead of aggressively regressing to a central tendency.
\end{tcolorbox}

\section{Mixture of Biases: Outlier Handling}
\label{sxn:application}

\begin{figure}[!htb]
    \centering
    \myvspace{-1\baselineskip}
    \includegraphics[width=1\linewidth]{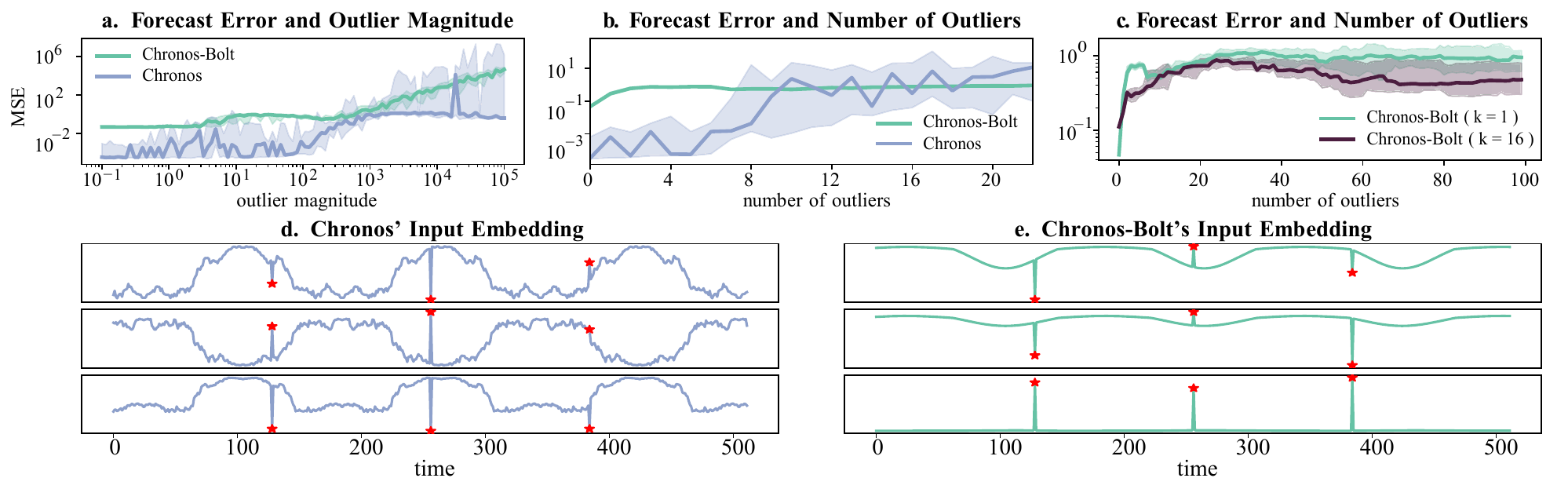}
    \myvspace{-2\baselineskip}
    \caption{Top row shows the MSE of the baseline models of forecasting a sinusoidal wave with randomly injected outliers. Panel (a) shows the effect of increasing the outlier's magnitude, and panels (b) and (c) show the MSE as we increase the number of outliers. We repeat the experiments are $30$ times. Panels (d) and (e) show the first principal component of the embedded input in Chronos and Chronos-Bolt as we increase the magnitude of the outlier.}
    \label{fig:outlier}
\end{figure}

In the previous three sections, we have identified three key knobs that affect three fundamental implicit biases in the development of TSFMs. 
Of course, in practical settings, these biases can interact in complex ways, and the effect of fiddling with a given knob can be nontrivial to characterize.
In this section, we present outlier handling as an illustration of how different biases can combine in natural ways.
We use a sinusoidal wave as the context; and we inject outliers into it, in  two scenarios: 
(1) increasing the magnitude of a fixed number of outliers; and 
(2) increasing the number of outliers sampled i.i.d.\ from a fixed Gaussian distribution.
Our results, as shown in~\Cref{fig:outlier}, are the following:

\myvspace{-0.5\baselineskip}
\begin{enumerate}[leftmargin=*]
    \item 
    \textbf{Temporal bias.} 
    Comparing the two Chronos-Bolt models with patch sizes $k=1$ and $k=16$ shows that, without outliers, the $k=1$ model performs better. As more outliers are injected, they act as high-frequency perturbations of the context, which makes the larger patch size more effective at denoising (see~\Cref{fig:outlier}c).
    \myvspace{-0.25\baselineskip}
    \item 
    \textbf{Geometric (Distance) bias.} 
    When outlier contamination is limited, i.e., their magnitude and number are relatively small, Chronos outperforms Chronos-Bolt by reducing the apparent scale of outliers through its quantization-based embedding (see~\Cref{fig:outlier}d).
    \\
    \textbf{Geometric (Norm) bias.} 
    Chronos-Bolt’s continuous embedding causes large-magnitude outliers to dominate the embedded context; and, as the outlier magnitude increases, this dominance leads to progressively worse forecasts (see~\Cref{fig:outlier}a,e).
    \myvspace{-0.25\baselineskip}
    \item 
    \textbf{Regression-to-the-Mean bias.} 
    Although Chronos handles outliers well on average, its performance variance is larger than that of Chronos-Bolt. This is because the regression-to-the-mean bias sometimes causes Chronos to produce an ``outlier'' when it is uncertain whether the observed outliers are part of the underlying context (see~\Cref{fig:outlier}b).
\end{enumerate}
\myvspace{-0.5\baselineskip}

\section{Conclusion}
\label{sxn:conclusion}

We identify and characterize three broad classes of inductive biases in TSFMs that influence their strengths and weaknesses across tasks in subtle and sometimes counterintuitive ways.
Understanding these biases helps explain why different models excel in different settings, and it helps to clarify trade-offs among architectural and training choices.
By making these biases explicit, and relating them to well-known knobs of the training process, practitioners can better match model design to application needs, leveraging favorable effects while mitigating harmful ones.
Our findings clearly have implications for learning ``simple'' models at scale (some initial ideas are in \Cref{app:simplicity}) and for developing TSFMs that are forward-compatible with orders of magnitude more data and compute---these are important follow-up directions.

\subsubsection*{Acknowledgments}

The authors would like to thank William Gilpin and Yuanzhao Zhang for many fruitful discussions on this work.

\bibliography{iclr2026_conference}

\begin{thebibliography}{77}
\providecommand{\natexlab}[1]{#1}
\providecommand{\url}[1]{\texttt{#1}}
\expandafter\ifx\csname urlstyle\endcsname\relax
  \providecommand{\doi}[1]{doi: #1}\else
  \providecommand{\doi}{doi: \begingroup \urlstyle{rm}\Url}\fi

\bibitem[Ansari et~al.(2024{\natexlab{a}})Ansari, Stella, Turkmen, Zhang,
  Mercado, Shen, Shchur, Rangapuram, Arango, Kapoor, et~al.]{ansari2024chronos}
Abdul~Fatir Ansari, Lorenzo Stella, Caner Turkmen, Xiyuan Zhang, Pedro Mercado,
  Huibin Shen, Oleksandr Shchur, Syama~Sundar Rangapuram, Sebastian~Pineda
  Arango, Shubham Kapoor, et~al.
\newblock Chronos: Learning the language of time series.
\newblock \emph{arXiv preprint arXiv:2403.07815}, 2024{\natexlab{a}}.

\bibitem[Ansari et~al.(2024{\natexlab{b}})Ansari, Turkmen, Shchur, and
  Stella]{ChronosBolt}
Abdul~Fatir Ansari, Caner Turkmen, Oleksandr Shchur, and Lorenzo Stella.
\newblock Fast and accurate zero-shot forecasting with {C}hronos-{B}olt and
  {A}utogluon, 2024{\natexlab{b}}.
\newblock URL
  \url{https://aws.amazon.com/blogs/machine-learning/fast-and-accurate-zero-shot-forecasting-with-chronos-bolt-and-autogluon/}.

\bibitem[Arora et~al.(2019)Arora, Du, Hu, Li, and Wang]{arora}
S.~Arora, S.~S. Du, W.~Hu, Z.~Li, and R.~Wang.
\newblock Fine-grained analysis of optimization and generalization for
  overparameterized two-layer neural networks.
\newblock In \emph{Inter. Conf. Mach. Learn.}, pp.\  322--332. PMLR, 2019.

\bibitem[Athanasopoulos et~al.(2011)Athanasopoulos, Hyndman, Song, and
  Wu]{tourism_kaggle}
George Athanasopoulos, Rob~J. Hyndman, Haiyan Song, and Doris~C. Wu.
\newblock The tourism forecasting competition.
\newblock \emph{International Journal of Forecasting}, 27\penalty0
  (3):\penalty0 822--844, 2011.

\bibitem[Basri et~al.(2019)Basri, Jacobs, Kasten, and Kritchman]{basri}
R.~Basri, D.~Jacobs, Y.~Kasten, and S.~Kritchman.
\newblock The convergence rate of neural networks for learned functions of
  different frequencies.
\newblock 32, 2019.

\bibitem[Basri et~al.(2020)Basri, Galun, Geifman, Jacobs, Kasten, and
  Kritchman]{basri2020frequency}
R.~Basri, M.~Galun, A.~Geifman, D.~Jacobs, Y.~Kasten, and S.~Kritchman.
\newblock Frequency bias in neural networks for input of non-uniform density.
\newblock In \emph{Inter. Conf. Mach. Learn.}, pp.\  685--694. PMLR, 2020.

\bibitem[Belrose et~al.(2024)Belrose, Pope, Quirke, Mallen, and
  Fern]{belrose2024neural}
Nora Belrose, Quintin Pope, Lucia Quirke, Alex Mallen, and Xiaoli Fern.
\newblock Neural networks learn statistics of increasing complexity.
\newblock \emph{arXiv preprint arXiv:2402.04362}, 2024.

\bibitem[Bietti \& Mairal(2019)Bietti and Mairal]{bietti2019inductive}
A.~Bietti and J.~Mairal.
\newblock On the inductive bias of neural tangent kernels.
\newblock 32, 2019.

\bibitem[Cao et~al.(2019)Cao, Fang, Wu, Zhou, and Gu]{cao2019towards}
Yuan Cao, Zhiying Fang, Yue Wu, Ding-Xuan Zhou, and Quanquan Gu.
\newblock Towards understanding the spectral bias of deep learning.
\newblock \emph{arXiv preprint arXiv:1912.01198}, 2019.

\bibitem[Czarnecki et~al.(2017)Czarnecki, Osindero, Jaderberg, Swirszcz, and
  Pascanu]{czarnecki2017sobolev}
W.~M. Czarnecki, S.~Osindero, M.~Jaderberg, G.~Swirszcz, and R.~Pascanu.
\newblock Sobolev training for neural networks.
\newblock 30, 2017.

\bibitem[Das et~al.(2024)Das, Kong, Sen, and Zhou]{das2024decoder}
Abhimanyu Das, Weihao Kong, Rajat Sen, and Yichen Zhou.
\newblock A decoder-only foundation model for time-series forecasting.
\newblock In \emph{Forty-first International Conference on Machine Learning},
  2024.

\bibitem[Doan et~al.(2020)Doan, Polifke, and Magri]{doan2020physics}
Nguyen Anh~Khoa Doan, Wolfgang Polifke, and Luca Magri.
\newblock Physics-informed echo state networks.
\newblock \emph{Journal of Computational Science}, 47:\penalty0 101237, 2020.

\bibitem[Eisenach et~al.(2022)Eisenach, Patel, and
  Madeka]{eisenach2022mqtransformermultihorizonforecastscontext}
Carson Eisenach, Yagna Patel, and Dhruv Madeka.
\newblock Mqtransformer: Multi-horizon forecasts with context dependent and
  feedback-aware attention.
\newblock \emph{arXiv preprint arXiv:2009.14799}, 2022.

\bibitem[Gardner~Jr(1985)]{gardner1985exponential}
Everette~S Gardner~Jr.
\newblock Exponential smoothing: The state of the art.
\newblock \emph{Journal of forecasting}, 4\penalty0 (1):\penalty0 1--28, 1985.

\bibitem[Gilpin(2021)]{gilpin2021chaos}
William Gilpin.
\newblock Chaos as an interpretable benchmark for forecasting and data-driven
  modelling.
\newblock \emph{arXiv preprint arXiv:2110.05266}, 2021.

\bibitem[Gneiting \& Katzfuss(2014)Gneiting and
  Katzfuss]{gneiting2014probabilistic}
Tilmann Gneiting and Matthias Katzfuss.
\newblock Probabilistic forecasting.
\newblock \emph{Annual Review of Statistics and Its Application}, 1\penalty0
  (1):\penalty0 125--151, 2014.

\bibitem[Gneiting et~al.(2007)Gneiting, Balabdaoui, and
  Raftery]{gneiting2007probabilistic}
Tilmann Gneiting, Fadoua Balabdaoui, and Adrian~E Raftery.
\newblock Probabilistic forecasts, calibration and sharpness.
\newblock \emph{Journal of the Royal Statistical Society Series B: Statistical
  Methodology}, 69\penalty0 (2):\penalty0 243--268, 2007.

\bibitem[Godahewa et~al.(2021)Godahewa, Bergmeir, Webb, Hyndman, and
  Montero-Manso]{godahewa2021}
Rakshitha Godahewa, Christoph Bergmeir, Geoffrey~I. Webb, Rob~J. Hyndman, and
  Pablo Montero-Manso.
\newblock Monash time series forecasting archive.
\newblock In \emph{Advances in neural information processing systems track on
  datasets and benchmarks}, 2021.

\bibitem[Goldblum et~al.(2024)Goldblum, Finzi, Rowan, and
  Wilson]{goldblum2023no}
Micah Goldblum, Marc Finzi, Keefer Rowan, and Andrew~Gordon Wilson.
\newblock The no free lunch theorem, kolmogorov complexity, and the role of
  inductive biases in machine learning.
\newblock \emph{PMLR}, 2024.

\bibitem[Goswami et~al.(2024)Goswami, Szafer, Choudhry, Cai, Li, and
  Dubrawski]{goswami2024moment}
Mononito Goswami, Konrad Szafer, Arjun Choudhry, Yifu Cai, Shuo Li, and Artur
  Dubrawski.
\newblock Moment: A family of open time-series foundation models.
\newblock \emph{arXiv preprint arXiv:2402.03885}, 2024.

\bibitem[Gruver et~al.(2023)Gruver, Finzi, Qiu, and Wilson]{gruver2023large}
Nate Gruver, Marc Finzi, Shikai Qiu, and Andrew~G Wilson.
\newblock Large language models are zero-shot time series forecasters.
\newblock \emph{Advances in Neural Information Processing Systems},
  36:\penalty0 19622--19635, 2023.

\bibitem[Gu \& Dao(2023)Gu and Dao]{gu2023mamba}
Albert Gu and Tri Dao.
\newblock Mamba: Linear-time sequence modeling with selective state spaces.
\newblock \emph{arXiv preprint arXiv:2312.00752}, 2023.

\bibitem[Hong et~al.(2016)Hong, Pinson, Fan, Zareipour, Troccoli, and
  Hyndman]{hong2016probabilistic}
Tao Hong, Pierre Pinson, Shu Fan, Hamidreza Zareipour, Alberto Troccoli, and
  Rob~J Hyndman.
\newblock Probabilistic energy forecasting: Global energy forecasting
  competition 2014 and beyond, 2016.

\bibitem[Huh et~al.(2021)Huh, Mobahi, Zhang, Cheung, Agrawal, and
  Isola]{huh2021low}
Minyoung Huh, Hossein Mobahi, Richard Zhang, Brian Cheung, Pulkit Agrawal, and
  Phillip Isola.
\newblock The low-rank simplicity bias in deep networks.
\newblock \emph{arXiv preprint arXiv:2103.10427}, 2021.

\bibitem[Hyndman \& Athanasopoulos(2018)Hyndman and
  Athanasopoulos]{hyndman2018forecasting}
Rob~J Hyndman and George Athanasopoulos.
\newblock \emph{Forecasting: principles and practice}.
\newblock OTexts, 2018.

\bibitem[Jacot et~al.(2018)Jacot, Gabriel, and Hongler]{jacot2018neural}
A.~Jacot, F.~Gabriel, and C.~Hongler.
\newblock Neural tangent kernel: Convergence and generalization in neural
  networks.
\newblock 31, 2018.

\bibitem[{James M. Kilts Center}(2020)]{dominick}
{James M. Kilts Center}.
\newblock Dominick's dataset, 2020.
\newblock URL
  \url{https://www.chicagobooth.edu/research/kilts/research-data/dominicks}.

\bibitem[Lai et~al.(2017)Lai, Chang, Yang, and Liu]{traffic}
G.~Lai, W.~Chang, Y.~Yang, and H.~Liu.
\newblock Modeling long- and short-term temporal patterns with deep neural
  networks.
\newblock \emph{arXiv preprint arXiv:1703.07015}, 2017.

\bibitem[Lai et~al.(2025)Lai, Bao, and Gilpin]{lai2025panda}
Jeffrey Lai, Anthony Bao, and William Gilpin.
\newblock Panda: A pretrained forecast model for universal representation of
  chaotic dynamics.
\newblock \emph{arXiv preprint arXiv:2505.13755}, 2025.

\bibitem[Liang et~al.(2025)Liang, Hou, Yao, Wang, Jiang, Han, and
  Huang]{liang2025TSGym}
Shuang Liang, Chaochuan Hou, Xu~Yao, Shiping Wang, Minqi Jiang, Songqiao Han,
  and Hailiang Huang.
\newblock Tsgym.
\newblock \url{https://github.com/SUFE-AILAB/TSGym}, 2025.

\bibitem[Liang et~al.(2024)Liang, Wen, Nie, Jiang, Jin, Song, Pan, and
  Wen]{liang2024foundation}
Yuxuan Liang, Haomin Wen, Yuqi Nie, Yushan Jiang, Ming Jin, Dongjin Song,
  Shirui Pan, and Qingsong Wen.
\newblock Foundation models for time series analysis: A tutorial and survey.
\newblock In \emph{Proceedings of the 30th ACM SIGKDD conference on knowledge
  discovery and data mining}, pp.\  6555--6565, 2024.

\bibitem[Lim et~al.(2021)Lim, Ar{\i}k, Loeff, and Pfister]{lim2021temporal}
Bryan Lim, Sercan~{\"O} Ar{\i}k, Nicolas Loeff, and Tomas Pfister.
\newblock Temporal fusion transformers for interpretable multi-horizon time
  series forecasting.
\newblock \emph{International journal of forecasting}, 37\penalty0
  (4):\penalty0 1748--1764, 2021.

\bibitem[Liu et~al.(2024{\natexlab{a}})Liu, Xu, Cao, and
  Zhang]{liu2024mitigating}
Xinliang Liu, Bo~Xu, Shuhao Cao, and Lei Zhang.
\newblock Mitigating spectral bias for the multiscale operator learning.
\newblock \emph{Journal of Computational Physics}, 506:\penalty0 112944,
  2024{\natexlab{a}}.
\newblock ISSN 0021-9991.

\bibitem[Liu et~al.(2024{\natexlab{b}})Liu, Hu, Li, Diao, Liang, Hooi, and
  Zimmermann]{liu2024unitime}
Xu~Liu, Junfeng Hu, Yuan Li, Shizhe Diao, Yuxuan Liang, Bryan Hooi, and Roger
  Zimmermann.
\newblock Unitime: A language-empowered unified model for cross-domain time
  series forecasting.
\newblock In \emph{Proceedings of the ACM Web Conference 2024}, pp.\
  4095--4106, 2024{\natexlab{b}}.

\bibitem[Liu et~al.(2023)Liu, Hu, Zhang, Wu, Wang, Ma, and
  Long]{liu2023itransformer}
Yong Liu, Tengge Hu, Haoran Zhang, Haixu Wu, Shiyu Wang, Lintao Ma, and
  Mingsheng Long.
\newblock itransformer: Inverted transformers are effective for time series
  forecasting.
\newblock \emph{arXiv preprint arXiv:2310.06625}, 2023.

\bibitem[Liu et~al.(2024{\natexlab{c}})Liu, Zhang, Li, Huang, Wang, and
  Long]{liu2024timer}
Yong Liu, Haoran Zhang, Chenyu Li, Xiangdong Huang, Jianmin Wang, and Mingsheng
  Long.
\newblock Timer: Generative pre-trained transformers are large time series
  models.
\newblock \emph{arXiv preprint arXiv:2402.02368}, 2024{\natexlab{c}}.

\bibitem[Ma et~al.(2024)Ma, Chen, Zhao, Yang, Ji, Xu, Liu, Jing, Liu, and
  Yang]{ma2024mamba}
Haoyu Ma, Yushu Chen, Wenlai Zhao, Jinzhe Yang, Yingsheng Ji, Xinghua Xu,
  Xiaozhu Liu, Hao Jing, Shengzhuo Liu, and Guangwen Yang.
\newblock A mamba foundation model for time series forecasting.
\newblock \emph{arXiv preprint arXiv:2411.02941}, 2024.

\bibitem[Makridakis \& Hibon(1997)Makridakis and Hibon]{makridakis1997arma}
Spyros Makridakis and Michele Hibon.
\newblock Arma models and the box--jenkins methodology.
\newblock \emph{Journal of forecasting}, 16\penalty0 (3):\penalty0 147--163,
  1997.

\bibitem[Makridakis et~al.(2020)Makridakis, Spiliotis, and
  Assimakopoulos]{makridakis2020m4}
Spyros Makridakis, Evangelos Spiliotis, and Vassilios Assimakopoulos.
\newblock The m4 competition: 100{,}000 time series and 61 forecasting methods.
\newblock \emph{International Journal of Forecasting}, 36\penalty0
  (1):\penalty0 54--74, 2020.
\newblock \doi{10.1016/j.ijforecast.2019.04.014}.

\bibitem[Masserano et~al.(2025)Masserano, Ansari, Han, Zhang, Faloutsos,
  Mahoney, Wilson, Park, Rangapuram, Maddix, et~al.]{masserano2024enhancing}
Luca Masserano, Abdul~Fatir Ansari, Boran Han, Xiyuan Zhang, Christos
  Faloutsos, Michael~W Mahoney, Andrew~Gordon Wilson, Youngsuk Park, Syama
  Rangapuram, Danielle~C Maddix, et~al.
\newblock Enhancing foundation models for time series forecasting via
  wavelet-based tokenization.
\newblock In \emph{International Conference on Machine Learning}. PMLR, 2025.

\bibitem[Morwani et~al.(2023)Morwani, Batra, Jain, and
  Netrapalli]{morwani2023simplicity}
Depen Morwani, Jatin Batra, Prateek Jain, and Praneeth Netrapalli.
\newblock Simplicity bias in 1-hidden layer neural networks.
\newblock \emph{Advances in Neural Information Processing Systems},
  36:\penalty0 8048--8075, 2023.

\bibitem[Nie et~al.(2022)Nie, Nguyen, Sinthong, and Kalagnanam]{nie2022time}
Yuqi Nie, Nam~H Nguyen, Phanwadee Sinthong, and Jayant Kalagnanam.
\newblock A time series is worth 64 words: Long-term forecasting with
  transformers.
\newblock \emph{arXiv preprint arXiv:2211.14730}, 2022.

\bibitem[Oreshkin et~al.(2020)Oreshkin, Carpov, Chapados, and
  Bengio]{oreshkin2020nbeats}
Boris~N. Oreshkin, Dmitri Carpov, Nicolas Chapados, and Yoshua Bengio.
\newblock {N}-{B}{E}{A}{T}{S}: {N}eural basis expansion analysis for
  interpretable time series forecasting.
\newblock \emph{Internation Conference on Learning Representations}, 2020.

\bibitem[Park et~al.(2022)Park, Maddix, Aubet, Kan, Gasthaus, and
  Wang]{pmlr-v151-park22a}
Youngsuk Park, Danielle~C. Maddix, Fran\c{c}ois-Xavier Aubet, Kelvin Kan, Jan
  Gasthaus, and Yuyang Wang.
\newblock Learning quantile functions without quantile crossing for
  distribution-free time series forecasting.
\newblock In \emph{Proceedings of The 25th International Conference on
  Artificial Intelligence and Statistics}, volume 151, pp.\  8127--8150. PMLR,
  2022.

\bibitem[Pathak et~al.(2018)Pathak, Wikner, Fussell, Chandra, Hunt, Girvan, and
  Ott]{pathak2018hybrid}
Jaideep Pathak, Alexander Wikner, Rebeckah Fussell, Sarthak Chandra, Brian~R
  Hunt, Michelle Girvan, and Edward Ott.
\newblock Hybrid forecasting of chaotic processes: Using machine learning in
  conjunction with a knowledge-based model.
\newblock \emph{Chaos: An interdisciplinary journal of nonlinear science},
  28\penalty0 (4), 2018.

\bibitem[Piao et~al.(2024)Piao, Chen, Murayama, Matsubara, and
  Sakurai]{piao2024fredformer}
Xihao Piao, Zheng Chen, Taichi Murayama, Yasuko Matsubara, and Yasushi Sakurai.
\newblock Fredformer: Frequency debiased transformer for time series
  forecasting.
\newblock In \emph{Proceedings of the 30th ACM SIGKDD conference on knowledge
  discovery and data mining}, pp.\  2400--2410, 2024.

\bibitem[Rackauckas(2025)]{rackauckas2025how}
Christopher Rackauckas.
\newblock How chaotic is chaos? how some ai for science / sciml papers are
  overstating accuracy claims.
\newblock
  \url{https://www.stochasticlifestyle.com/how-chaotic-is-chaos-how-some-ai-for-science-sciml-papers-are-overstating-accuracy-claims/},
  2025.

\bibitem[Rahaman et~al.(2019)Rahaman, Baratin, Arpit, Draxler, Lin, Hamprecht,
  Bengio, and Courville]{rahaman2019spectral}
N.~Rahaman, A.~Baratin, D.~Arpit, F.~Draxler, M.~Lin, F.~Hamprecht, Y.~Bengio,
  and A.~Courville.
\newblock On the spectral bias of neural networks.
\newblock In \emph{Inter. Conf. Mach. Learn.}, pp.\  5301--5310. PMLR, 2019.

\bibitem[Rende et~al.(2024)Rende, Gerace, Laio, and
  Goldt]{rende2024distributional}
Riccardo Rende, Federica Gerace, Alessandro Laio, and Sebastian Goldt.
\newblock A distributional simplicity bias in the learning dynamics of
  transformers.
\newblock \emph{Advances in Neural Information Processing Systems},
  37:\penalty0 96207--96228, 2024.

\bibitem[Salinas et~al.(2020)Salinas, Flunkert, Gasthaus, and
  Januschowski]{salinas2020deepar}
David Salinas, Valentin Flunkert, Jan Gasthaus, and Tim Januschowski.
\newblock Deepar: Probabilistic forecasting with autoregressive recurrent
  networks.
\newblock \emph{International journal of forecasting}, 36\penalty0
  (3):\penalty0 1181--1191, 2020.

\bibitem[Schumaker \& Yu(2022)Schumaker and Yu]{schumaker2022approximation}
Larry~L Schumaker and Annan Yu.
\newblock Approximation by polynomial splines on curved triangulations.
\newblock \emph{Computer Aided Geometric Design}, 92:\penalty0 102050, 2022.

\bibitem[Shi et~al.(2025)Shi, Wang, Nie, Li, Ye, Wen, and Jin]{shi2024time}
Xiaoming Shi, Shiyu Wang, Yuqi Nie, Dianqi Li, Zhou Ye, Qingsong Wen, and Ming
  Jin.
\newblock Time-moe: Billion-scale time series foundation models with mixture of
  experts.
\newblock \emph{International Conference on Learning Representations}, 2025.

\bibitem[Son et~al.(2021)Son, Jang, Han, and Hwang]{son2021sobolev}
H.~Son, J.W. Jang, W.J. Han, and H.J. Hwang.
\newblock Sobolev training for the neural network solutions of {P}{D}{E}s.
\newblock \emph{arXiv preprint arXiv:2101.08932}, 2021.

\bibitem[Son(2023)]{sonsobolev}
Hwijae Son.
\newblock Sobolev acceleration for neural networks.
\newblock 2023.

\bibitem[Steger et~al.(2022)Steger, Rohrhofer, and Geiger]{steger2022pinns}
Sophie Steger, Franz~M Rohrhofer, and Bernhard~C Geiger.
\newblock How pinns cheat: Predicting chaotic motion of a double pendulum.
\newblock In \emph{The Symbiosis of Deep Learning and Differential Equations
  II}, 2022.

\bibitem[Stewart et~al.(2023)Stewart, Bach, Berthet, and
  Vert]{stewart2023regressionclassificationinfluencetask}
Lawrence Stewart, Francis Bach, Quentin Berthet, and Jean-Philippe Vert.
\newblock Regression as classification: Influence of task formulation on neural
  network features.
\newblock \emph{arXiv preprint arXiv:2211.05641}, 2023.

\bibitem[Su \& Yang(2019)Su and Yang]{suyang}
L.~Su and P.~Yang.
\newblock On learning over-parameterized neural networks: A functional
  approximation perspective.
\newblock volume~32. Curran Associates, Inc., 2019.

\bibitem[Talukder et~al.(2024)Talukder, Yue, and Gkioxari]{talukder2024totem}
Sabera Talukder, Yisong Yue, and Georgia Gkioxari.
\newblock Totem: Tokenized time series embeddings for general time series
  analysis.
\newblock \emph{arXiv preprint arXiv:2402.16412}, 2024.

\bibitem[Trindade(2015)]{trindade2015electricity}
Artur Trindade.
\newblock Electricityloaddiagrams20112014, 2015.
\newblock URL
  \url{https://archive.ics.uci.edu/dataset/321/electricityloaddiagrams20112014}.

\bibitem[Tsay(2021)]{tsay2021sobolev}
C.~Tsay.
\newblock Sobolev trained neural network surrogate models for optimization.
\newblock \emph{Comp. Chem. Eng.}, 153:\penalty0 107419, 2021.

\bibitem[Vlassis \& Sun(2021)Vlassis and Sun]{vlassis2021sobolev}
N.N. Vlassis and W.~Sun.
\newblock Sobolev training of thermodynamic-informed neural networks for
  interpretable elasto-plasticity models with level set hardening.
\newblock \emph{Compu. Meth. Appl. Mech. Eng.}, 377:\penalty0 113695, 2021.

\bibitem[Wen et~al.(2018)Wen, Torkkola, Narayanaswamy, and
  Madeka]{wen2018multihorizonquantilerecurrentforecaster}
Ruofeng Wen, Kari Torkkola, Balakrishnan Narayanaswamy, and Dhruv Madeka.
\newblock A multi-horizon quantile recurrent forecaster.
\newblock \emph{arXiv preprint arXiv:1711.11053}, 2018.

\bibitem[Wilson(2025)]{wilson2025deep}
Andrew~Gordon Wilson.
\newblock Deep learning is not so mysterious or different.
\newblock \emph{PMLR}, 2025.

\bibitem[Wolpert \& Macready(2002)Wolpert and Macready]{wolpert2002no}
David~H Wolpert and William~G Macready.
\newblock No free lunch theorems for optimization.
\newblock \emph{IEEE transactions on evolutionary computation}, 1\penalty0
  (1):\penalty0 67--82, 2002.

\bibitem[Woo et~al.(2024)Woo, Liu, Kumar, Xiong, Savarese, and
  Sahoo]{woo2024unified}
Gerald Woo, Chenghao Liu, Akshat Kumar, Caiming Xiong, Silvio Savarese, and
  Doyen Sahoo.
\newblock Unified training of universal time series forecasting transformers.
\newblock 2024.

\bibitem[Xu(2020)]{xu2020frequency}
Z.-Q.~J. Xu.
\newblock Frequency principle: Fourier analysis sheds light on deep neural
  networks.
\newblock \emph{Commun. Comput. Phys.}, 28\penalty0 (5):\penalty0 1746--1767,
  2020.

\bibitem[Yang \& Salman(2019)Yang and Salman]{yang2019fine}
G.~Yang and H.~Salman.
\newblock A fine-grained spectral perspective on neural networks.
\newblock \emph{arXiv preprint arXiv:1907.10599}, 2019.

\bibitem[Yu et~al.(2023{\natexlab{a}})Yu, Nigmetov, Morozov, Mahoney, and
  Erichson]{yu2023robustifying}
Annan Yu, Arnur Nigmetov, Dmitriy Morozov, Michael~W Mahoney, and N~Benjamin
  Erichson.
\newblock Robustifying state-space models for long sequences via approximate
  diagonalization.
\newblock \emph{arXiv preprint arXiv:2310.01698}, 2023{\natexlab{a}}.

\bibitem[Yu et~al.(2023{\natexlab{b}})Yu, Yang, and Townsend]{yu2022tuning}
Annan Yu, Yunan Yang, and Alex Townsend.
\newblock Tuning frequency bias in neural network training with nonuniform
  data.
\newblock \emph{International Conference on Learning Representations},
  2023{\natexlab{b}}.

\bibitem[Yu et~al.(2025{\natexlab{a}})Yu, Lyu, Lim, Mahoney, and
  Erichson]{yu2024tuning}
Annan Yu, Dongwei Lyu, Soon~Hoe Lim, Michael~W Mahoney, and N~Benjamin
  Erichson.
\newblock Tuning frequency bias of state space models.
\newblock \emph{International Conference on Learning Representations},
  2025{\natexlab{a}}.

\bibitem[Yu et~al.(2025{\natexlab{b}})Yu, Maddix, Han, Zhang, Ansari, Shchur,
  Faloutsos, Wilson, Mahoney, and Wang]{yu2025transformer}
Annan Yu, Danielle~C. Maddix, Boran Han, Xiyuan Zhang, Abdul~Fatir Ansari,
  Oleksandr Shchur, Christos Faloutsos, Andrew~Gordon Wilson, Michael~W.
  Mahoney, and Yuyang Wang.
\newblock Understanding transformers for time series: Rank structure,
  flow-of-ranks, and compressibility.
\newblock \emph{arXiv preprint arXiv:2510.03358}, 2025{\natexlab{b}}.

\bibitem[Zhang \& Gilpin(2025{\natexlab{a}})Zhang and Gilpin]{zhang2024zero}
Yuanzhao Zhang and William Gilpin.
\newblock Zero-shot forecasting of chaotic systems.
\newblock \emph{International Conference on Learning Representations},
  2025{\natexlab{a}}.

\bibitem[Zhang \& Gilpin(2025{\natexlab{b}})Zhang and Gilpin]{zhang2025context}
Yuanzhao Zhang and William Gilpin.
\newblock Context parroting: A simple but tough-to-beat baseline for foundation
  models in scientific machine learning.
\newblock \emph{arXiv preprint arXiv:2505.11349}, 2025{\natexlab{b}}.

\bibitem[Zhao et~al.(2025)Zhao, Ni, Xu, Liu, Jin, and
  Prakash]{zhao2025timerecipe}
Zhiyuan Zhao, Juntong Ni, Shangqing Xu, Haoxin Liu, Wei Jin, and B~Aditya
  Prakash.
\newblock Timerecipe: A time-series forecasting recipe via benchmarking module
  level effectiveness.
\newblock \emph{arXiv preprint arXiv:2506.06482}, 2025.

\bibitem[Zhou et~al.(2021)Zhou, Zhang, Peng, Zhang, Li, Xiong, and
  Zhang]{zhou2021informer}
Haoyi Zhou, Shanghang Zhang, Jieqi Peng, Shuai Zhang, Jianxin Li, Hui Xiong,
  and Wancai Zhang.
\newblock Informer: Beyond efficient transformer for long sequence time-series
  forecasting.
\newblock In \emph{Proceedings of the AAAI Conference on Artificial
  Intelligence}, volume~35, pp.\  11106--11115. AAAI Press, 2021.

\bibitem[Zhou et~al.(2023)Zhou, Niu, Sun, Jin, et~al.]{zhou2023one}
Tian Zhou, Peisong Niu, Liang Sun, Rong Jin, et~al.
\newblock One fits all: Power general time series analysis by pretrained
  {L}{M}.
\newblock \emph{Advances in neural information processing systems},
  36:\penalty0 43322--43355, 2023.

\bibitem[Zhu et~al.(2021)Zhu, Hu, Lou, and Yang]{yang2021implicit}
B.~Zhu, J.~Hu, Y.~Lou, and Y.~Yang.
\newblock Implicit regularization effects of the {Sobolev} norms in image
  processing.
\newblock \emph{arXiv preprint arXiv:2109.06255}, 2021.

\end{thebibliography}
\bibliographystyle{iclr2026_conference}

\clearpage

\appendix

\section{Two Sides of Each Bias}
\label{app:goodandbad}

In the main article, we showed many examples in which a bias may be ``good'' or ``bad'' in certain situations; and we also emphasized that each bias controls a trade-off, and that there are important applications on both ends of the spectrum of a bias. 
In this section, we show concrete synthetic examples illustrating when a bias may be ``good'' or ``bad.'' 
(These examples illustrate learnings from developing Chronos-Bolt to ``improve'' Chronos on well-established benchmarks, and the observation that Chronos-Bolt performs worse than Chronos on the qualitatively-different chaotic nonlinear dynamics data.)
The examples in this appendix are \emph{not} cherry-picked, but they are highly context-specific; for a more generic and systematic analysis, see the main article.
See \Cref{tab:goodbad} for an overview of our discussion.

The way to read~\Cref{tab:goodbad} is as follows: for each bias, we compare two models on the two different ends of the spectrum. Then, we come up with two examples: in the first example, we show that having this bias is good for this particular example; in the second example, we show the opposite. A green background always indicates that a forecast is qualitatively better, while a red background shows otherwise.

\newcolumntype{Y}{>{\centering\arraybackslash}X}

\setlength{\tabcolsep}{6pt}
\renewcommand{\arraystretch}{1.02}
\newlength{\BiasCellH}
\setlength{\BiasCellH}{\dimexpr(\textheight - 10\baselineskip)/12\relax}
\newcommand{\CellFig}[3]{%
  \begin{tikzpicture}
    \node[inner sep=0] (img) {\includegraphics[height=\BiasCellH]{#1}};
    \node[anchor=north west, xshift=4pt, yshift=-4pt,
          fill=white, fill opacity=0.88, text opacity=1,
          rounded corners=2pt, inner sep=2pt, draw=black!10, line width=0.2pt,
          font=\scriptsize\sffamily]
         at (img.north west) {#2};
    \node[anchor=south west, xshift=4pt, yshift=4pt,
          fill=white, fill opacity=0.88, text opacity=1,
          rounded corners=2pt, inner sep=2pt, draw=black!10, line width=0.2pt,
          font=\scriptsize\sffamily\itshape]
         at (img.south west) {#3};
  \end{tikzpicture}
}

\setlength{\heavyrulewidth}{1.4pt}
\setlength{\lightrulewidth}{1pt}

\newcommand{\BiasBlock}[6]{
  \CellFig{./figures/examples/#601.pdf}{\textbf{#1} — GOOD}{\texttt{\textup{#2}}} &
  \CellFig{./figures/examples/#600.pdf}{}{\texttt{\textup{#3}}} \\
  \CellFig{./figures/examples/#610.pdf}{\textbf{#1} — BAD}{} &
  \CellFig{./figures/examples/#611.pdf}{}{} \\
  \addlinespace[0.15em]\midrule
}

\begin{table*}[!htb]
  \captionsetup{font=small}
  \caption{Illustration of when inductive biases ``help'' and when they ``hurt'' performance.}
  \myvspace{-0.75\baselineskip}
  \centering
    \begin{tabularx}{\textwidth}{YY}
      \toprule
      \BiasBlock
        {Frequency}
        {Chronos-Bolt ($k=16$)}
        {Chronos-Bolt ($k=1$)}
        {Model C (with freq. bias)}
        {Model D (alt./reduced freq. bias)}
        {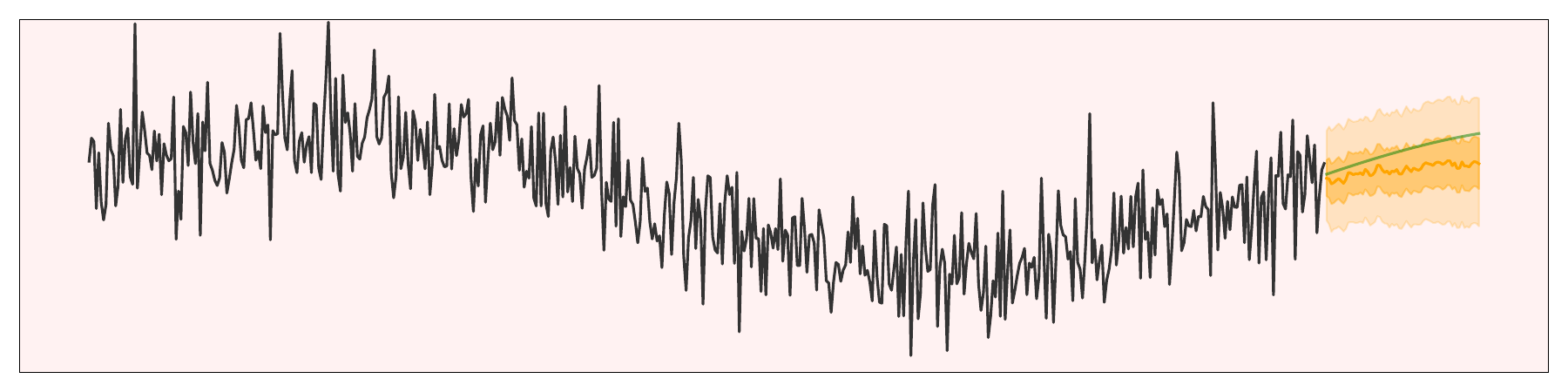}

      \BiasBlock
        {Regression}
        {Chronos}
        {Chronos-Bolt}
        {Model G (regression bias)}
        {Model H (alternative)}
        {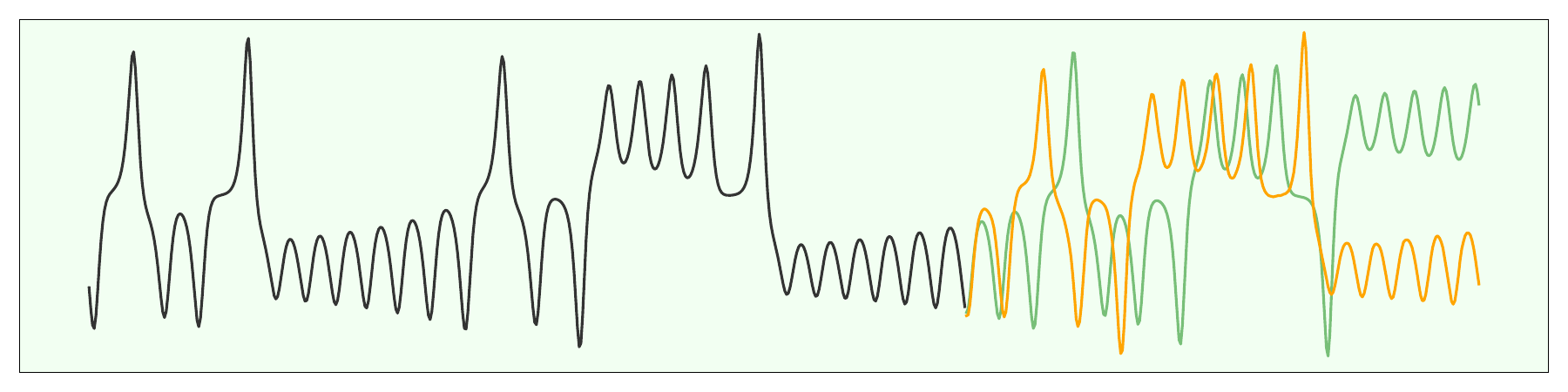}

      \BiasBlock
        {Locality}
        {Chronos}
        {Chronos-Bolt}
        {Model K (strong locality)}
        {Model L (contextual)}
        {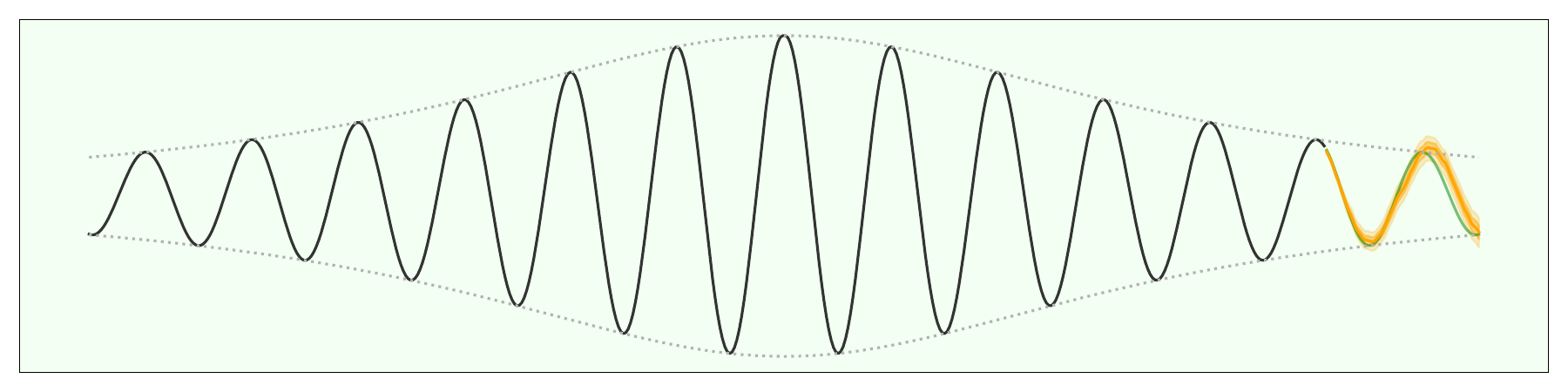}

      \BiasBlock
        {Scale}
        {Chronos-Bolt}
        {Chronos}
        {Model O (scale-aware)}
        {Model P (robust-to-scale)}
        {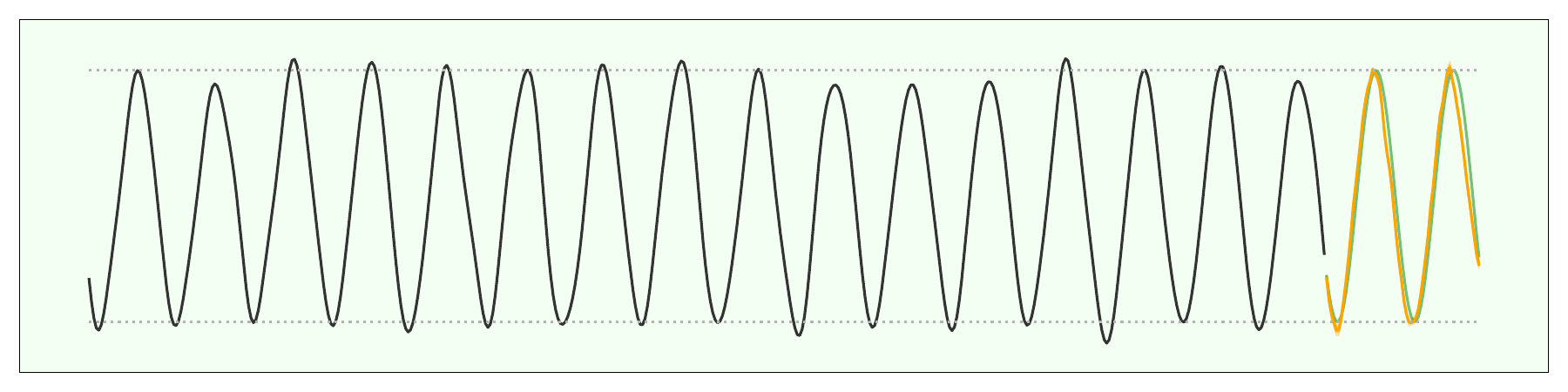}

      \BiasBlock
        {Offset}
        {Chronos-Bolt}
        {Chronos}
        {Model S (offset-invariant)}
        {Model T (re-centered)}
        {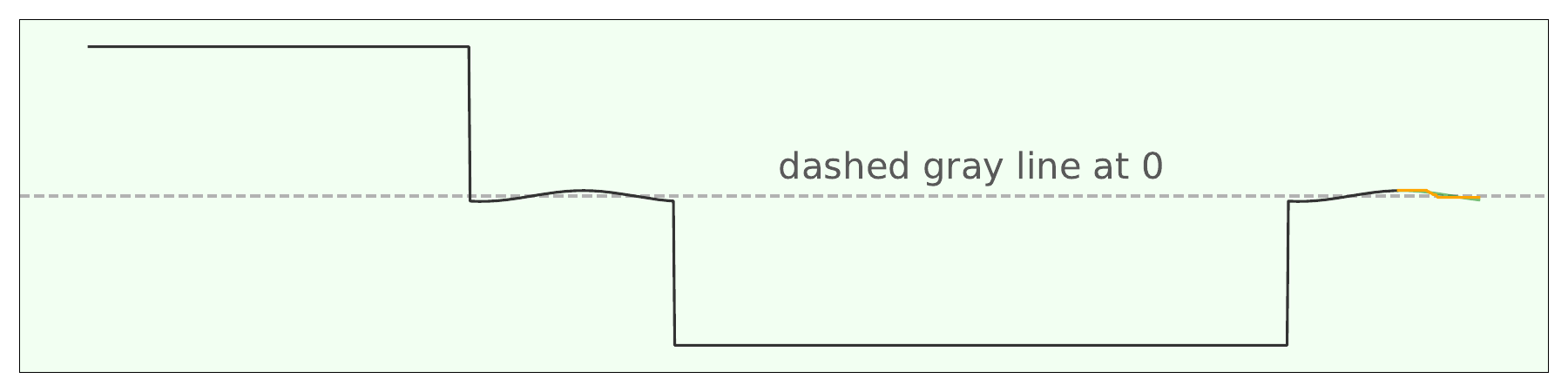}

      \BiasBlock
        {Periodicity}
        {Chronos-Bolt}
        {TimesFM}
        {Model W (periodicity bias)}
        {Model X (adaptive)}
        {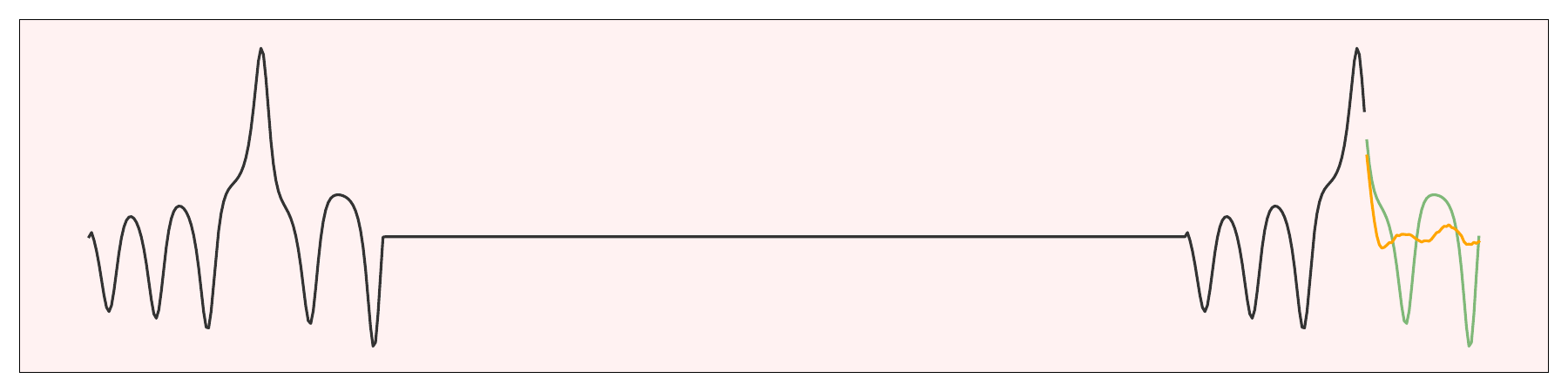}
      \bottomrule
    \end{tabularx}%
    \label{tab:goodbad}
\end{table*}

\subsection{Temporal Bias}

As described in \Cref{sec:frequency}, there are two biases related to the temporal bias: the frequency bias; and the periodicity bias.

\subsubsection{Frequency Bias}

The frequency bias has been known to the deep learning community for almost a decade; and it is often thought of as an implicit regularization. 
Driven by this, we first come up with an example where the context is
\begin{equation}\label{eq.frequencygood}
	x(t) = \sin(t) + \text{noise},
\end{equation}
where the noise is obtained by sampling every entry from the discrete sequence i.i.d.\ according to the standard Gaussian distribution $\mathcal{N}(0,1)$. In this case, $\sin(t)$ is the useful low-frequency information that one wants to learn, while the Gaussian noise is high-frequency content that we do not wish to learn. 
We see that with a larger patch size, and therefore a stronger frequency bias, Chronos-Bolt is more capable of ``denoising.''

On the other hand, the example in~\Cref{fig:frequency} has already shown a case where biasing towards low frequencies is not desirable:
\begin{equation}\label{eq.frequencybad}
	x(t) = \sin(\alpha t) + \sin(\beta t), \qquad \alpha \ll \beta.
\end{equation}
In this case, both the low-frequency information $\sin(\alpha t)$ and the high-frequency information $\sin(\beta t)$ contain useful information, but Chronos-Bolt with a patch size of $16$ only enables us to learn the low-frequency content.

\subsubsection{Periodicity Bias}

To understand the pros and cons of preserving the periodicity bias, we can first consider a signal formed from a chaotic system's data:
\begin{equation}\label{eq.periodicbad}
	x(t) = \mathbbm{1}_{[0,\Delta]}(t) F(t) + \mathbbm{1}_{[T-\Delta,T]}(t) F(t-T+\Delta),
\end{equation}
where $F(\cdot): [0,\Delta] \rightarrow \R$ is a motif taken from a Lorentz oscillator. Since $F$ is repeated twice in the context, and that the task can be solved by simply ``matching'' the repeating motifs, we see that a model that preserves periodicity better, i.e., Chronos-Bolt, performs better than one that destroys periodicity more, i.e., TimesFM.

However, if one focuses too much on the periodic patterns, then one may overlook the important non-periodicities. 
Consider the following context:
\begin{equation}\label{eq.periodicgood}
	x(t) = \mathbbm{1}_{R_1} t\sin(t) + \mathbbm{1}_{R_2} \sin(t),
\end{equation}
where $R_1$ is the disjoint union of all even ones from an evenly divided set of intervals, and $R_2 = [0,\infty) \setminus R_1$ is the complement of it. 
We see in \Cref{tab:goodbad} that Chronos-Bolt focuses heavily on the periodic patterns on $R_2$, resulting in a bad forecast, whereas TimesFM gives a more accurate partition of $R_1$ and $R_2$ in its forecast.

\subsection{Geometric Bias}
As described in~\cref{sec:geometry}, there are three parts to the geometric bias: an angular bias, which concerns the angles between the embeddings; a distance bias, which is due to the distance between the embeddings; and a norm bias, which originates from the homogeneity of the ReLU activation. 
We show interesting examples for each of them.

\subsubsection{Angular Bias}

A model that has an angular bias, e.g., Chronos, tends to tailor the ``parroting'' strategy. 
When there is a matching motif in the earlier context, then the model will parrot what is after it. 
Therefore, if we set the context to be a perfectly periodic signal
\begin{equation}\label{eq.localitygood}
	x(t) = \sin(t),
\end{equation}
then we see that Chronos' forecast is pinpoint, while Chronos-Bolt suffers from certain small errors.

However, parroting can also be dangerous, not only because of the fact that sometimes there is not too much to parrot, but also since it can prevent the model from identifying more subtle patterns. 
Consider the following context:
\begin{equation}\label{eq.localitygood}
	x(t) = \sin(t) S(t),
\end{equation}
where $S(t)$ is a bidirectional scaling factor that vanishes as $t \rightarrow \pm \infty$. 
In this case, Chronos is able to identify the recent context to a previous motif. 
Unfortunately, this ``drags'' Chronos into a caveat: simply parroting the previous context results in a forecast with an increasing trend, whereas the actual trend is clearly decreasing over time. 
In this case, Chronos-Bolt gets a better forecast than Chronos.

\subsubsection{Distance Bias}

Similar to the frequency bias, fine-scale oscillation may sometimes be simply due to random noise. In that sense, the distance bias could be helpful for denoising. Consider the following context:
\begin{equation}\label{eq.scalegood}
	x(t) = \sin(t) + \text{noise},
\end{equation}
where the noise is obtained by sampling every entry from the discrete sequence i.i.d.\ according to the standard Gaussian distribution $\mathcal{N}(0,\alpha)$ for some small $\alpha$, and then passing it through a low-pass filter. The reason why we choose $\alpha$ small is that we want to make the noise small-scale, and the reason why we pass it through the low-pass filter is to decouple the example from the frequency bias. Here, we see that Chronos learns these small-scale noises, while Chronos-Bolt, having a distance bias, only learns the large-scale trend.

Of course, when small-scale properties appear to be important, Chronos performs better than Chronos-Bolt.
One example is shown in~\Cref{fig:scale}, where
\begin{equation}\label{eq.scalebad}
	x(t) = \sin(t) S(t),
\end{equation}
where $S(t)$ is a scaling factor that periodically alternates between large and small values.

\subsubsection{Norm Bias}

Regarding the norm bias, it may not be clear why one should treat signals at different offsets differently. 
Indeed, if we consider the signal labeled ``Offset---BAD'' in~\Cref{tab:goodbad}, we see that Chronos is much better at continuing the signal near zero than Chronos-Bolt.
However, sometimes the small signals in the input domain are really ``physically'' less important than the large signals. 
One important example is with a sparse signal. 
For example, consider the signal labeled ``Offset---GOOD'' in~\Cref{tab:goodbad}. This signal is sparse, consisting of mostly zeros and a few non-zero motifs. Essentially, the most important information is contained in these non-zero motifs. Chronos-Bolt's MLP embedding assigns the zero values small embeddings, and it focuses more on the non-zero parts, resulting in a better prediction than Chronos.

\subsection{Regression-to-the-mean Bias}

As described in \Cref{sec:regression}, it may be intuitive that regression algorithms should regress to the mean, but an example where this bias is not desired is provided by the chaotic systems data from~\citet{zhang2024zero}. 
(Probabilistic forecasting has become an important paradigm in real-world applications, from weather and climate prediction~\citep{gneiting2014probabilistic}, to energy demand forecasting~\citep{hong2016probabilistic}, to financial risk management~\citep{gneiting2007probabilistic}; and evidence suggests that similar results may hold more generally for TSFMs.) 
In particular, we use the Lorentz oscillator to generate our context. We see that since the system is chaotic, neither forecast stays close to the ground-truth, but Chronos' prediction resembles the context much more than Chronos-Bolt's. That is, while from an MSE or MAE perspective, Chronos is not doing so much better than Chronos-Bolt, it performs better on other aspects, such as the frequency loss or the fractal dimension~\citep{zhang2024zero}. More colloquially, without showing the ground-truth, Chronos gives a more credible continuation of the context and Chronos-Bolt.

Clearly, of course, in many other cases, regression-to-the-mean is desirable.
A significant example of this arises when outliers are introduced. 
For example, consider the following context:
\begin{equation}\label{eq.regressionbad}
	x(t) = \sin(t) + \text{Bernoulli}(p) \delta_t,
\end{equation}
where $\delta_t$ is the Dirac-delta and $\text{Bernoulli}(p)$ is i.i.d.\ random variable with a small probability $0 < p < 1$. 
This signal can model a demand curve that follows some seasonal trend but can be excited by several random events. 
In this case, we see that Chronos sometimes predicts these outliers, which we do not want, while Chronos-Bolt generally avoids them.

\clearpage

\section{Related Work}
\label{app:relatedwork}

In this section, we describe several lines of related work.

\subsection{Time series Modeling and Foundation Models}

\begin{figure}[!htb]
    \centering
    \includegraphics[width=1\linewidth]{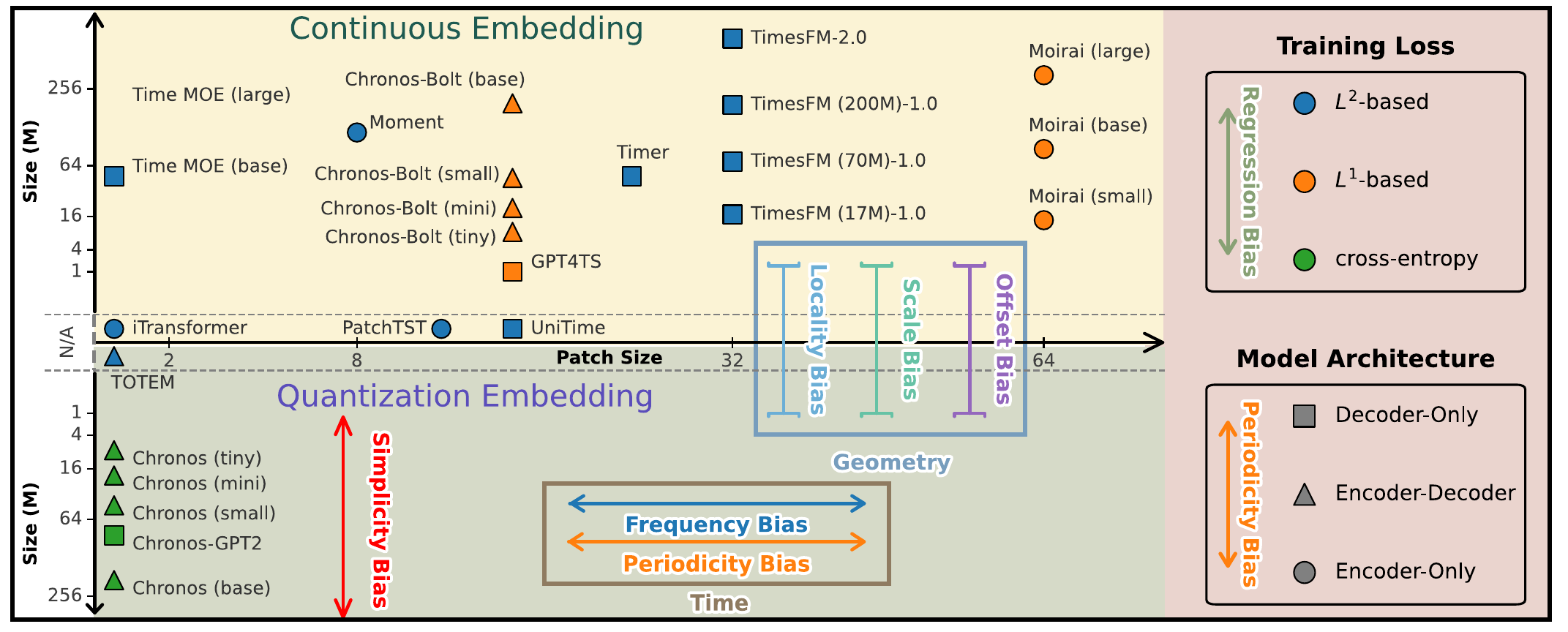}
    \caption{Design choices of different TSFMs and how they affect the implicit biases.}
    \label{fig:designchoices}
\end{figure}

Classical time series forecasting~\citep{hyndman2018forecasting} is grounded in statistical methods such as ARIMA~\citep{makridakis1997arma} and exponential smoothing~\citep{gardner1985exponential}.  
More recently, deep learning approaches, e.g., N-BEATS~\citep{oreshkin2020nbeats}, have become prominent. 
Sequence models such as DeepAR~\citep{salinas2020deepar} popularized probabilistic forecasting at scale, while the Temporal Fusion Transformer~\citep{lim2021temporal} combined attention mechanisms with interpretability for multi-horizon prediction.
Building on these advances, recent work has focused on pretraining general-purpose time series models that transfer across datasets and tasks.  
Notable examples include TimesFM~\citep{das2024decoder}, Chronos~\citep{ansari2024chronos}, Moirai~\citep{woo2024unified}, and Time-MoE~\citep{shi2024time}.  
These models vary in tokenization strategies, training corpora, and zero-shot protocols, but they all share the common goal of developing a single ``foundation model'' that generalizes across domains and horizons. 
Such TSFMs, which we compare along several axes in~\Cref{fig:designchoices}, include Chronos~\citep{ansari2024chronos}, TimesFM~\citep{das2024decoder}, Moirai~\citep{woo2024unified}, PatchTST~\citep{nie2022time}, Time MOE~\citep{shi2024time}, Chronos-Bolt~\citep{ChronosBolt}, iTransformer~\citep{liu2023itransformer}, Moment~\citep{goswami2024moment}, Timer~\citep{liu2024timer}, GPT4TS~\citep{zhou2023one}, UniTime~\citep{liu2024unitime}, and TOTEM~\citep{talukder2024totem}.

\subsection{Design Choices of TSFMs}

Design choices for TSFMs have caught research attention, e.g., with empirical work in~\citet{liang2025TSGym,zhao2025timerecipe}. 
Beyond these benchmark studies, several papers have investigated specific design dimensions of TSFMs.  
For instance, recent work has examined the impact of tokenization strategies~\citep{nie2022time,talukder2024totem}, scaling laws for model size and dataset coverage~\citep{woo2024unified,shi2024time}, and architectural choices such as state-space models versus Transformers~\citep{gu2023mamba, ma2024mamba}.  
Other efforts have focused on evaluation protocols and generalization across domains, highlighting the importance of systematic comparisons to guide the development of more reliable TSFMs.

\subsection{Deep Learning Models for Chaotic Systems Forecasting}

We have used time series from chaotic dynamical systems data as part of our investigation. 
Recent work~\citep{rackauckas2025how} has highlighted the promise and limitations of using deep learning models to learn chaotic systems. 
A major line of work from Gilpin’s group has made fundamental contributions, including providing a benchmark of many state-of-the-art methods---ranging from reservoir computing to TSFMs---showing that large, domain-agnostic models can achieve accurate predictions up to two dozen Lyapunov times into the future~\citep{gilpin2021chaos}.  
Further empirical studies demonstrate that TSFMs like Chronos can perform zero-shot forecasting of chaotic dynamics, preserving the geometric and statistical properties of chaotic attractors even when point forecasts degrade~\citep{zhang2024zero,zhang2025context,lai2025panda}. 
Beyond TSFMs, physics-informed networks leverage known physical constraints to extend predictability limits~\citep{doan2020physics,steger2022pinns}, while knowledge-based hybrid models combining mechanistic and data-driven components reliably forecast complex chaotic behaviors far longer than either approach alone~\citep{pathak2018hybrid}.

\subsection{Frequency Bias}

The frequency bias is related to the spectral bias of neural networks, which was observed and studied in~\citet{rahaman2019spectral,yang2019fine,xu2020frequency}.  
The term ``spectral bias'' comes from the spectral decomposition of neural tangent kernels (NTKs)~\citep{jacot2018neural}, which approximate the training dynamics of overparameterized networks~\citep{arora,suyang,cao2019towards}.  
By analyzing the eigenfunctions of NTKs,~\citet{basri,bietti2019inductive} formally proved the presence of spectral bias in two-layer overparameterized networks with uniform input distributions; and these results were later extended to nonuniform inputs~\citep{basri2020frequency,yu2022tuning}.  
Several approaches have since been proposed to tune the strength of this bias, including Sobolev-norm-based training strategies~\citep{vlassis2021sobolev,yu2022tuning,tsay2021sobolev,son2021sobolev,czarnecki2017sobolev,yang2021implicit,sonsobolev,liu2024mitigating} and frequency-aware initialization methods~\citep{yu2023robustifying,yu2024tuning}.

\clearpage

\section{Details of the Temporal Bias (from \Cref{sec:frequency})}
\label{app:temporal}

In this section, we provide more details on the temporal bias, which was discussed in \Cref{sec:frequency}.
We discuss both the frequency bias (\Cref{app:frequency}) and the periodicity bias (\Cref{app:periodicity}).

\subsection{Details of the Frequency Bias}
\label{app:frequency}

In this subsection, we provide more details on the frequency bias, one of the two temporal biases (the other being the periodicity bias; see \Cref{app:periodicity}), from~\cref{sec:frequency}. 
Our discussion is split into several parts. 
First, we prove~\Cref{thm.frequencybias} and discuss some of its consequences (\Cref{app:proof_of_freq_bias_thm}). 
Then, we provide more results on the attention matrices in a Transformer with patching to corroborate the frequency bias (\Cref{app:frequency-visualization-attention}). 
Finally, we show the existence and non-existence of the frequency bias in two more interesting scenarios (\Cref{app:frequency-additional-evidence}).

\subsubsection{Proof and Discussion of~\Cref{thm.frequencybias}}
\label{app:proof_of_freq_bias_thm}

\subsubsection*{The Proof}

The proof of~\Cref{thm.frequencybias} is an application of concentration inequalities and matrix norm bounds once we realize that the inverse Fourier transform is an orthogonal transformation, mapping distinct Fourier modes to orthogonal subspaces in the patch space $\R^k$.

\begin{proof}[Proof of~\Cref{thm.frequencybias}]
    We prove the two statements separately.

    \noindent \textbf{Proof of the First Statement.} If the supports of $\hat{\mathbf{v}}_1, \ldots, \hat{\mathbf{v}}_n$ all belong to a set $B_j$ with a cardinality no more than $\omega$, then the vectors $\hat{\mathbf{v}}_1, \ldots, \hat{\mathbf{v}}_n$ are contained in a subspace of dimension $\leq \omega$. Since the discrete Fourier transform is an orthogonal operator, the vectors ${\mathbf{v}}_1, \ldots, {\mathbf{v}}_n$ are also contained in a subspace of dimension $\leq \omega$. Therefore, we have that $\text{rank}(\mathbf{V}) \leq \omega$ and that
    \[
        \|\mathbf{V}\|_F \leq \sqrt{\sigma_1(\mathbf{V})^2 + \cdots  + \sigma_n(\mathbf{V})^2} \leq \sqrt{\omega \,\sigma_1(\mathbf{V})^2} = \sqrt{\omega} \,\sigma_1(\mathbf{V}).
    \]
    Since the $\text{ReLU}$ activation function cannot increase the norm of an input matrix, we therefore have
    \begin{align*}
        \|\Phi(\mathbf{V})\|_F &\leq \|\mathbf{W}_2 \,\text{ReLU}(\mathbf{W}_1 \mathbf{V})\|_F + \|\mathbf{b} \,\boldsymbol{1}_{n}^\top\|_F \\
        &= \mathcal{O}(1) \|\mathbf{W}_2 \,\text{ReLU}(\mathbf{W}_1 \mathbf{V})\|_F \leq \mathcal{O}(1) \|\mathbf{W}_2\|_2 \|\text{ReLU}(\mathbf{W}_1 \mathbf{V})\|_F  \\
        &\leq \mathcal{O}(1)\|\mathbf{W}_2\|_2 \|\mathbf{W}_1 \mathbf{V}\|_F \leq \mathcal{O}(1) \|\mathbf{W}_2\|_2 \|\mathbf{W}_1\|_2 \|\mathbf{V}\|_F \leq \mathcal{O}(\sqrt{\omega}) \|\mathbf{W}_2\|_2 \|\mathbf{W}_1\|_2 \|\mathbf{V}\|_2.
    \end{align*}
    By our assumption, we have that $\sigma_1(\Phi(\mathbf{V})) \geq \Omega(1) \|\mathbf{W}_2\|_2 \|\mathbf{W}_1\|_2 \|\mathbf{V}\|_2$, and the statement about the stable rank follows immediately. Thus, we have
    \begin{equation*}
        \begin{aligned}
            \text{rank}_\varepsilon(\Phi(\mathbf{V})) &\leq \frac{\sum_{j=1}^n \sigma_j(\Phi(\mathbf{V}))^2}{ \varepsilon^2 \sigma_1(\Phi(\mathbf{V}))^2} \leq \frac{\|\Phi(\mathbf{V})\|^2_F}{ \varepsilon^2 \sigma_1(\Phi(\mathbf{V}))^2} = \left(\frac{\|\Phi(\mathbf{V})\|_F}{ \varepsilon \sigma_1(\Phi(\mathbf{V}))}\right)^2 \\
            &\leq \left(\frac{\sqrt{\omega} \|\mathbf{W}_2\|_2 \|\mathbf{W}_1\|_2 \|\mathbf{V}\|_2}{ \varepsilon \Omega(1) \|\mathbf{W}_2\|_2 \|\mathbf{W}_1\|_2 \|\mathbf{V}\|_2}\right)^2 = \mathcal{O}(1) \varepsilon^{-2} \omega.
        \end{aligned}
    \end{equation*}

    \noindent \textbf{Proof of the Second Statement.} We will break the proof into three steps. In the first step, we lower-bound the Frobenius norm of $\Phi(\mathbf{U})$. In the second step, we upper-bound the spectral norm of $\Phi(\mathbf{U})$. This gives us an estimate of the number of large singular values, which we formally prove as the last step.

    \begin{itemize}[leftmargin=*]
        \item \textbf{Step I:} Since the ReLU activation function is homogeneous on $[0,\infty)$, we may assume, without loss of generality, that $\alpha = \beta = 1$. Since the vectors $\hat{\mathbf{u}}_1, \ldots, \hat{\mathbf{u}}_n$ are supported on mutually disjoint sets of indices, they are orthogonal to each other. Moreover, since each of them is a unit vector, they are orthonormal. Since the inverse discrete Fourier transform is an orthogonal operator, we have that the matrix $\mathbf{U}$ has orthonormal columns. Therefore, the matrix $\mathbf{W}_1 \mathbf{U}$ is still a random Gaussian matrix whose entries are i.i.d.\ distributed like $\mathcal{N}(0,1)$. That is, we clearly have that
        \begin{equation*}
            \mathbb{E}\left[\|\text{ReLU}(\mathbf{W}_1 \mathbf{U})\|^2_F\right] = \frac{1}{2} mn,
        \end{equation*}
        where the $1/2$ comes from the fact that $\text{ReLU}$ fires each neuron with a probability of $1/2$. Moreover, by estimating the lower tail of the $\chi^2$ distribution, we know that for some $m = \Omega(\log(1/\delta))$, with probability no less than $1 - \delta / 5$, we have that
        \begin{equation}\label{eq.postrelubound}
            \|\text{ReLU}(\mathbf{W}_1 \mathbf{U})\|^2_F \geq \frac{3}{7} mn.
        \end{equation}
        From now on, we instantiate $\mathbf{W}_1$ and assume that~\cref{eq.postrelubound} is achieved. Let $\mathbf{Z} = \text{ReLU}(\mathbf{W}_1 \mathbf{U})$, $N_1 = \|\mathbf{Z}\|_F$, and let $\mathbf{Y} = \mathbf{W}_2 \mathbf{Z}$. Let $\mathbf{w}_i^\top$ be the $i$th row of the matrix $\mathbf{W}_2$. We have
        \begin{equation*}
        \begin{aligned}
            \mathbb{E}[\|\mathbf{Y}\|_F^2] &= \mathbb{E}[\|\mathbf{W}_2 \mathbf{Z}\|_F^2] = \mathbb{E}\left[\sum_{i=1}^d \|\mathbf{w}_i^\top \mathbf{Z}\|_2^2\right] = \sum_{i=1}^d \mathbb{E}\left[\|\mathbf{w}_i^\top \mathbf{Z}\|_2^2\right] \\
            &= d \mathbb{E}\left[\|\mathbf{w}_1^\top \mathbf{Z}\|_2^2\right] = d \sum_{j=1}^n \mathbb{E}\left[(\mathbf{w}_1^\top \mathbf{z}_j)^2\right].
        \end{aligned}
        \end{equation*}
        However, since $\mathbf{w}_1^\top \sim \mathcal{N}(\boldsymbol{0}, \mathbf{I}_k)$, we have 
        \[
            \mathbb{E}\left[(\mathbf{w}_1^\top \mathbf{z}_j)^2\right] = \|\mathbf{z}_j\|_2^2.
        \]
        This gives us 
        \[
            \mathbb{E}[\|\mathbf{Y}\|_F^2] = d \sum_{j=1}^n \mathbb{E}\left[(\mathbf{w}_1^\top \mathbf{z}_j)^2\right] = d \sum_{j=1}^n \|\mathbf{z}_j\|_2^2 = d \|\mathbf{Z}\|_F^2.
        \]
        By another Chernoff bound, we have that for $d = \Omega(\log(n/\delta))$, with probability no less than $1-\delta/5$, that
        \begin{equation}\label{eq.outputbound}
            \|\mathbf{Y}\|_F^2 \geq \Omega(1) d\|\mathbf{Z}\|_F^2 \geq \frac{2}{5} dmn.
        \end{equation}
        Clearly, for $d = \Omega(\log(n/\delta))$, with probability no less than $1 - \delta / 5$, we have
        \begin{equation}\label{eq.biadbound}
            \|\mathbf{W}_2 \boldsymbol{1}_m\|_2^2 \leq \frac{6}{5} dm \quad \xRightarrow \quad\quad \|\mathbf{W}_2 \boldsymbol{1}_{m \times n} / \sqrt{2\pi}\|_F^2 \leq \frac{6}{5 \times 2\pi} dmn.
        \end{equation}
        By a union bound, with probability no less than $1 - 3\delta/5$, we have that~\cref{eq.postrelubound},~(\ref{eq.outputbound}), and~(\ref{eq.biadbound}) hold. When these hold, the triangle inequality of the Frobenius norm gives us
        \begin{equation}\label{eq.finaloutputbound}
            \|\Phi(\mathbf{U})\|_F^2 \geq \|\mathbf{Y}\|_F^2 - \|\mathbf{W}_2 \boldsymbol{1}_{m \times n} / \sqrt{2\pi}\|_F^2 \geq \left(\frac{2}{5} - \frac{3}{5\pi}\right)dmn > \frac{1}{5} dmn.
        \end{equation}

        \item \textbf{Step II:} We can write the output of the neural network as
        \[
            \Phi(\mathbf{U}) = \mathbf{W}_2 \text{ReLU}(\mathbf{W}_1 \mathbf{U}) - \frac{1}{\sqrt{2\pi}} \mathbf{W}_2 \boldsymbol{1}_{m \times n} = \mathbf{W}_2 \left(\underbrace{\text{ReLU}(\mathbf{W}_1 \mathbf{U}) - \frac{1}{\sqrt{2\pi}} \boldsymbol{1}_{m \times n}}_{\mathbf{R}}\right).
        \]
        First, since $\mathbf{W}_2$ is a random Gaussian matrix whose entries are i.i.d.\ $\mathcal{N}(0,1)$, for $d = \Omega(\log(n/\delta))$ and $m = \Omega(\log(1/\delta))$ defined above and with probability no less than $\delta/5$, we have
        \begin{equation}\label{eq.spectralW2}
            \|\mathbf{W}_2\|_2 = \mathcal{O}(1) (\sqrt{d} + \sqrt{m}).
        \end{equation}
        Now, consider the matrix $\mathbf{R}$. Since $\mathbf{U}$ has orthonormal columns, we know that the matrix $\mathbf{W}_1 \mathbf{U}$ still has i.i.d.\ entries, and so does $\mathbf{R}$. Moreover, each entry in $\mathbf{R}$ clearly has a sub-Gaussian distribution. Let $R$ be an entry of $\mathbf{R}$, then we have
        \[
            \mathbb{E}[R] = \mathbb{E}\left[\text{ReLU}(Z) - \frac{1}{\sqrt{2\pi}}\right] = \mathbb{E}\left[\text{ReLU}(Z)\right] - \frac{1}{\sqrt{2\pi}} = 0,
        \]
        where $Z \sim \mathcal{N}(0,1)$ and it is well-known that $\mathbb{E}\left[\text{ReLU}(Z)\right] = 1/\sqrt{2\pi}$. Since we assumed that $m = \Omega(\log(1/\delta))$, with probability no less than $1 - \delta / 5$, we have
        \begin{equation}\label{eq.spectralR}
            \|\mathbf{R}\|_2 = \mathcal{O}(1) (\sqrt{m} + \sqrt{n}).
        \end{equation}
        By a union bound,~\cref{eq.spectralW2} and~(\ref{eq.spectralR}) hold with a probability no less than $1 - 2\delta/5$. When these two equations hold, we have by the sub-multiplicativity of the spectral norm that
        \begin{equation}\label{eq.firstsingval}
            \|\Phi(\mathbf{U})\|_2 = \|\mathbf{W}_2 \mathbf{R}\|_2 \leq \|\mathbf{W}_2\|_2 \|\mathbf{R}\|_2 = \mathcal{O}(1) (\sqrt{dm} + m) = \mathcal{O}(1) \sqrt{dm},
        \end{equation}
        where the second last step comes from the fact that $n < m$ and the last step follows from the fact that $d/m = \Theta(1)$.

        \item \textbf{Step III:} By a final union bound, we can assume that~\cref{eq.finaloutputbound} and~(\ref{eq.firstsingval}) hold with a probability of no less than $1 - \delta$. Let $\sigma_1, \ldots, \sigma_n$ be the singular values of $\Phi(\mathbf{U})$. Then, we have that
        \[
            \frac{\sigma_1^2 + \cdots + \sigma_n^2}{\sigma_1^2} = \frac{\|\Phi(\mathbf{U})\|^2_F}{\|\Phi(\mathbf{U})\|^2_2} = \Omega(1) \frac{dmn}{dm} = \Omega(n).
        \]
        Since the constant in the $\Omega$-notation is universal, there must exist a universal $\varepsilon > 0$ such that at least $\Omega(n)$ of the relative singular values $\sigma_j / \sigma_1 = \sigma_j^2 / \sigma_1^2$ are $> \varepsilon$. The proof is complete.   
    \end{itemize}
\end{proof}

\subsubsection*{Interpretation and Discussion of Technical Details}
\label{sec:technicalfreqbias}

\begin{figure}[!htb]
    \centering
    \includegraphics[width=0.95\linewidth]{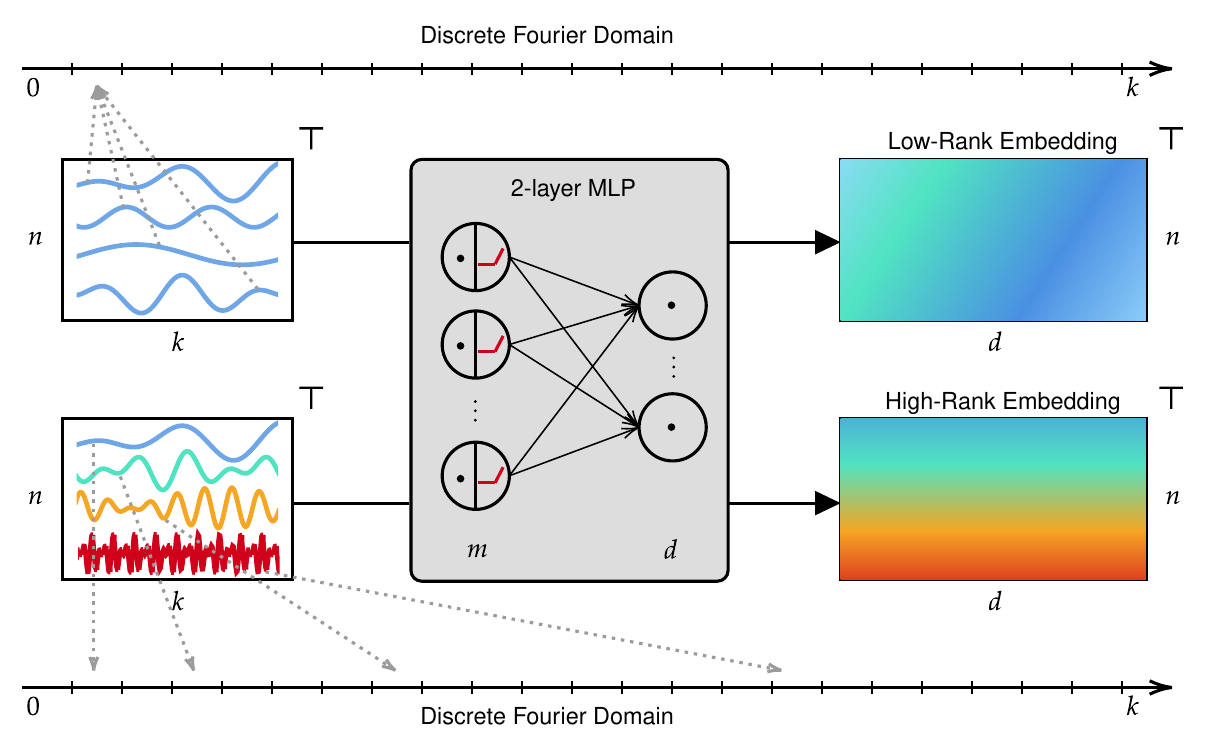}
    \caption{An illustration of~\Cref{thm.frequencybias}: if we sample a number of patches from the same narrow frequency band, then they are embedded into a small subspace; while if we sample the patches with different frequency contents, then they are embedded into different subspaces of the hidden space.}
    \label{fig:frequencybiasthm}
\end{figure}

A pictorial description of~\Cref{thm.frequencybias} can be found in~\Cref{fig:frequencybiasthm}.

As a next step, notice that our theorem relies on a list of crucial assumptions. We anticipate some questions on them, and now we discuss why certain assumptions are made.

\begin{itemize}
    \item \textit{Why do we assume that the support of a vector $\mathbf{u}$ or $\mathbf{v}$ is strictly in a band?}

    We make this assumption for the simplicity of the proof and the cleanliness of the final statement. In practice, when $\omega$ is small, one hardly encounters a patch whose support is strictly limited within a narrow band; instead, it is often the case that a patch has its Fourier coefficients dense in a band and are supported with small values elsewhere. To prove a result in this case, one can view these patches as small perturbations of strictly band-limited patches, and a statement will follow using a Weyl-type perturbation analysis.

    \item \textit{In the first statement, why do we assume that $\|\Phi(\mathbf{V})\| = \Omega(1) \|\mathbf{W}_2\|_2 \|\mathbf{W}_1\|_2 \|\mathbf{V}\|_2$?}

    If we assume no ReLU activation is used, then this condition simply reduces to the general tightness of the sub-multiplicativity of the spectral norm. However, introducing the ReLU activation function usually only increases the spectral norm of the output. Hence, the lower bound is natural to assume.

    \item \textit{In the first statement, why do we assume that $\|\mathbf{b} \boldsymbol{1}_n^\top\|_F = \mathcal{O}(1) \|\mathbf{W}_2 \text{ReLU}(\mathbf{W}_1 \mathbf{V})\|_F$?}

    This assumption simply says that when computing the output of a two-layer ReLU-activated MLP, we are not asymptotically dominated by the bias term, which is a natural one to make.

    \item \textit{In the second statement, why do we assume that $\mathbf{b} = -\mathbf{W}_2 \boldsymbol{1}_m / \sqrt{2\pi}$?}

    This is perhaps the most arguable assumption that we have made. It is a technical assumption that we will further empirically justify in our numerical experiments. At a high level, if we assume that $\mathbf{W}_1$ and $\mathbf{W}_2$ are random Gaussian matrices, the ReLU's output $\text{ReLU}(\mathbf{W}_1 \mathbf{V})$ does not have a mean of $0$. Since it does not have a mean of zero, in expectation, it will introduce a rank-$1$ perturbation in the output of the MLP. The bias term is there to recenter the intermediate output $\text{ReLU}(\mathbf{W}_1 \mathbf{V})$ to avoid the rank-$1$ component. Without the bias term, the output will be the sum of a large rank-$1$ matrix and a full-rank matrix, making our statement about the stable rank invalid, but the statement about the $\varepsilon$-rank remains true.
\end{itemize}

\subsubsection*{Numerical Experiment}
\label{sec:numexpfreqbias}

\begin{figure}[H]
    \centering
    \includegraphics[width=0.9\linewidth]{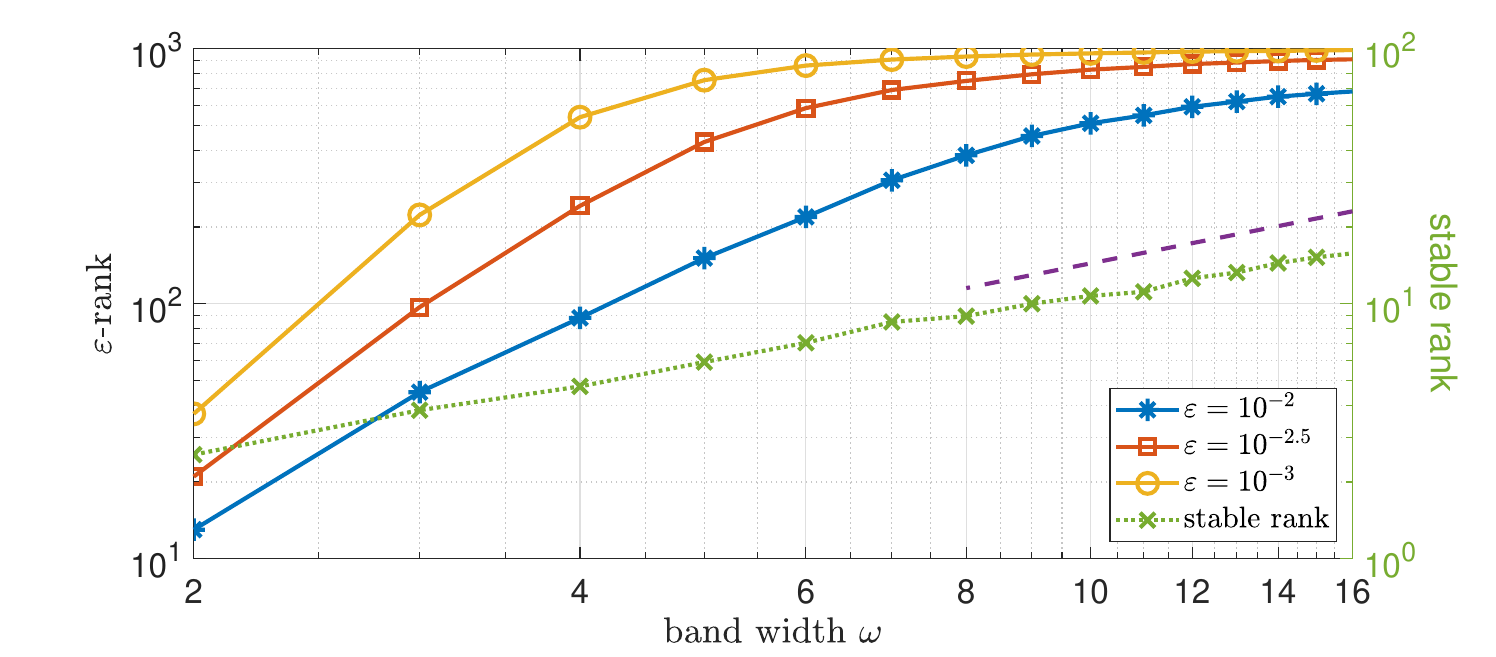}
    \caption{The $\varepsilon$-ranks and stable ranks of the output matrix of a randomly sampled MLP layer, given an input matrix sampled from a single frequency band whose width is limited by $\omega$. As $\omega$ increases, the ranks of the output matrix grow. The reference line has a slope of $1$ in this log-log plot.}
    \label{fig:freqbiasrank}
\end{figure}

We conclude our discussion of~\Cref{thm.frequencybias} by presenting three numerical experiments to corroborate the theorem statement and elaborate on some technical assumptions made. Our first experiment validates that the bandwidth $\omega$ limits the rank of the input embedding. For this experiment, we set $d = 1024$, $m = 4096$, $n = 1024$, and $k = 64$. Our MLP is obtained by randomly sampling Gaussian matrices $\mathbf{W}_1$ and $\mathbf{W}_2$ and defining the bias term to be the centering factor $\mathbf{W}_2 \boldsymbol{1}_m$ as explained earlier. For a fixed bandwidth $\omega$, we sample $n$ patches whose supports are contained in a fixed band of width $\omega$. We then simulate the two-layer MLP to compute the output matrix.

In~\Cref{fig:freqbiasrank}, we see that as $\omega$ increases, the stable rank of the output matrix follows the linear growth, which corroborates the first statement in~\Cref{thm.frequencybias}. The $\varepsilon$-ranks are slightly more complicated because they are upper bounded by $\min(d,n)$ and saturate easily. In practice, we see that the growth of the $\varepsilon$-ranks is initially faster than linear and eventually slows down.

In the second experiment, we compare the two statements made in~\Cref{thm.frequencybias}. To this end, we design an experiment, where we fix the bandwidth to $\omega = 2$. We still use $d = 1024$, $m = 4096$, and $k = 64$. However, we change $n$, the number of patches sampled. For each $n$, we sample these patches in two ways:
\begin{itemize}
    \item \textbf{A sampling that resembles the first statement:} we sample the $n$ patches from a fixed single band that has a width of $\omega$.
    \item \textbf{A sampling that resembles the second statement:} we sample the $n$ patches from $n$ mutually disjoint bands in the Fourier domain. Note that this requires $\omega \cdot n \geq k$, which is the reason why we choose a small $\omega$ for this experiment.
\end{itemize}
We then simulate the two-layer MLP sampled in the same way as explained above.

\begin{figure}[H]
    \centering
    \begin{minipage}{0.47\textwidth}
        \includegraphics[width=1\linewidth]{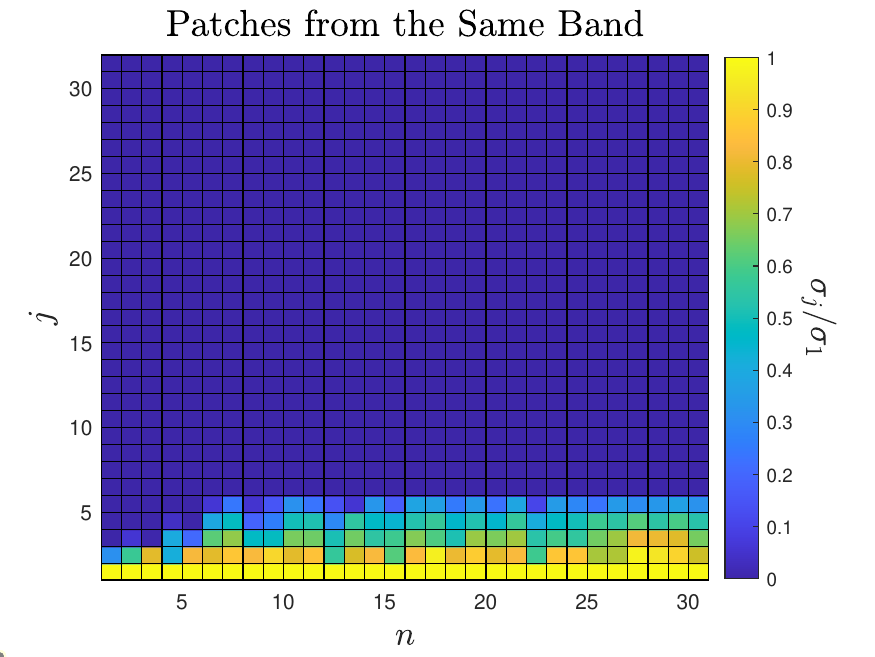}
    \end{minipage}
    \hfill
    \begin{minipage}{0.47\textwidth}
        \includegraphics[width=1\linewidth]{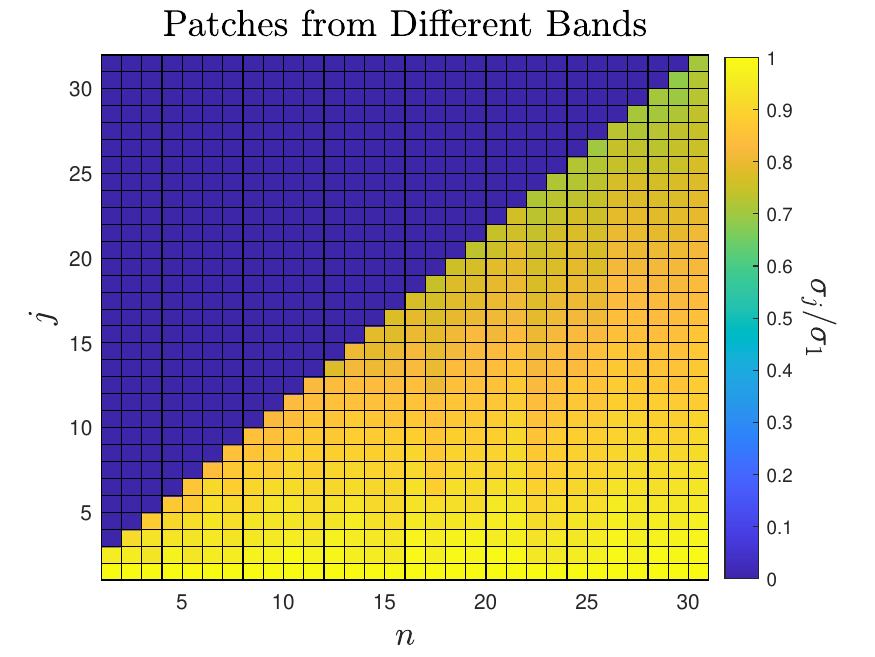}
    \end{minipage}
    \caption{The relative singular values of the output matrix. On the left, we sample all $n$ patches in the input matrix from the same $\omega$-band in the Fourier domain. On the right, we sample $n$ patches from the $n$ mutually disjoint $\omega$-bands in the Fourier domain.}
    \label{fig:diffranksfreq}
\end{figure}

From~\Cref{fig:diffranksfreq}, we see that if we sample the $n$ patches from the same band, then there is a limited number of large singular values, no matter how large $n$ is. This corroborates the first statement. On the other hand, if we sample the $n$ patches from $n$ different Fourier bands, and therefore inject different frequencies into each of them, then all singular values are fairly large --- in fact, none of the singular values are below $\sigma_1 / 2$ --- indicating that the output matrix is numerically full-rank. This corroborates the second statement in~\Cref{thm.frequencybias}.

The last experiment we show concerns the necessity of the bias term $\mathbf{b}$ in the second statement of~\Cref{thm.frequencybias}. As explained earlier, this term is used to centralize the intermediate output. In this experiment, we do not use the bias term, and we repeat the experiment shown in~\Cref{fig:diffranksfreq}. We see that compared to the case with the bias term, the output matrices are essentially the sum of a large rank-$1$ matrix and a smaller full-rank matrix. That is, for a reasonably small $n$ and a reasonably large $\varepsilon$, the $\varepsilon$-rank of the output matrix is still large and growing with $n$.

\begin{figure}[H]
    \centering
    \begin{minipage}{0.47\textwidth}
        \includegraphics[width=1\linewidth]{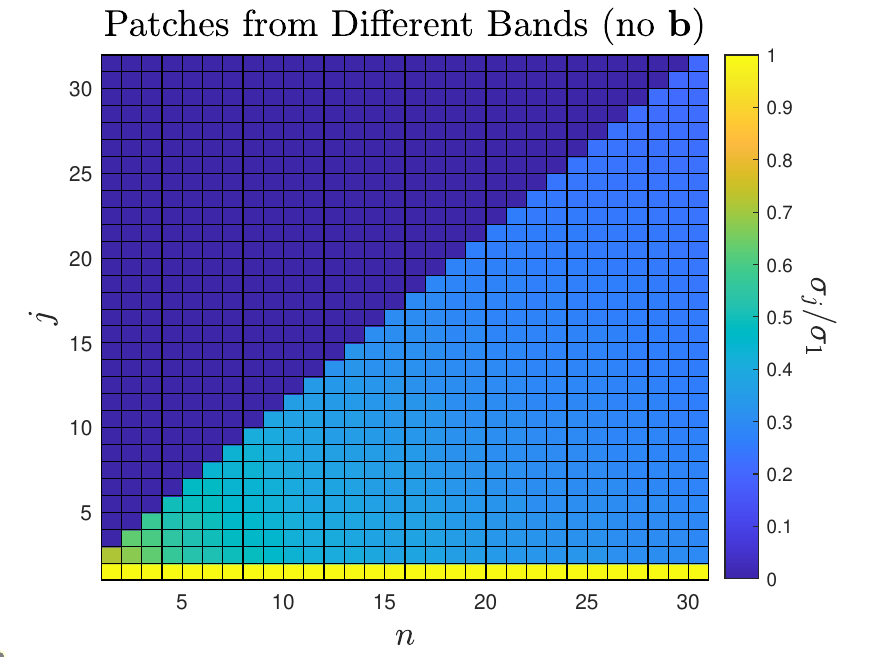}
    \end{minipage}
    \hfill
    \begin{minipage}{0.47\textwidth}
        \includegraphics[width=1\linewidth]{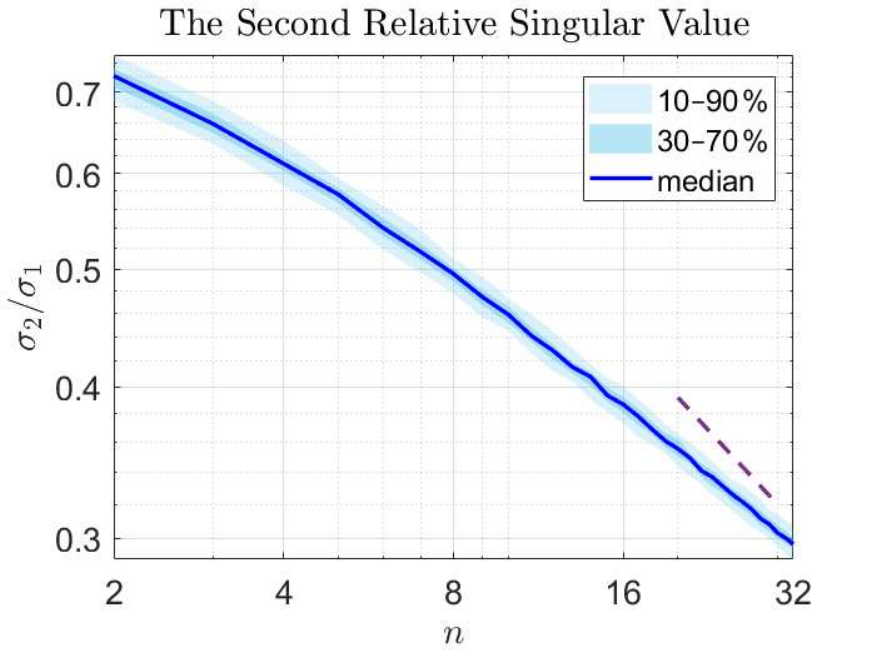}
    \end{minipage}
    \caption{On the left, we reproduce the experiment in~\Cref{fig:diffranksfreq}, but without the bias term $\mathbf{b}$ in the neural network. We see that removing the bias term makes the full-rank output a sum of a large rank-one matrix and a small full-rank matrix. On the right, we show the gap between the rank-one term and the full-rank term as the number of sampled patches $n$ increases, by measuring the second relative singular value $\sigma_2/\sigma_1$ of the output matrix. The reference line has a slope of $-1/2$.}
    \label{fig:freqnob}
\end{figure}

As $n$ increases, one needs to set $\varepsilon$ smaller to still have most of the relative singular values of the output larger than $\varepsilon$. In fact, theoretically, it is relatively easy to show that we need $\varepsilon \sim 1/\sqrt{n}$, so that the $\varepsilon$-rank of the output matrix, without the bias term $\boldsymbol{b}$, is proportional to $n$ as it increases. This is still much larger than the output of patches from the same band. In~\Cref{fig:freqnob}, we also repeat this experiment for $100$ randomly sampled neural networks without the bias term, and we show the second relative singular value $\sigma_2 / \sigma_1$ of the output matrix, which measures the gap between the large rank-one matrix and the small full-rank matrix. We see that this gap seems to grow slightly slower than $\sqrt{n}$, which is suggested by the theory.

\subsubsection{Visualization of Attention Scores}
\label{app:frequency-visualization-attention}

\begin{figure}[H]
    \centering
    \includegraphics[width=0.9\linewidth, trim = 5cm 6cm 5cm 6cm, clip]{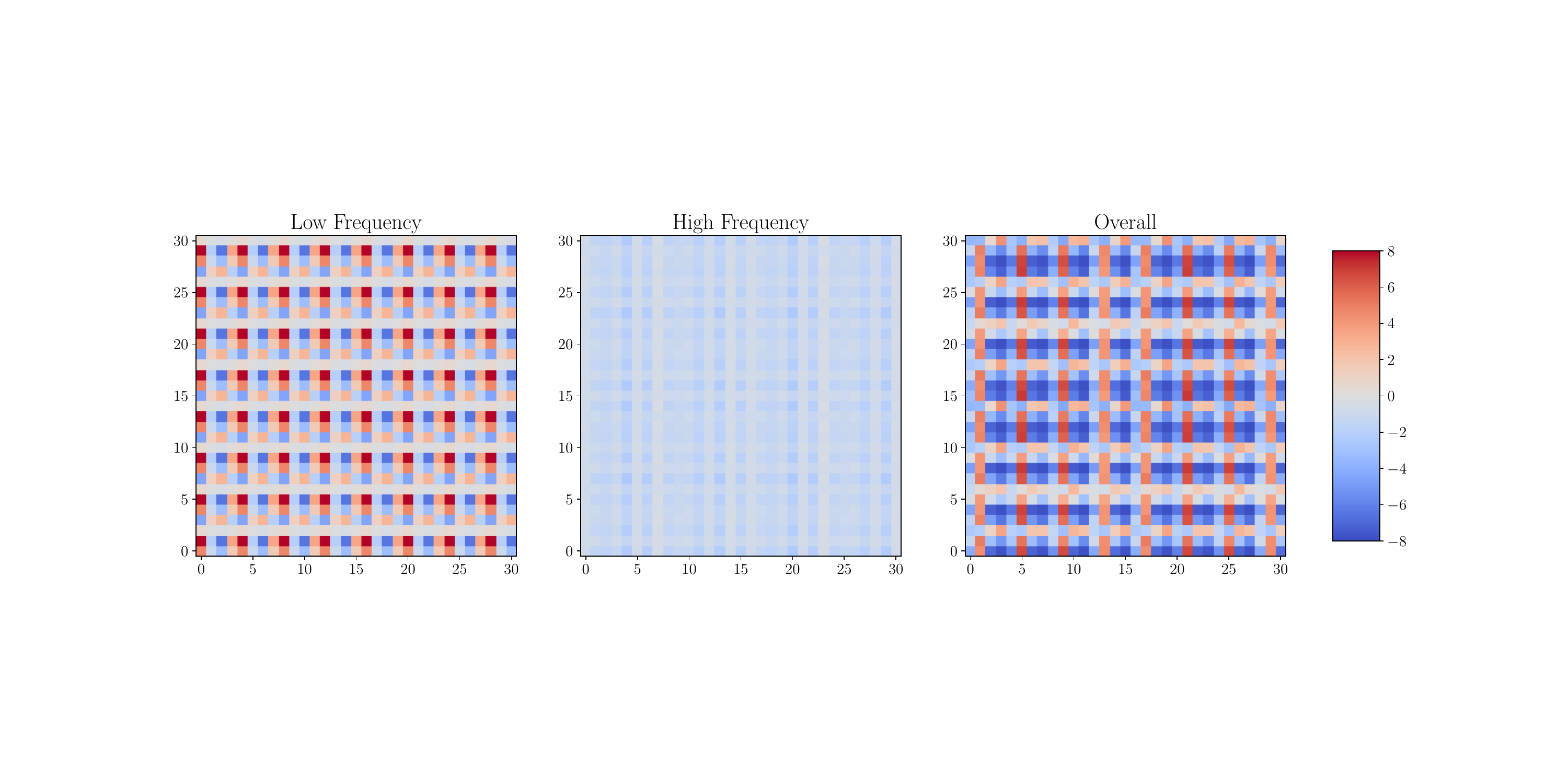}
    \caption{The attention scores before softmax, i.e., $\mathbf{Q}^\top \mathbf{K} \in \R^{L \times L}$, where $L$ is the length of the input sequence (i.e., the number of patches). We see that overall, if the patches are low-frequency, then the attention scores have much larger magnitudes than if the patches are high-frequency. Therefore, when a signal contains both low-frequency content and high-frequency one, the attention scores are dominated by the (periodicity of) the low-frequency information.}
    \label{fig:freqhighlowatt}
\end{figure}

Recall that, in~\Cref{fig:frequency}, we saw that the low- and high-frequency patches are embedded in two nearly orthogonal subspaces of the hidden space. 
It is also the case that the attention scores corresponding to the low-frequency patches are much larger than those corresponding to the high-frequency ones. 
In~\Cref{fig:freqhighlowatt}, we show the magnitudes of the averaged (across all heads) attention scores in the first layer of the Chronos-Bolt's encoder. 
The first one shows the attention scores when we give $\sin(8t)$ as the input, and the second one shows the attention scores when we input $\sin(117t)$. 
We can see that since the attention kernel $\mathbf{W}_Q^\top \mathbf{W}_K$ is much better aligned with the low-frequency patches, the scores in the first panel have a much higher order of magnitude than the scores in the second panel. 
Therefore, when we superpose the two Fourier modes to form the composite input, the attention scores mainly follow the periodicity of $8$ instead of $117$.

\subsubsection{Additional Evidence: Superposing Two Frequencies without Aliasing}
\label{app:frequency-additional-evidence}

Recall that, in~\Cref{fig:frequency}, we saw an example where our context is sampled from the signal
\[
    f(t) = \sin(8t) + \sin(117 t).
\]
In that case, we know that if we use a patch size of $k = 16$, then the low- and high-frequency information is stored separately into different subspaces. Moreover, since the attention scores are dominated by the low-frequency information, the high-frequency patterns are ruined during attention, resulting in a poor learning capability of the high-frequency information.

\begin{figure}[H]
    \centering
    \begin{minipage}{0.47\textwidth}
        \includegraphics[width=1\linewidth, trim = 0cm 0cm 0cm 1cm, clip]{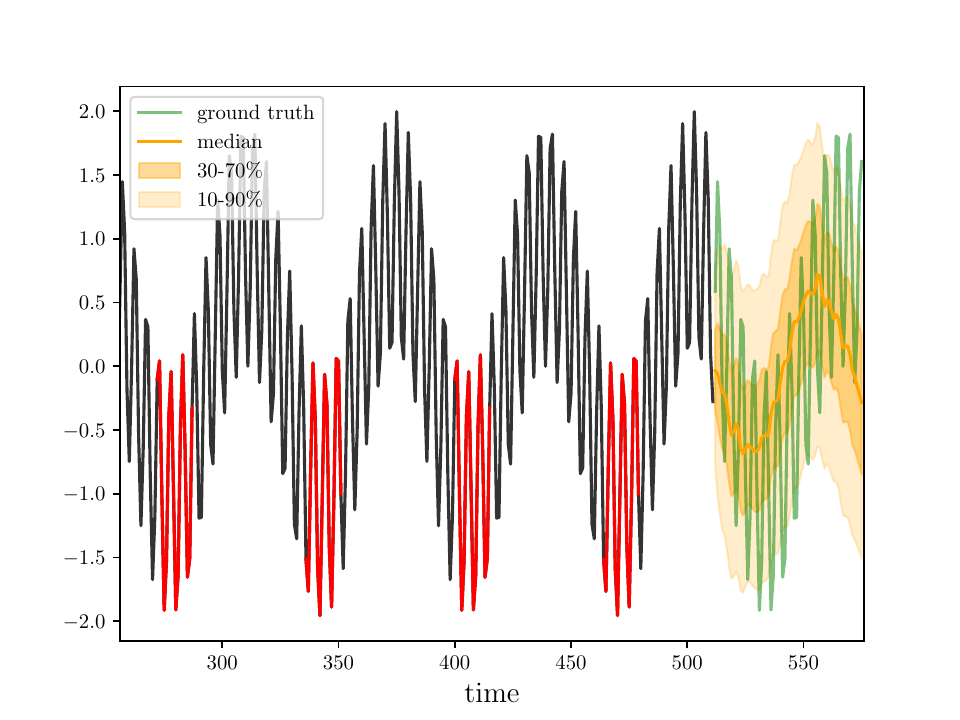}
    \end{minipage}
    \hfill
    \begin{minipage}{0.47\textwidth}
        \includegraphics[width=1\linewidth, trim = 0cm 0cm 0cm 1cm, clip]{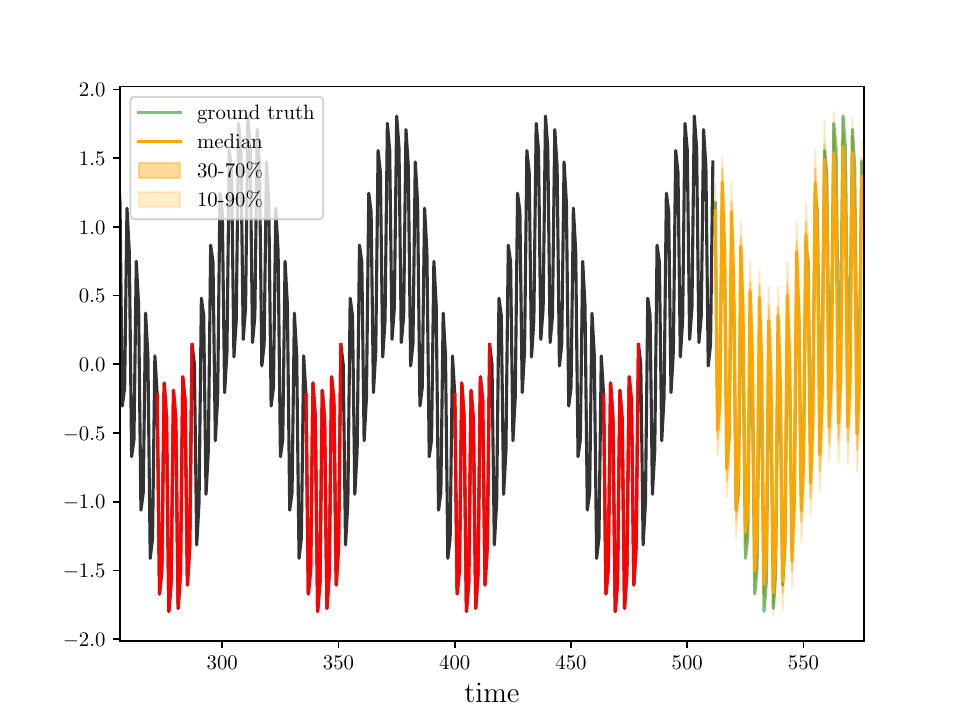}
    \end{minipage}

    \myvspace{-0.2cm}
    
    \caption{On the left, we show the forecast of a Chronos-Bolt model (with patch size $k=16$) where the periodicity of high-frequency information in the context does not align with that of the low-frequency information. On the right, we show the forecast when the periodicities of the high- and low-frequency align. We use a red color to indicate the same motif in the low-frequency sinusoidal wave, where we see that the high-frequency information in different motifs looks very different on the left, while it looks the same on the right.}
    \label{fig:superposingalignment}
\end{figure}

Here, we show one additional example where if we change the high-frequency mode by a little bit, then we can completely change the models' performance. In particular, consider the following new context:
\[
    f(t) = \sin(8t) + \sin(128 t).
\]
In some sense, $128$ is even ``larger'' than $117$; but when we sample this signal and use a Chronos-Bolt model with $k = 16$ to predict, we see that the performance becomes much better (see~\Cref{fig:superposingalignment})! 
What makes that happen? 
Notice that $128$ is an integral multiple of $8$, but $117$ is not. Therefore, even if the high-frequency information is attended in the way dominated by the low-frequency, it does not cause any issue in preserving the high-frequency information. 
That is, we do not suffer from aliasing errors. 
(As an analogy, when you use your cellphone's camera to take a picture of the screen of your television, you often notice some stripes --- this is due to the mismatch between the resolutions of your digital camera and the television screen --- but there are certain cases where the stripes do not appear, which happen exactly when the resolutions match each other.) 
For example, in~\Cref{fig:superposingalignment}, we use red colors to indicate the same motif in each period of the low-frequency wave $\sin(8t)$. 
On the left, when the periodicities of $\sin(8t)$ and $\sin(117t)$ do not match, we see that the high-frequency information in each motif is different; but since the periodicities of $\sin(8t)$ and $\sin(128t)$ match each other, we have exactly the same high-frequency content in each red motif.

The significance of this example is two-fold:
\begin{enumerate}[leftmargin=*]
    \item First, it validates our hypothesis that the attention scores are mainly driven by the low-frequency content that results in the frequency bias when using a large patch size.
    \item Second, it shows that the frequency bias is not a guaranteed notion. When the periodicities of the high- and low-frequency match each other, its impact is significantly reduced.
\end{enumerate}

\paragraph{Another Use Case of the Frequency Bias: Concatenation.}

Previously, our discussion mainly focused on the case where we have a superposition of a low- and high-frequency, because this is usually the way that people care about the frequency bias. 
However, there is another way that two different frequencies can come into the same context --- via concatenation.

\begin{figure}[H]
    \centering
    \begin{minipage}{0.32\textwidth}
        \includegraphics[width=1\linewidth, trim = 0cm 0cm 0cm 0.5cm, clip]{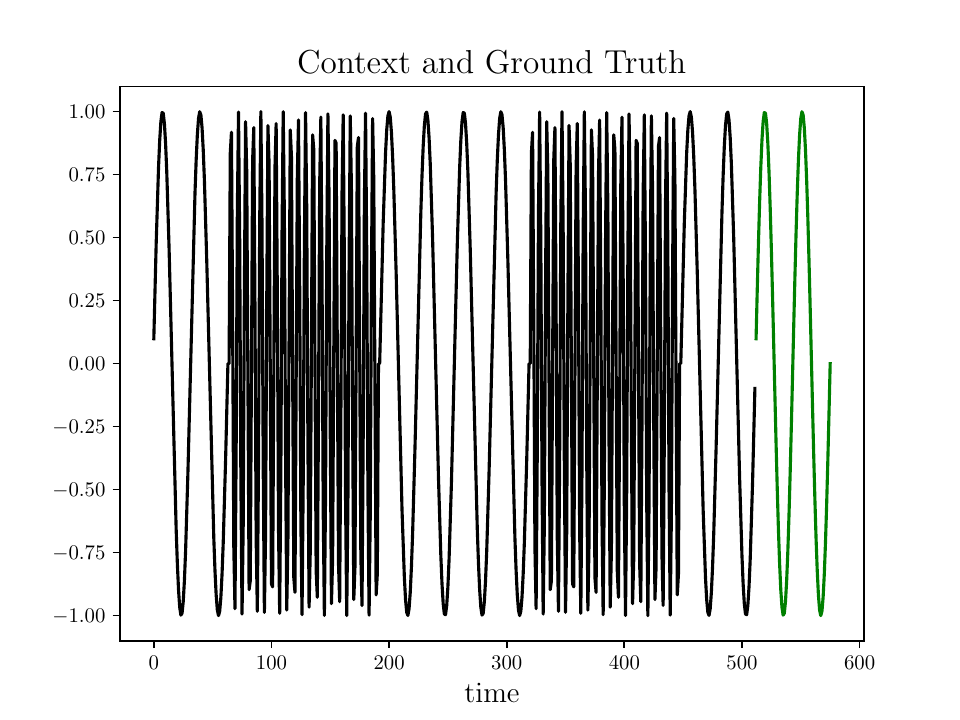}
    \end{minipage}
    \hfill
    \begin{minipage}{0.32\textwidth}
        \includegraphics[width=1\linewidth, trim = 0cm 0cm 0cm 0.5cm, clip]{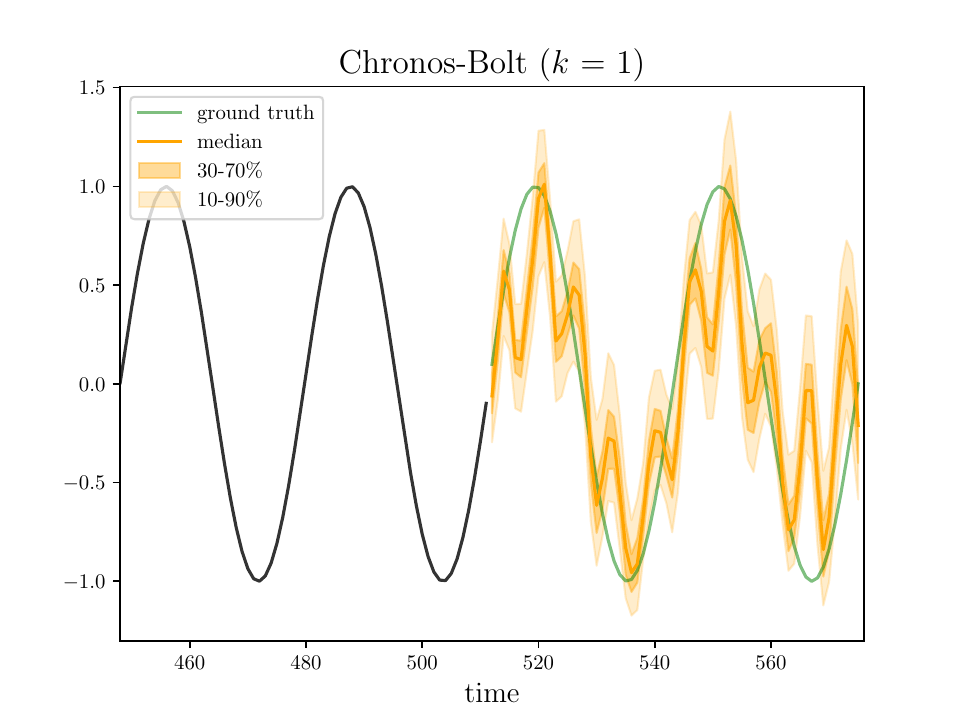}
    \end{minipage}
    \hfill
    \begin{minipage}{0.32\textwidth}
        \includegraphics[width=1\linewidth, trim = 0cm 0cm 0cm 0.5cm, clip]{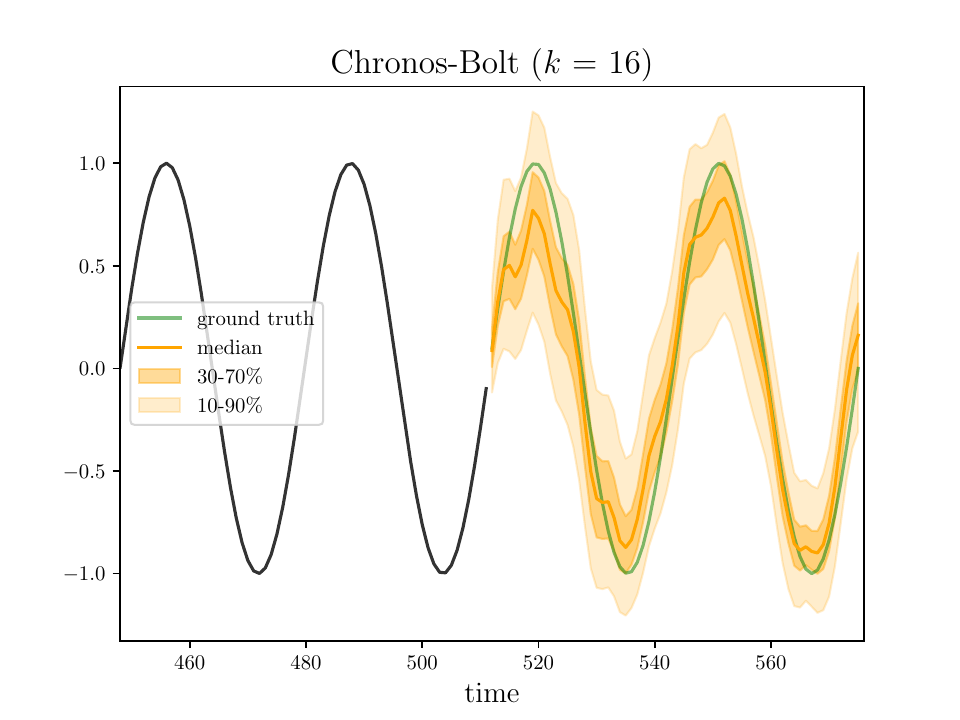}
    \end{minipage}
    \caption{We show the forecast of two models, Chronos-Bolt with a patch size $k = 1$ and one with $k = 16$, given a context formed by concatenating low-frequency modes and high-frequency ones. We see that the high-frequency information significantly bleeds into Chronos-Bolt's forecast when $k = 1$, but the forecast is better when the patch size $k = 16$.}
    \label{fig:freqconcatenation}
\end{figure}

In~\Cref{fig:freqconcatenation}, we show the forecast of Chronos-Bolt models, with a patch size of $k = 1$ and $16$, respectively, given an input formed by concatenating low- and high-frequency motifs. In this setting, we see that having a larger patch size helps prevent the high-frequency information from bleeding into the low-frequency forecast.

\begin{figure}[H]
    \myvspace{1\baselineskip}
    \begin{minipage}{0.45\textwidth}
        \begin{overpic}[width=1\linewidth, trim = 0cm 0cm 0cm 1.5cm, clip]{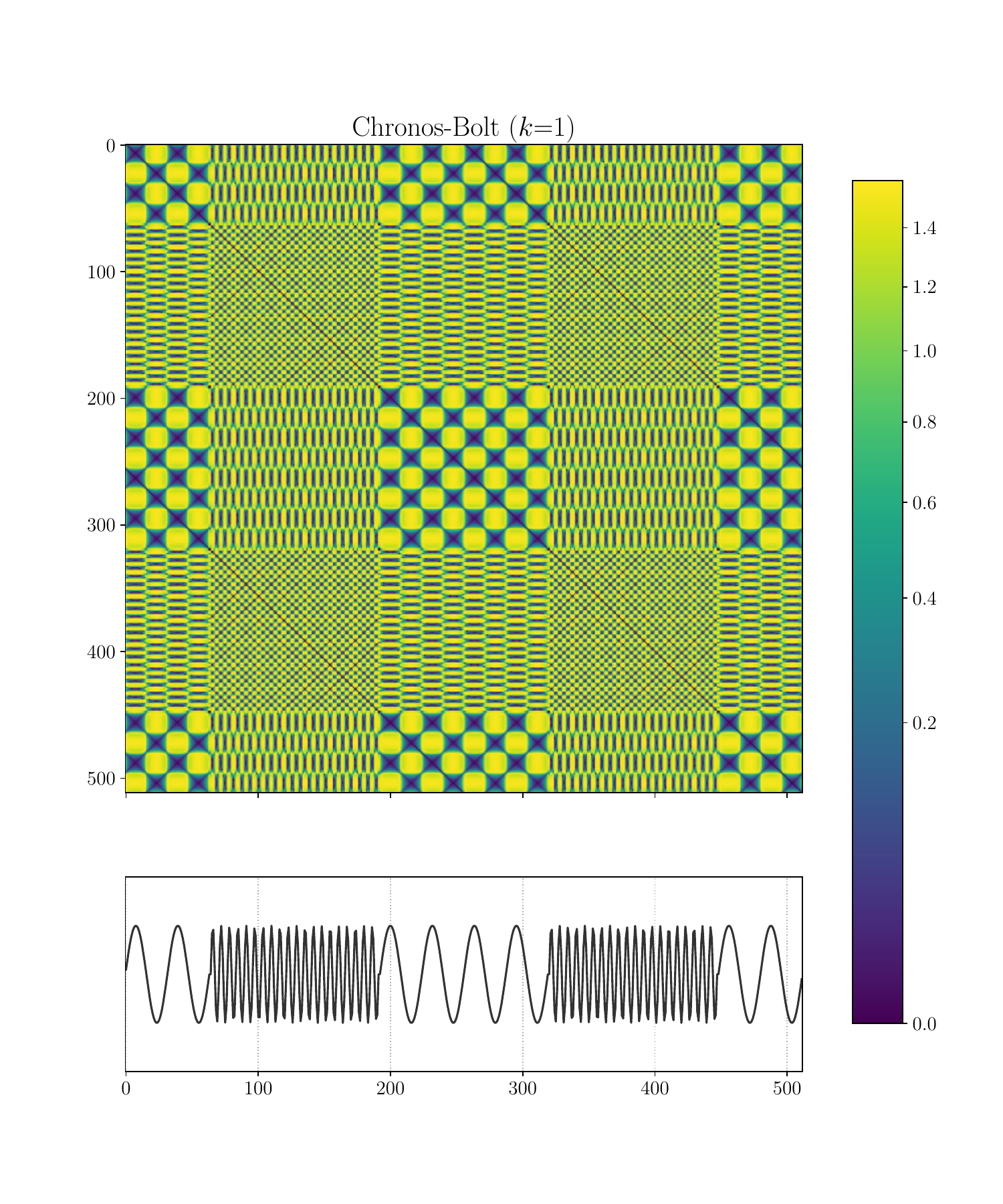}
            \put(56,102){\textbf{Angles between Embedded Vectors}}
        \end{overpic}
    \end{minipage}
    \hfill
    \begin{minipage}{0.45\textwidth}
        \includegraphics[width=1\linewidth, trim = 0cm 0cm 0cm 1.5cm, clip]{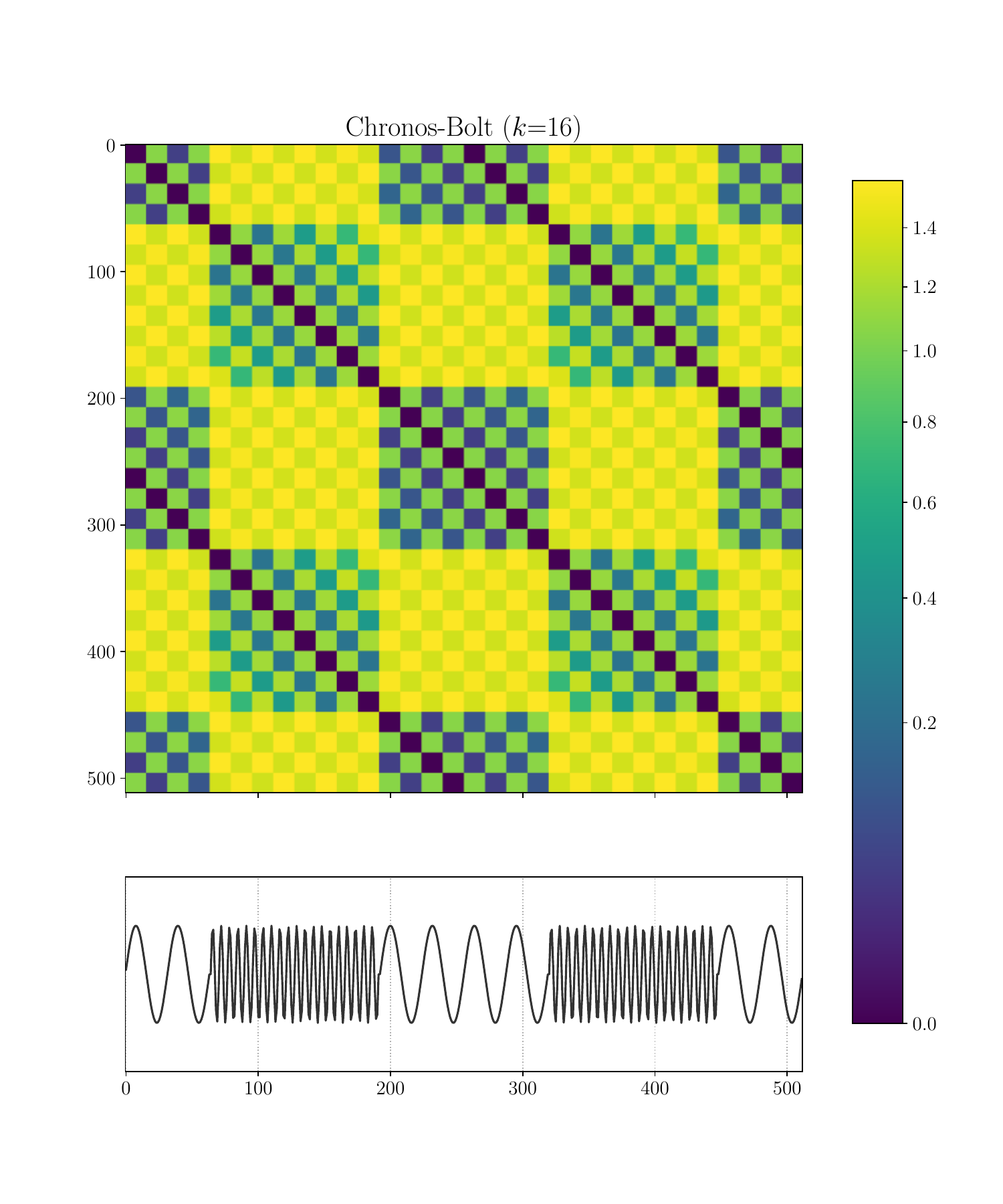}
    \end{minipage}

    \vspace{1.5\baselineskip}

    \begin{minipage}{0.45\textwidth}
        \begin{overpic}[width=1\linewidth, trim = 0cm 1cm 0cm 1.5cm, clip]{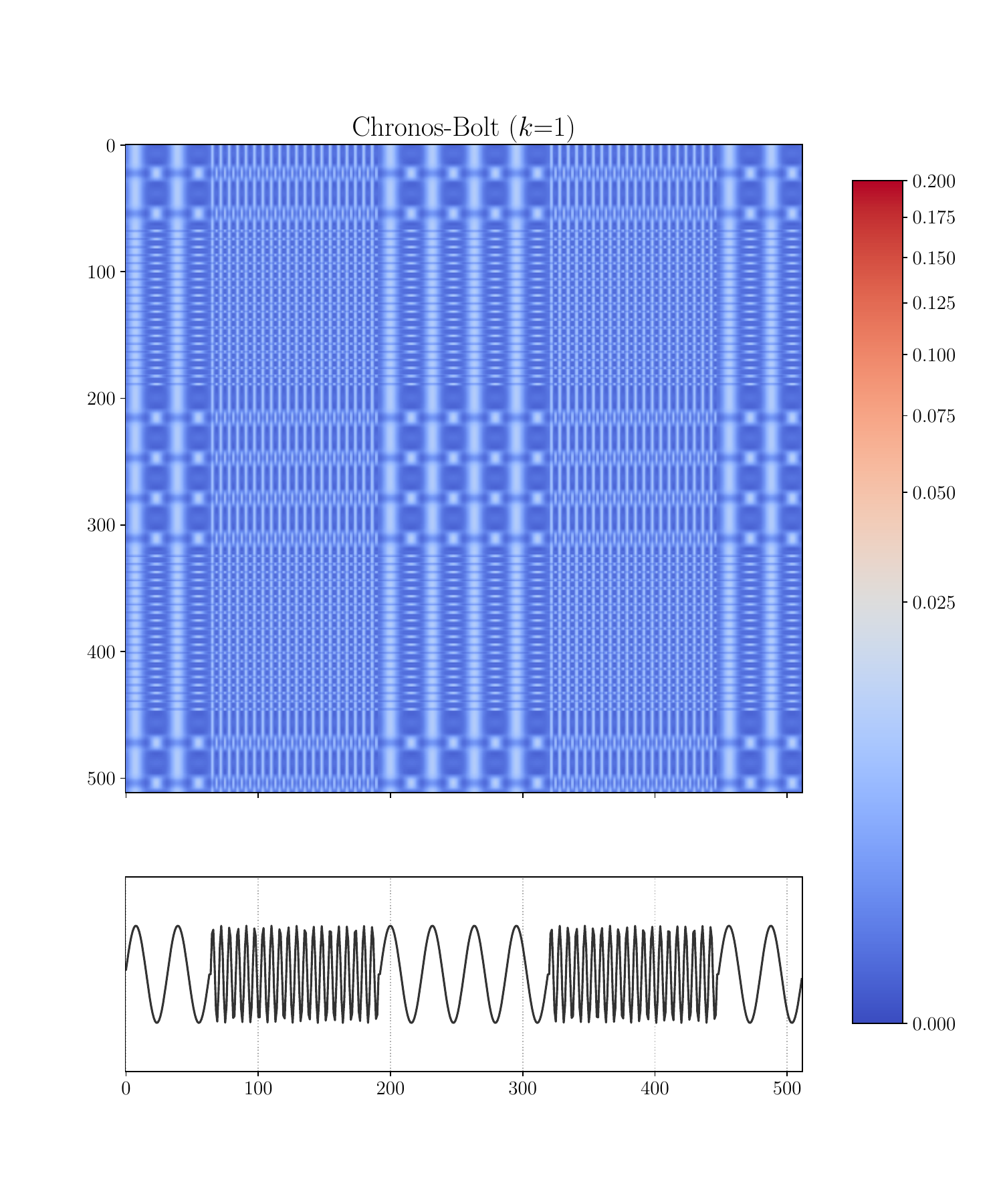}
            \put(62,102){\textbf{Magtinude of Attention Scores}}
        \end{overpic}
    \end{minipage}
    \hfill
    \begin{minipage}{0.45\textwidth}
        \includegraphics[width=1\linewidth, trim = 0cm 1cm 0cm 1.5cm, clip]{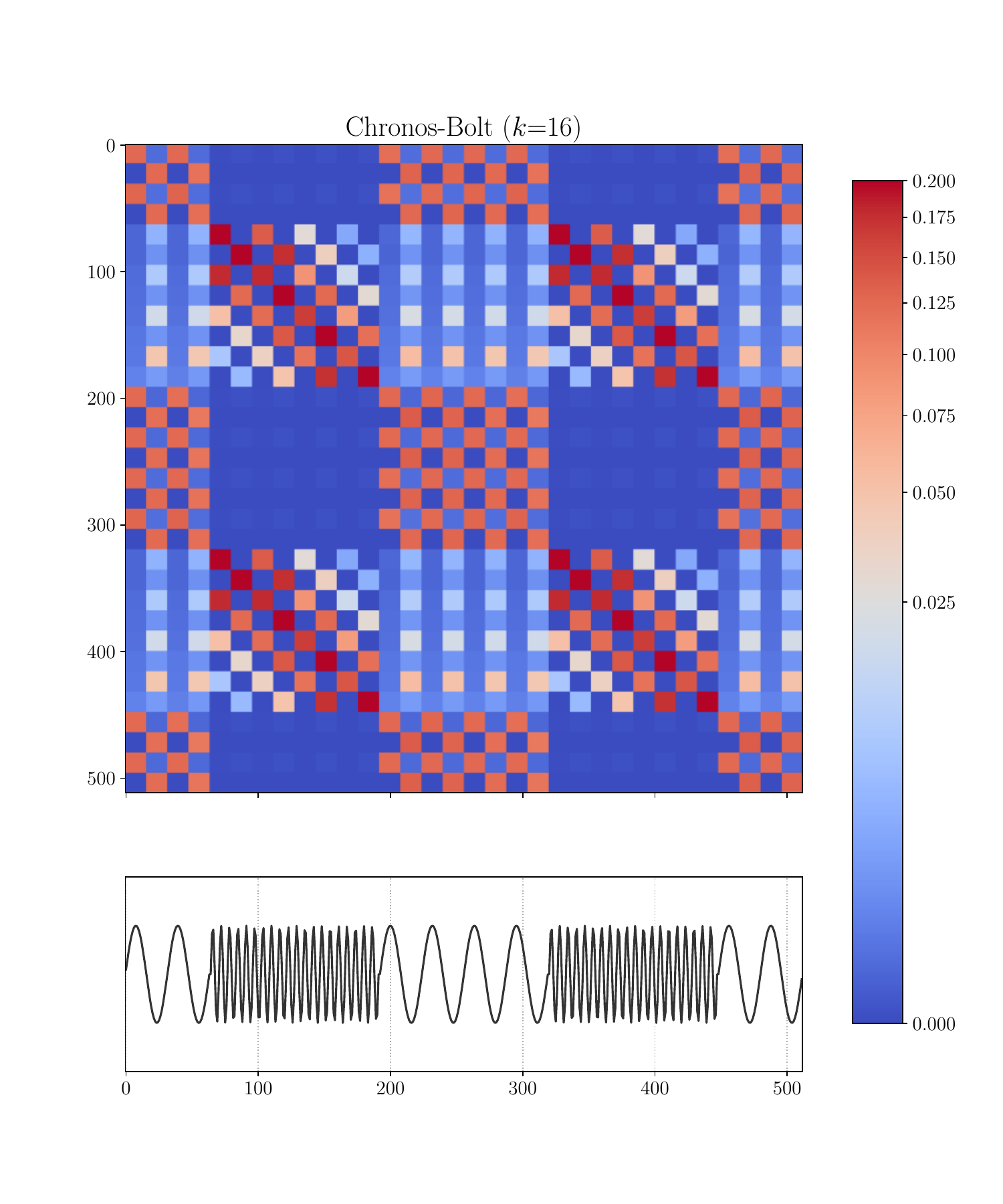}
    \end{minipage}
    
    \caption{We show the angles between different pairs of embedded patches in the hidden space, given the context in~\Cref{fig:freqconcatenation}, as well as the magnitudes of the attention scores.}
    \label{fig:freqconcateatt}
\end{figure}

This finding is not surprising, and it provides yet another example of~\Cref{thm.frequencybias}. 
The fact that high- and low-frequency patches are orthogonal to each other in the hidden space means that they are unlikely to interact with each other in a Transformer (see the top row of~\Cref{fig:freqconcateatt}). 
From the attention scores shown in~\Cref{fig:freqconcateatt}, we see that with a patch size of $1$, a lot of the high-frequency information is attended to the low-frequency information.
However, when $k = 16$, the high-frequency patches mainly talk to the high-frequency ones, and similarly for the low-frequency patches, but the low-frequency patches rarely attend to the high-frequency ones. This explains why we have less contamination using a larger patch size.


\subsection{Details of the Periodicity Bias}
\label{app:periodicity}

In this subsection, we provide more details on the periodicity bias, one of the two temporal biases (the other being the frequency bias; see \Cref{app:frequency}), from~\cref{sec:frequency}. 
Our discussion is split into two parts.
First, we perform a more detailed analysis of the \texttt{[REG]} token's role in the preservation of the periodicity (\Cref{app:periodicity-reg}).
Then, we briefly discuss how we measure the periodicity bias induced by patching (\Cref{app:periodicity-design}).

\subsubsection{An Investigation of the [REG] Token}
\label{app:periodicity-reg}

\begin{center}
    \includegraphics[width=0.86\linewidth, trim= 0 38cm 0 0, clip]{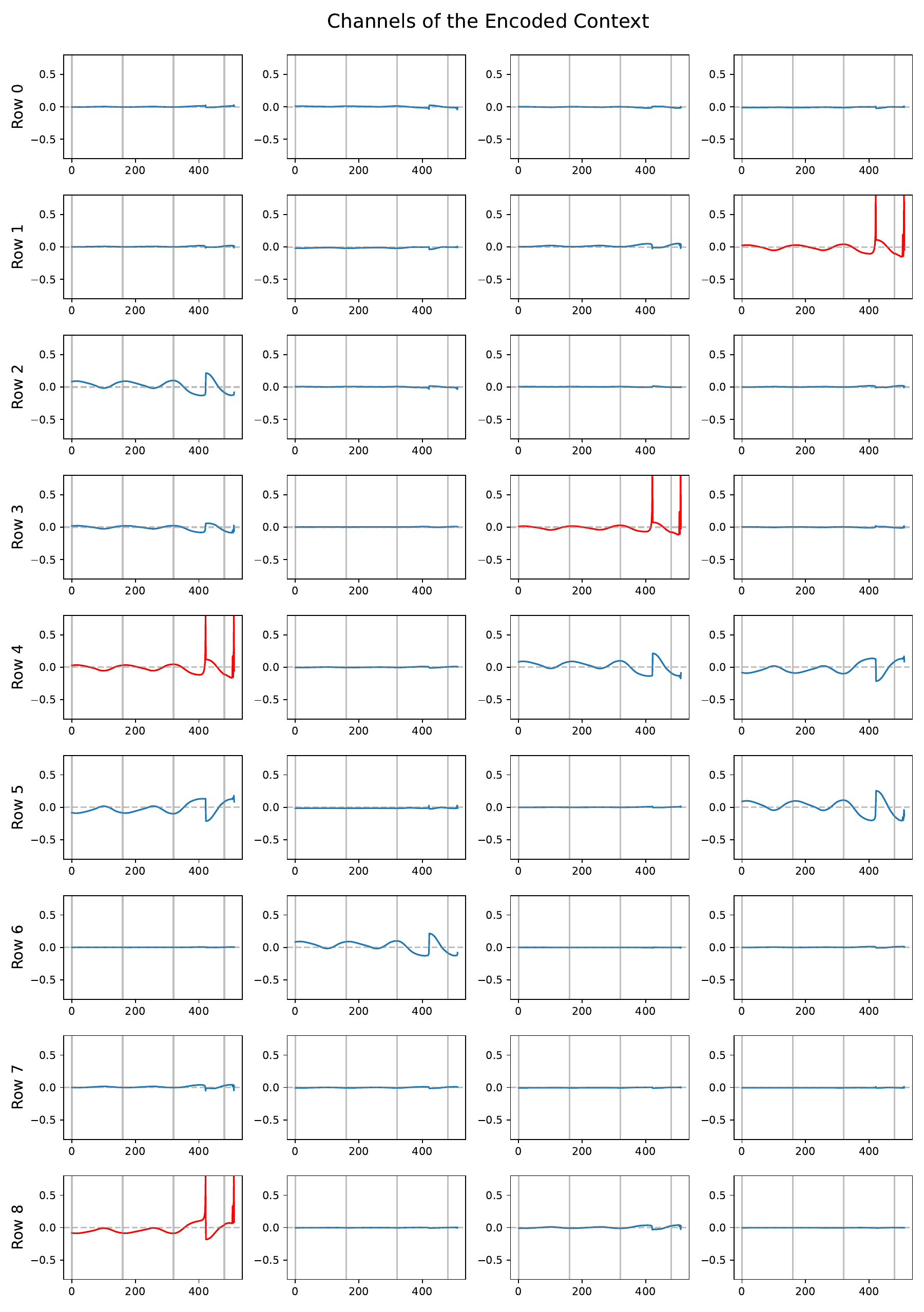}
\end{center}

\begin{figure}[H]
    \centering
    \includegraphics[width=0.86\linewidth, trim= 0 0 0 43cm, clip]{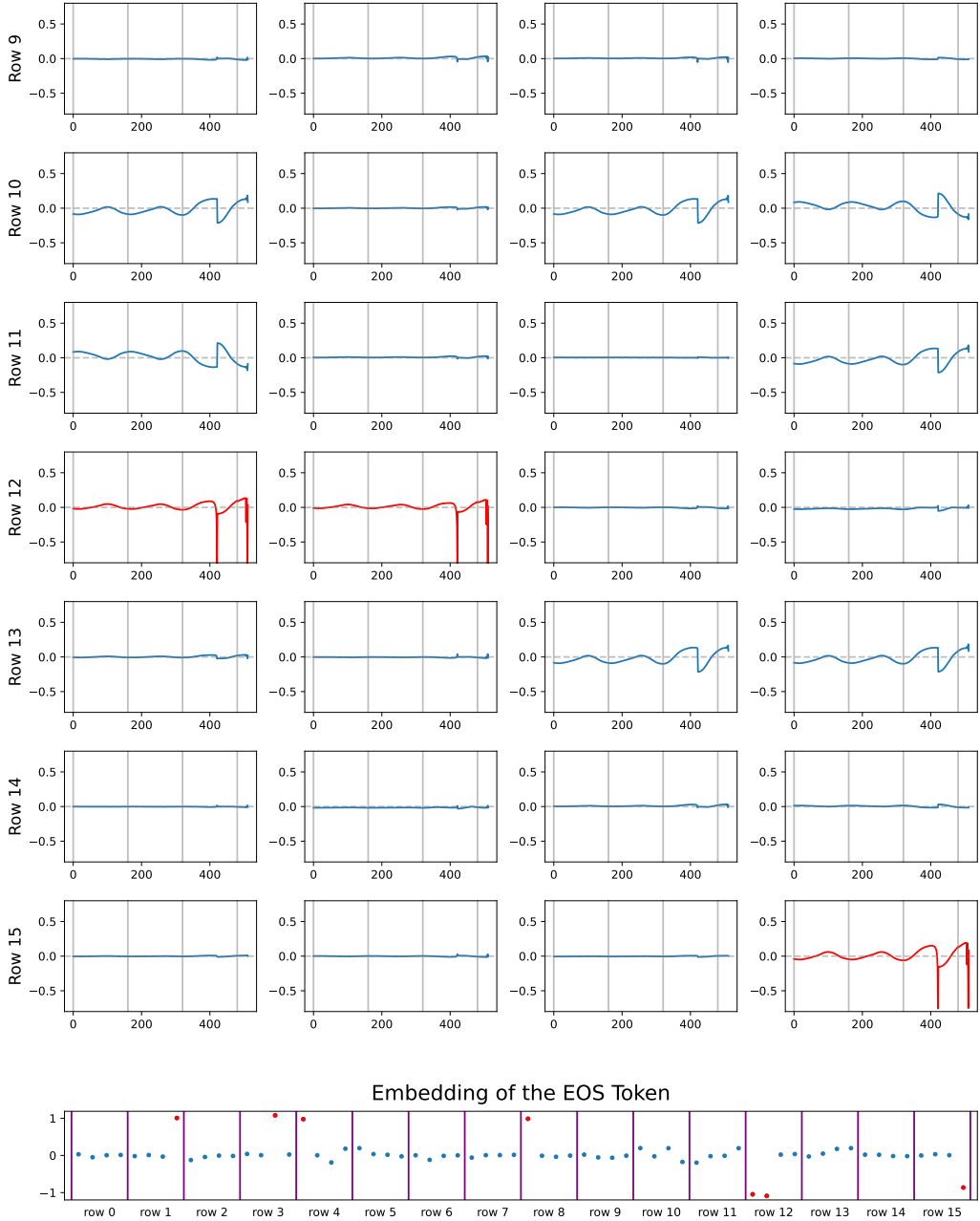}
    \caption{We give a sinusoidal wave to the encoder of a Chronos-Bolt model (patch size $k=1$) and take the pre-$\mathbf{W}_O$ output from the first attention layer. We plot the $64$ channels in the first head of this output, in the row-major order. We see that the periodicity in several channels, highlighted in red, is much less-well preserved than in others. The bottom panel shows the $64$ values corresponding to the first head of the \texttt{[REG]} token in the first attention layer, where we see the large values correspond to exactly the very non-periodic channels.}
    \label{fig:periodallchannels}
\end{figure}

Recall~\Cref{fig:periodicity}, where we noted that when the \texttt{[REG]} token is present, Chronos-Bolt's Transformer becomes significantly worse at preserving periodic patterns. To understand why this happens, we present a case study where our input is a sinusoidal wave. We pass this sinusoidal wave into Chronos-Bolt's encoder (patch size $k = 1$) and compute the context $\mathbf{z} \in \R^{d \times L}$ after applying the first attention layer. Here, we have $d = 512$ and $L = 512$. There are $8$ heads, so the first head of $\mathbf{z}$ is a matrix $\mathbf{z}^{(1)} \in \R^{64 \times 512}$. We plot all $64$ channels of $\mathbf{z}^{(1)}$ in~\Cref{fig:periodallchannels}. We see clearly some channels where the periodicity of the input context is seriously affected. Those channels are highlighted in red.

Now, we claim that these serious losses of periodicity are primarily due to the existence of the \texttt{[REG]} token. In the bottom panel of~\Cref{fig:periodallchannels}, we show all $64$ values corresponding to the \texttt{[REG]} token. We see that there are clearly a couple of them that are large in magnitude. More strikingly, there is a strong correlation between the magnitude of the \texttt{[REG]} token and the loss of periodicity. That is, the entries of large values in \texttt{[REG]} are exactly the channels in $\mathbf{z}^{(1)}$ that have a poor periodicity.

\begin{figure}[H]
    \centering
    \includegraphics[width=0.9\linewidth]{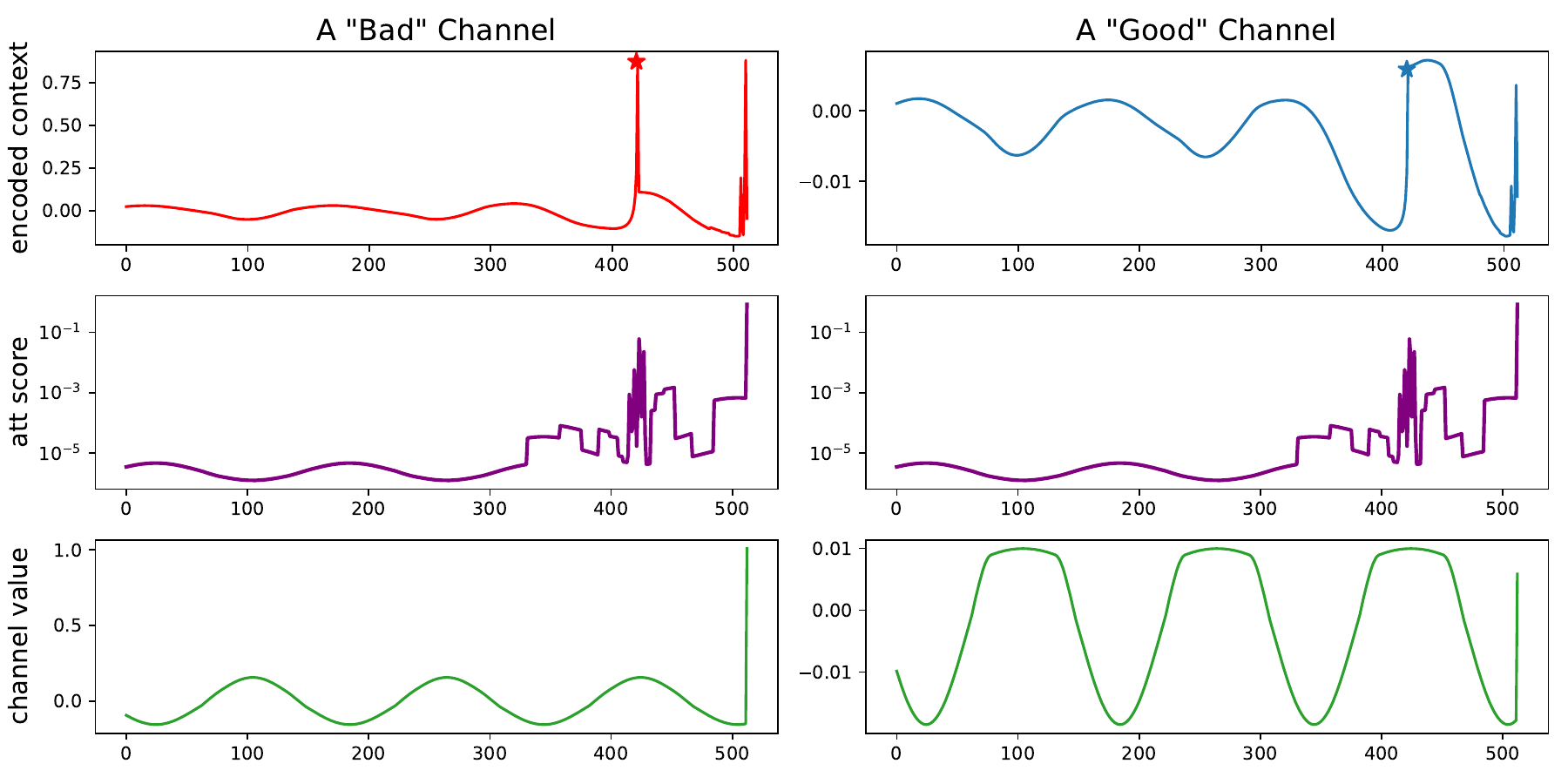}
    \caption{We show how a very non-periodic channel and a more periodic channel from~\Cref{fig:periodallchannels} are computed. On the top row, we plot these two channels, and the stars are the positions that we investigate in the bottom two panels of the figure. The starred position is computed by taking the inner product of the attention scores vector, shown in the middle row, and the value vector, shown in the bottom row. We see that the attention scores used to compute the starred position are particularly large at the \texttt{[REG]} token, and the large value of the \texttt{[REG]} token seriously destroys the periodicity of the signal.}
    \label{fig:periodicwithREG}
\end{figure}

This is not hard to understand. If the \texttt{[REG]} token has a large value, without masking, this value is written into a previous position that has a large attention score on it, perturbing the periodic pattern. To illustrate this with a clear example, we take a ``bad'' channel of $\mathbf{z}^{(1)}$, where the periodicity is severely destroyed, and a ``good'' channel. In~\Cref{fig:periodicwithREG}, we see that the main difference happens at the starred position, where in the bad channel, the output blows up at this position. In the middle row of~\Cref{fig:periodicwithREG}, we show the attention scores that are used to compute this position. We see that the last attention score is very large, meaning that a lot of the \texttt{[REG]} token's value is written into the starred position. More precisely, the starred position is computed by taking the inner product of the attention scores and the channel values:
\[
    \mathbf{z}_{k,j^*} = \sum_{i=1}^L s_{j^*}^{i} V_{k,i},
\]
where $(s_{j^*}^{1}, \ldots, s_{j^*}^{L})$ are the self-attention scores shown in purple and $(V_{k,1}, \ldots, V_{k,L})$ are the values in the $k$th channel shown in green. Since $s_{j^*}^{L}$ is large, whether the periodicity is well-preserved reduces to a question of whether $V_{k,L}$ is large. That is why we see the strong coupling of the periodicity and the magnitude of the \texttt{[REG]} token's value in~\Cref{fig:periodallchannels}.

\begin{figure}[H]
    \centering
    \includegraphics[width=0.9\linewidth]{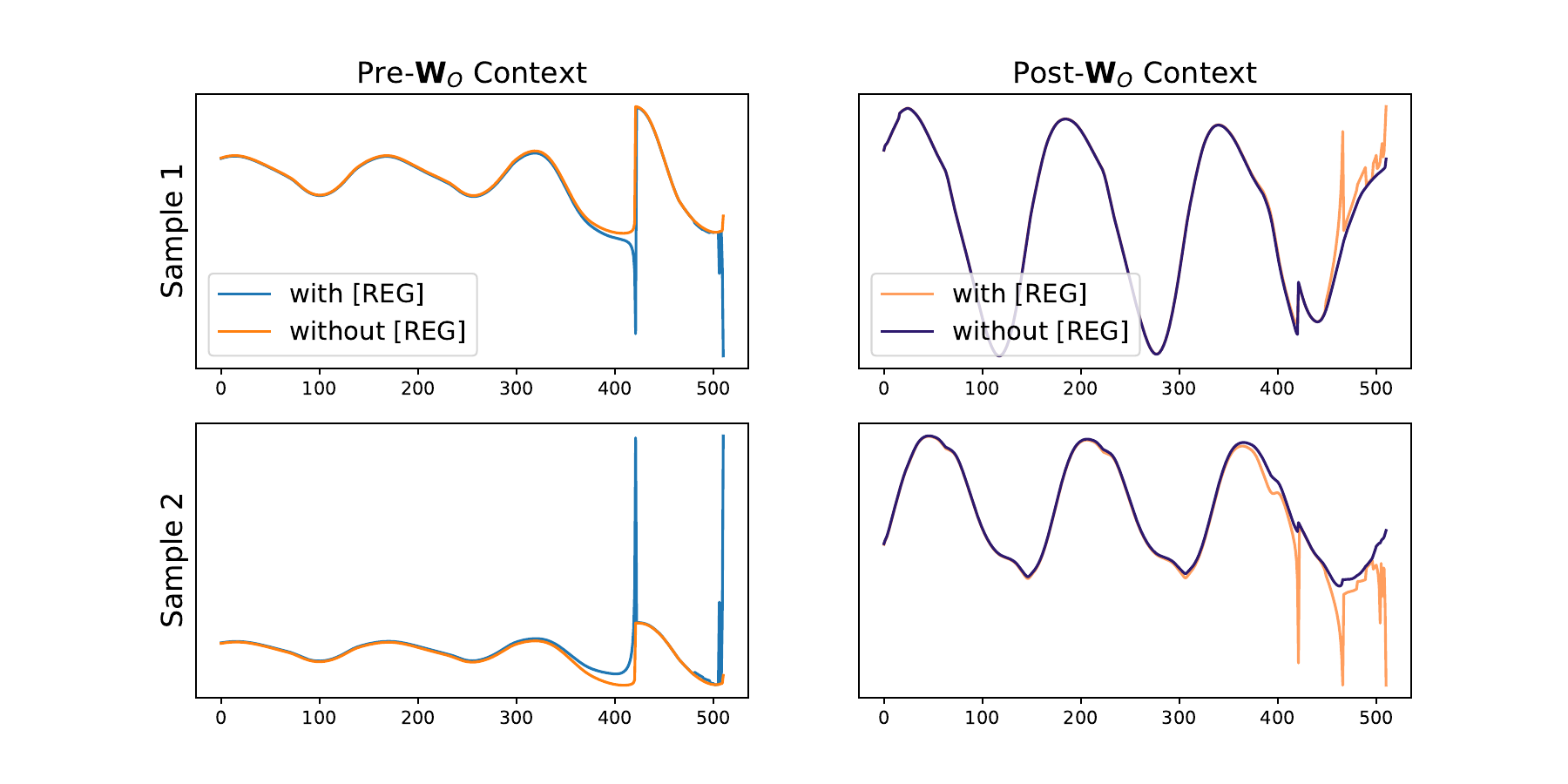}
    \caption{We give a sinusoidal wave to the encoder of a Chronos-Bolt model (patch size $k=1$) and take the pre-$\mathbf{W}_O$ and post-$\mathbf{W}_O$ output from the first attention layer. We compute the output in two ways: using the default setting where every position is attended to another position; and applying a mask to prevent the value of the \texttt{[REG]} token from being written into the previous outputs. We see that with masking, the periodicity is much better preserved.}
    \label{fig:REGornot}
\end{figure}

The question that remains is: does removing the \texttt{[REG]} token help? The answer to this is positive. In~\Cref{fig:REGornot}, we show two models: in the first model, we compute Chronos-Bolt's encoded values as usual, which corresponds to exactly the figures shown in~\Cref{fig:periodallchannels}; in the second model, we apply a masking to prevent the \texttt{[REG]} token's value from being written into the previous tokens, i.e., the \texttt{[REG]} token is write-only. We show both random channels before and after applying the channel mixing layer $\mathbf{W}_O$. Clearly, masking the \texttt{[REG]} token helps the preservation of the periodicity.

\subsubsection{Design of Experiments in~\Cref{fig:periodicity}}
\label{app:periodicity-design}

Recall~\Cref{fig:periodicity}, where we explained that patching also hides the periodic patterns. This is also not surprising. The simplest way to view this is by considering aliasing. If the patch's periodicity mismatches the periodicity of the input signal, then we lose periodic information by patching. To show this, we generate two histograms in~\Cref{fig:periodicity} as follows:

\begin{enumerate}[leftmargin=*]
    \item Randomly sample a sequence $\mathbf{x} \in \R^{L}$ from Chronos' training dataset~\citep{ansari2024chronos}, which includes both real-world and synthetic time series. Sample until we have $L \gg 64$.

    \item Take the last $64$ elements of $\mathbf{x}$: $\mathbf{x}^* = \mathbf{x}_{(L-64+1):L}$.

    \item Break the context into patches and form motifs of length $64$: $\mathbf{x}^{(1)} = \mathbf{x}_{1:64}$, $\mathbf{x}^{(2)} = \mathbf{x}_{(1+k):(64+k)}$, $\mathbf{x}^{(3)} = \mathbf{x}_{(1+2k):(64+2k)}, \ldots$, where none of them overlaps $\mathbf{x}^*$.

    \item Compute the $R^2$-score between $\mathbf{x}^*$ defined in step 2 and each $\mathbf{x}^{(i)}$ formed in step 3. We call the maximum $R^2$ score the ``best matching score.''
\end{enumerate}

\begin{figure}[H]
    \centering
    \includegraphics[width=0.9\linewidth]{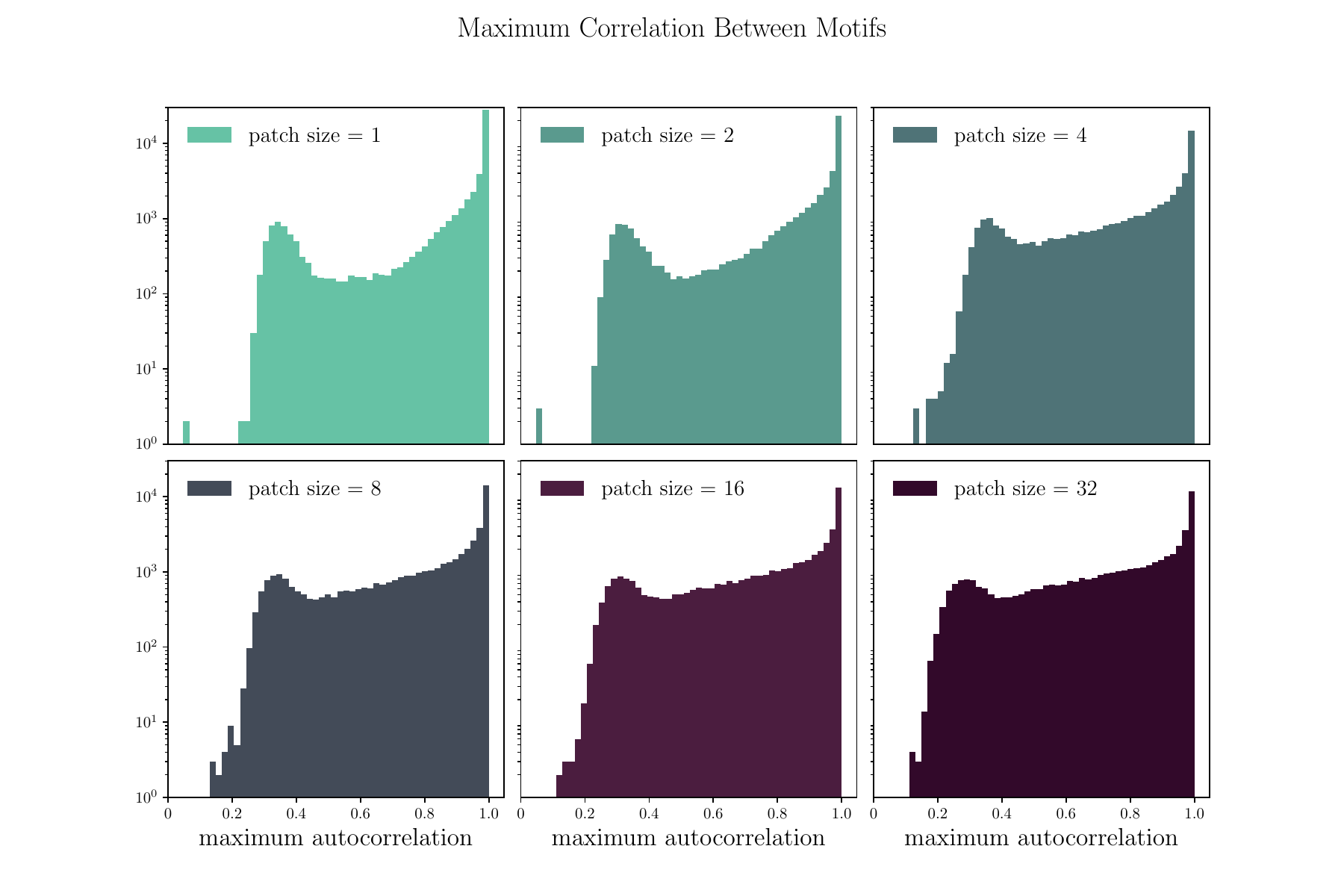}
    \caption{The maximum correlation between the last motif in a context and each motif in the earlier context, which are defined by patching. We see that as the patch size increases, correlations generally become smaller, indicating more challenges in identifying the periodicity in the context.}
    \label{fig:all_correlations_patch}
\end{figure}

We repeat this step for over $10,000$ sampled sequences to form the histograms in~\Cref{fig:periodicity}. \Cref{fig:all_correlations_patch} shows more results for different patch sizes $k$.

\begin{figure}[H]
    \centering
    \includegraphics[width=0.9\linewidth]{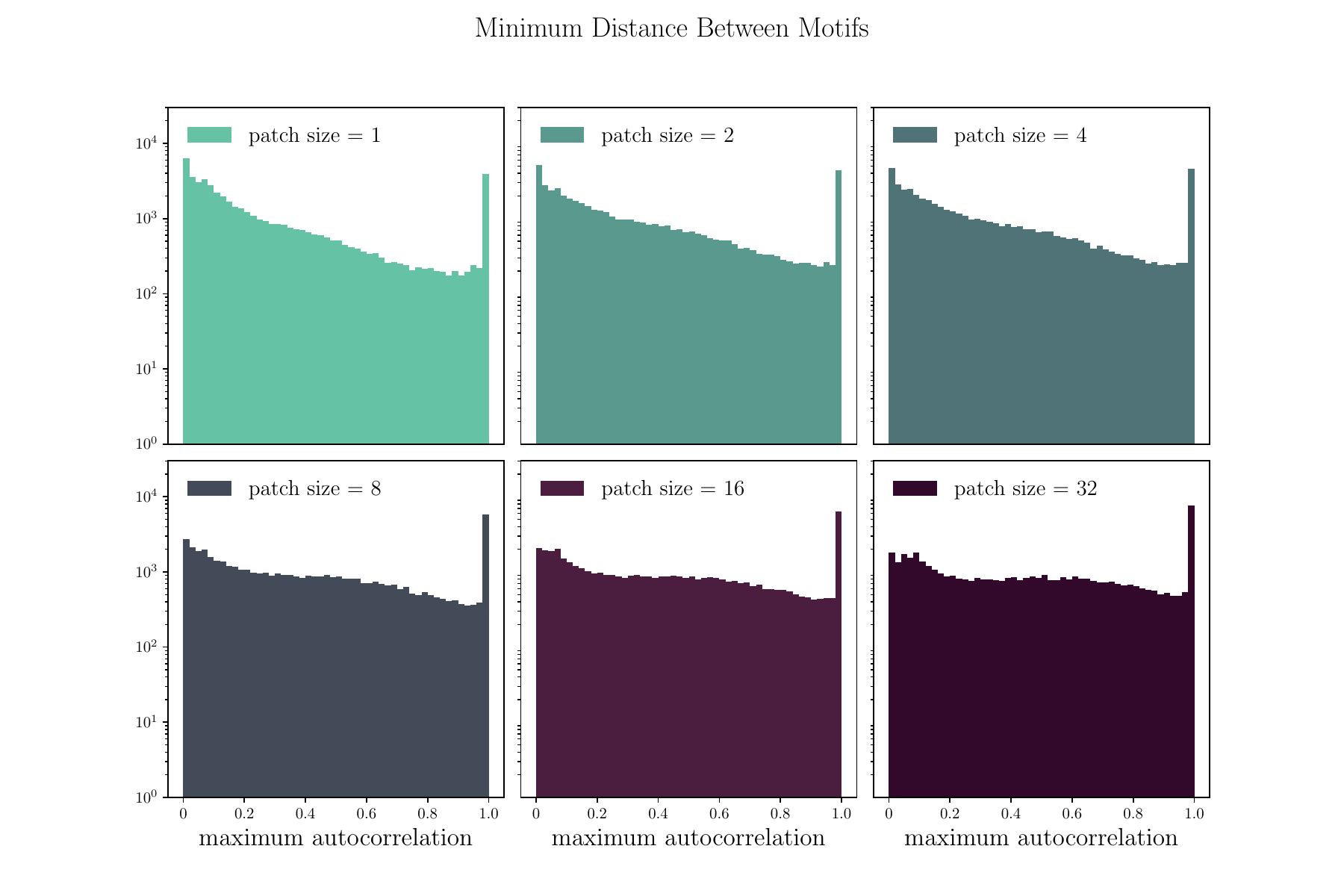}
    \caption{The minimum distance between the last motif in a context and each motif in the earlier context, which are defined by patching. We see that as the patch size increases, distances generally become larger, indicating more challenges in identifying the periodicity in the context.}
    \label{fig:all_distances_patch}
\end{figure}

In addition to the $R^2$-score, we also consider another metric, which measures the relative distance between the last motif and earlier ones. This is done by changing step 4 from above into the following:
\begin{itemize}[leftmargin=0.6cm]
    \item [4'.] Compute the relative distance between $\mathbf{x}^*$ defined in step 2 and each $\mathbf{x}^{(i)}$ formed in step 3:
    \[
        \|\mathbf{x}^* - \mathbf{x}^{(i)}\|_2 / (\|\mathbf{x}^*\| + 10^{-8}),
    \]
    where $10^{-8}$ is there to prevent a zero denominator.
\end{itemize}

In~\Cref{fig:all_distances_patch}, we see that as the patch size increases, distances generally become larger, indicating more challenges in identifying the periodicity in the context.

\clearpage

\section{Details of the Geometric Bias (from \Cref{sec:geometry})}
\label{app:geomtric}

In this section, we provide more details on the geometric bias, which was discussed in \Cref{sec:geometry}.
The discussion is made up of three parts: 
the \emph{angles} between the embedded vectors, which lead to a \emph{locality preference} (\Cref{app:geomtric-angular}); 
the \emph{distance} between the embedded vectors, which gives rise to a \emph{scale-dependent treatment} (\Cref{app:geomtric-distance}); and 
the \emph{norm} of the embedded vectors, yielding the \emph{offset-aware embedding} (\Cref{app:geomtric-norm}). 
This appendix is broken into three parts, corresponding to each of the three biases, respectively.

\subsection{Details of the Angular Bias}
\label{app:geomtric-angular}

\subsubsection{Design of Experiments in~\Cref{fig:locality}}

Recall~\Cref{fig:locality}, where we presented an experiment where we collected the self-attention scores of both the Chronos and Chronos-Bolt models, which revealed that the self-attention scores of the Chronos model have more of a bimodal distribution, while those of the Chronos-Bolt model followed a more unimodal distribution. 
To collect this set of self-attention scores, we sample $1000$ random samples from Chronos and Chronos-Bolt's in-domain evaluation corpus, and we pass them into the two models to collect all self-attention scores. 
Due to the space limitation, however, we were only able to show the histogram in the log--log scale, which hid some useful information.

\begin{figure}[H]
    \centering
    
    \begin{minipage}{0.24\textwidth}
        \includegraphics[width=1\textwidth]{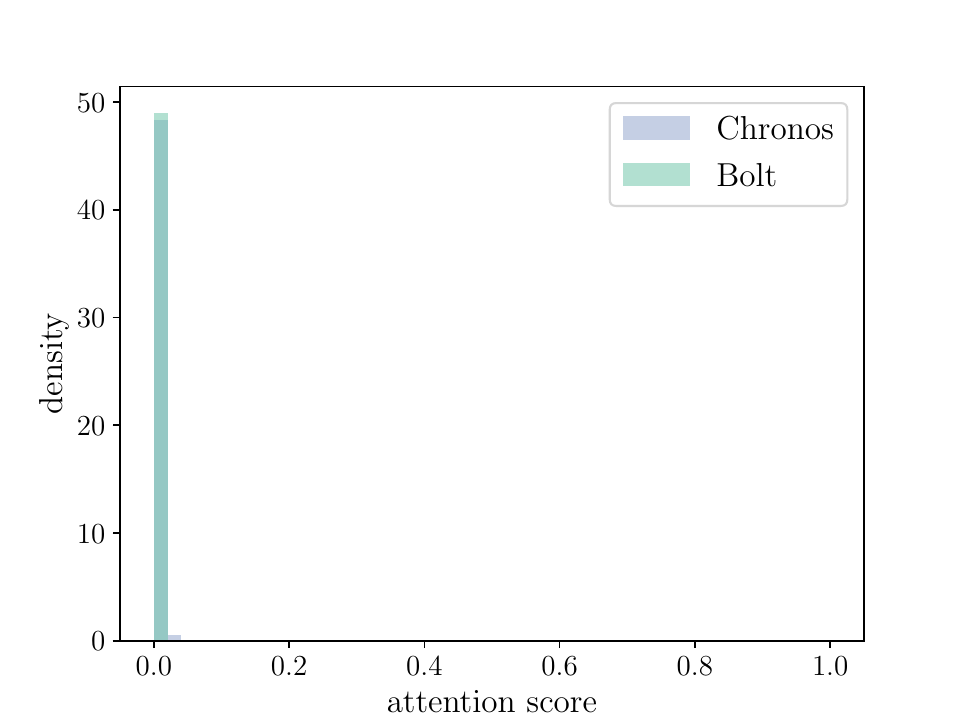}
    \end{minipage}
    \hfill
    \begin{minipage}{0.24\textwidth}
        \includegraphics[width=1\textwidth]{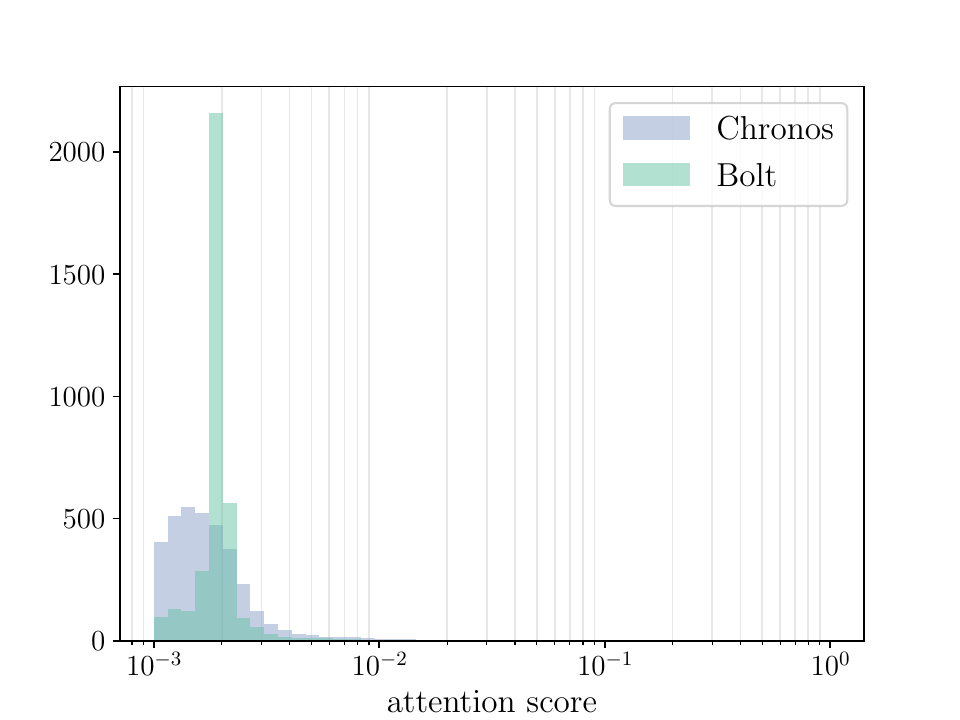}
    \end{minipage}
    \hfill
    \begin{minipage}{0.24\textwidth}
        \includegraphics[width=1\textwidth]{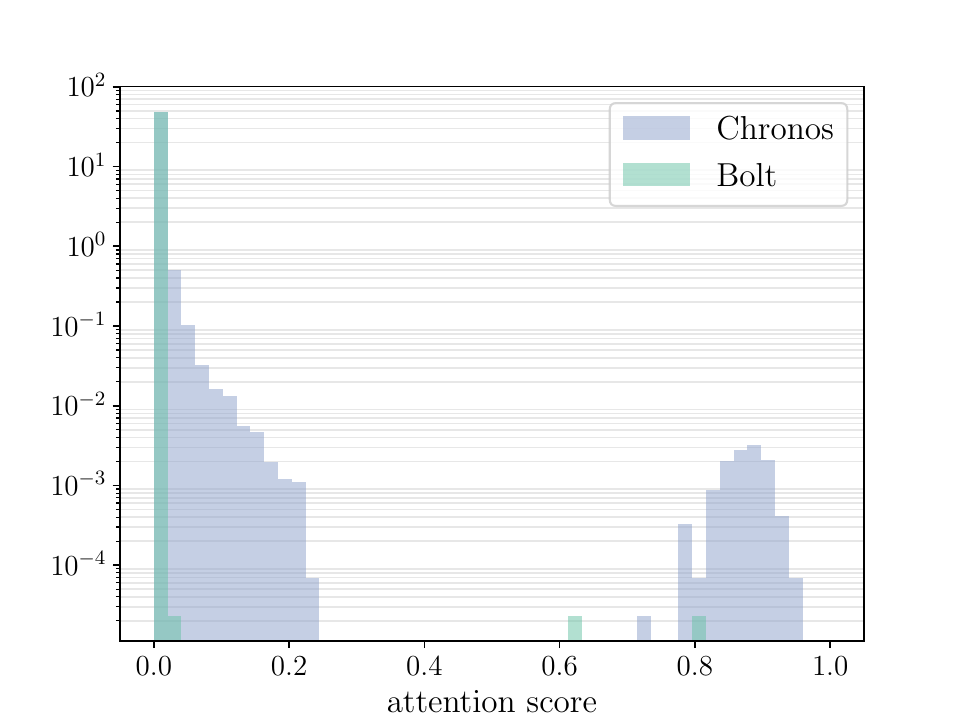}
    \end{minipage}
    \hfill
    \begin{minipage}{0.24\textwidth}
        \includegraphics[width=1\textwidth]{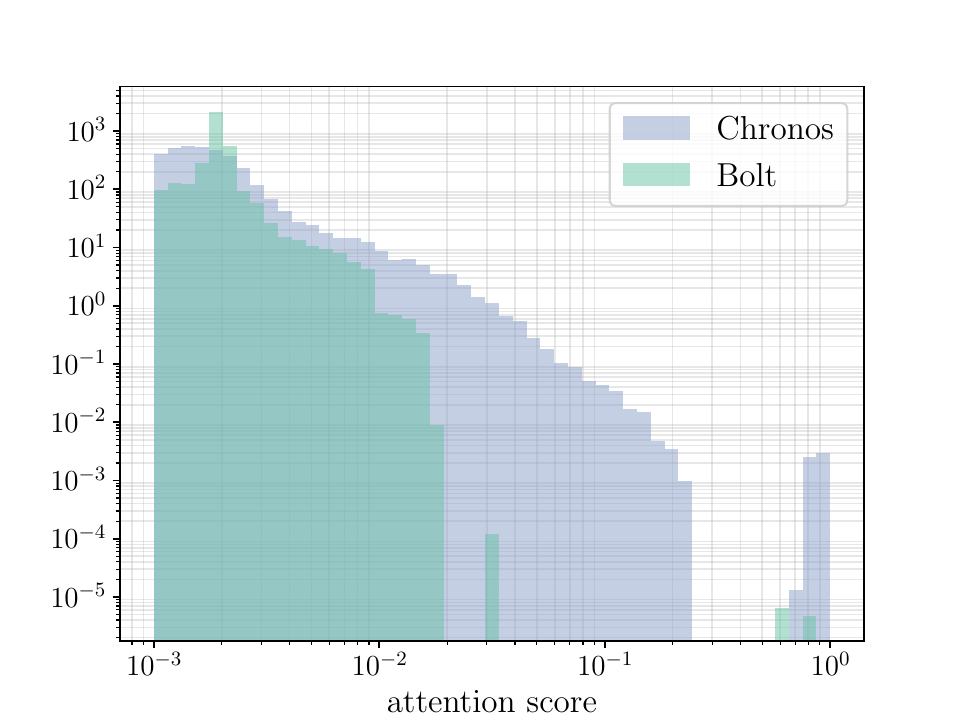}
    \end{minipage}

    \caption{We reproduce the histograms in~\Cref{fig:locality} on four different scales: linear, semilog-x, semilog-y, and log--log, respectively. We can clearly see the bimodal distribution of Chronos' self-attention scores in the semilog-y plot, while the semilog-x plot highlights the existence of small self-attention scores in Chronos, compared to Chronos-Bolt's mediocre ones.}
    \label{fig:fourscales}
\end{figure}

In~\Cref{fig:fourscales}, we reproduce this histogram on all four common scales: the linear scale, the semilog-x scale, the semilog-y scale, and the log--log scale, respectively. From these plots, we can see at least two clean messages:
\begin{enumerate}[leftmargin=*]
    \item From the semilog-y plot, i.e., the third panel, we see a clear bimodal distribution in Chronos' self-attention scores, which does not appear in Chronos-Bolt's self-attention scores. Note that the modes in Chronos' bimodal distribution do not have the same weight --- that is essentially why the two modes are hard to visualize when not using a logarithmic y-scale. This is not surprising, however, because if a single self-attention score is large, then it implies that the remaining $L-1$ singular values are all small, so the weight of the large-score mode must be much lower than the weight of the small-score mode.
    \item From the semilog-x plot, i.e., the second panel, we see that the small attention scores in Chronos are much smaller than the attention scores in Chronos-Bolt. For Chronos-Bolt, most self-attention scores are small, but not tiny. This corroborates our intuition that Chronos-Bolt performs more context mixing by putting a relatively small and more uniformly distributed attention score in each element across the sequence.
\end{enumerate}

\subsubsection{Three Examples to Illustrate ``Parroting''}

In the main article, and in previous work, e.g.,~\citet{zhang2024zero,zhang2025context}, the idea of ``parroting'' arises. 
This basically says that Chronos, in contrast to Chronos-Bolt and other models that use a continuous embedding, tends to compute the output by ``matching'' the motif immediately prior to the forecasting start point to a motif in the earlier context, and then predicting by copying what comes immediately after. 
We identified that this ``parroting'' strategy (described in detail in~\citet{zhang2024zero,zhang2025context}) is mainly due to the angular bias of the model. 
Here, we provide three examples to clearly illustrate the idea and the mechanism of ``parroting,'' and we suggest when it may be ``good'' or ``bad.''

\begin{figure}[H]
    \centering
    
    \begin{minipage}{0.48\textwidth}
        \includegraphics[width=1\textwidth]{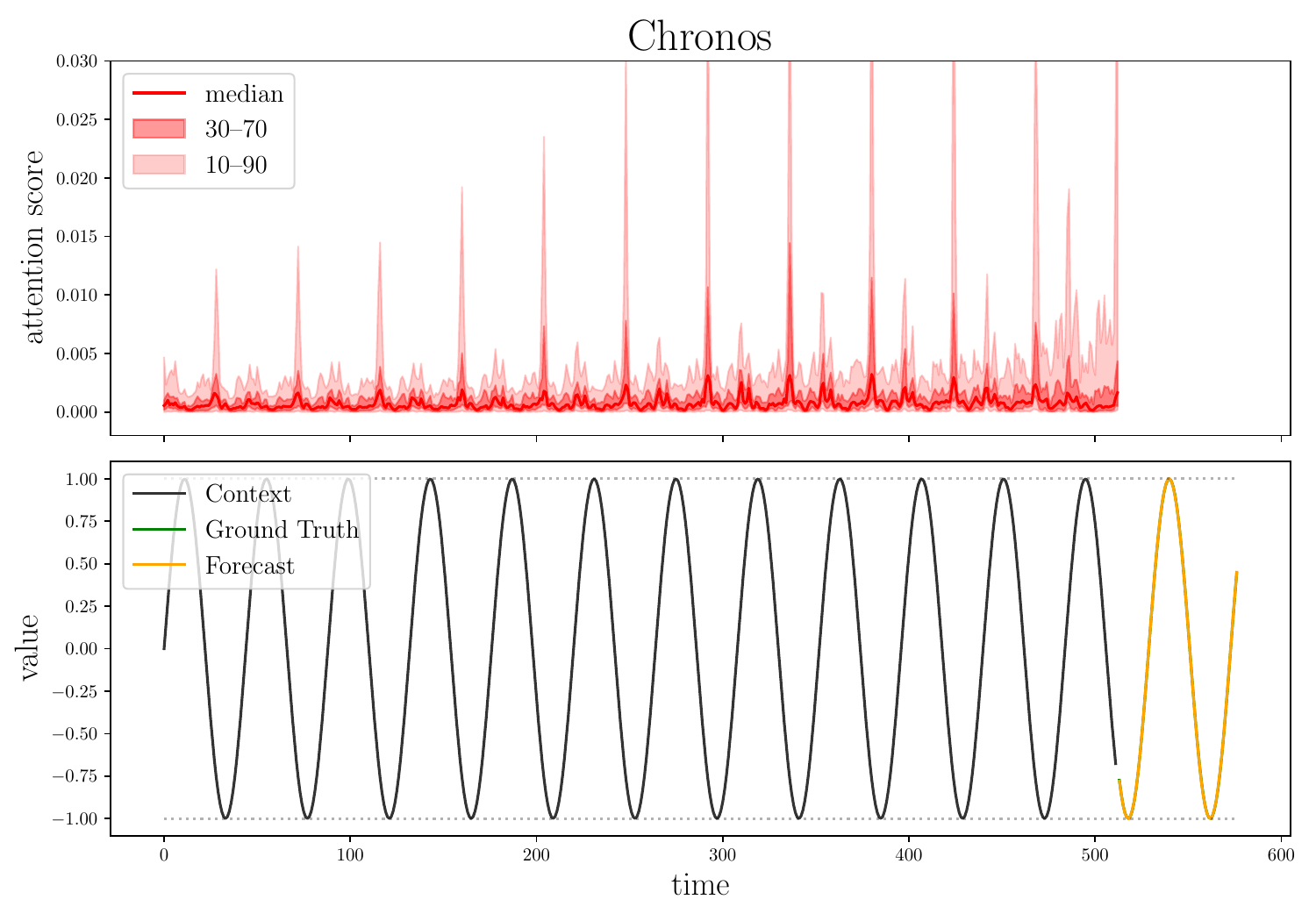}
    \end{minipage}
    \hfill
    \begin{minipage}{0.48\textwidth}
        \includegraphics[width=1\textwidth]{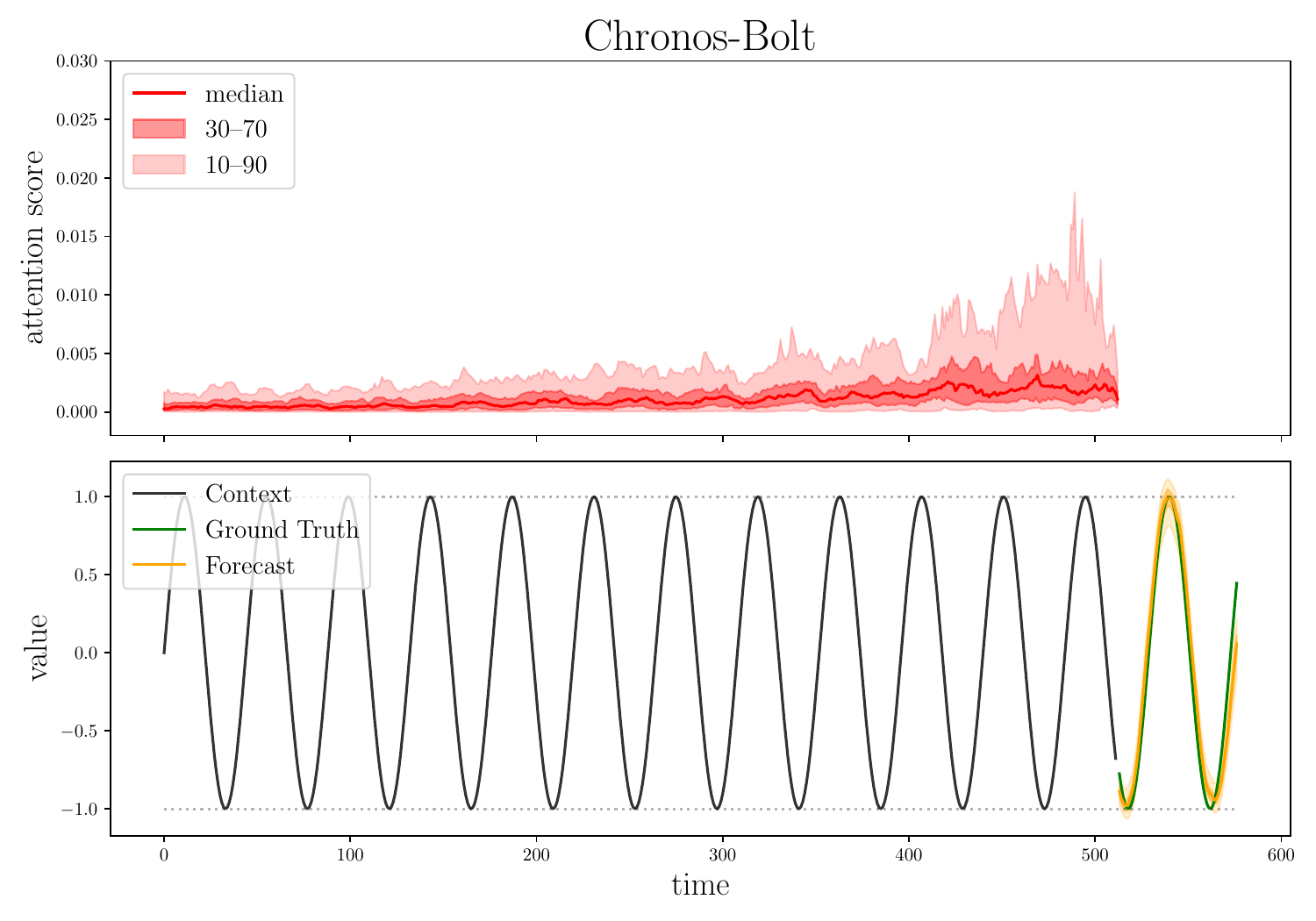}
    \end{minipage}

    \caption{We show an example where Chronos' parroting strategy makes it perform better than Chronos-Bolt on a perfectly periodic signal. Above the forecast plots, we show the density of all cross-attention scores in the decoder of the models. We see that Chronos' cross-attention scores are highly dense where the next forecast corresponds in the earlier periods, while Chronos-Bolt's cross-attention scores are more evenly distributed.}
    \label{fig:goodlocality}
\end{figure}

\textbf{Example I: Parroting helps Chronos to pinpoint a periodic context.} 
The example in~\Cref{fig:goodlocality} features a sinusoidal context wave. For this context, it is very easy for the Chronos model to parrot the context. In the top panels, we show the distribution of all cross-attention scores when Chronos and Chronos-Bolt (patch size $k=1$) make the first forecast. In Chronos, self-attention scores are put very heavily and locally near the position corresponding to the forecast position in every period, whereas Chronos-Bolt puts the cross-attention scores more evenly across the entire context. As a result, one cannot eyeball a difference between Chronos' forecast and the ground truth, but Chronos-Bolt does not achieve such a pinpoint prediction.

\textbf{Example II: Parroting makes Chronos lazy and prevents advanced ``reasoning.''}
In the next example, we still base our context on the same sinusoidal wave, but we scale it into a bidirectional envelope. This envelope is maximized in the middle and decays towards the two ends. The finesse of this design is that if a model naively performs parroting, then it is easy to match the last motif to a motif at the beginning, and predict the increasing context after it. By looking at the cross-attention scores, we note that this is exactly what Chronos does, and consequently, the signal it predicts has an increasing trend, while it should be decreasing. Chronos-Bolt, on the other hand, makes a relatively accurate prediction by attending mostly to the later context than the earlier ones to apply more advanced ``reasoning.''

\begin{figure}[H]
    \centering
    
    \begin{minipage}{0.48\textwidth}
        \includegraphics[width=1\textwidth]{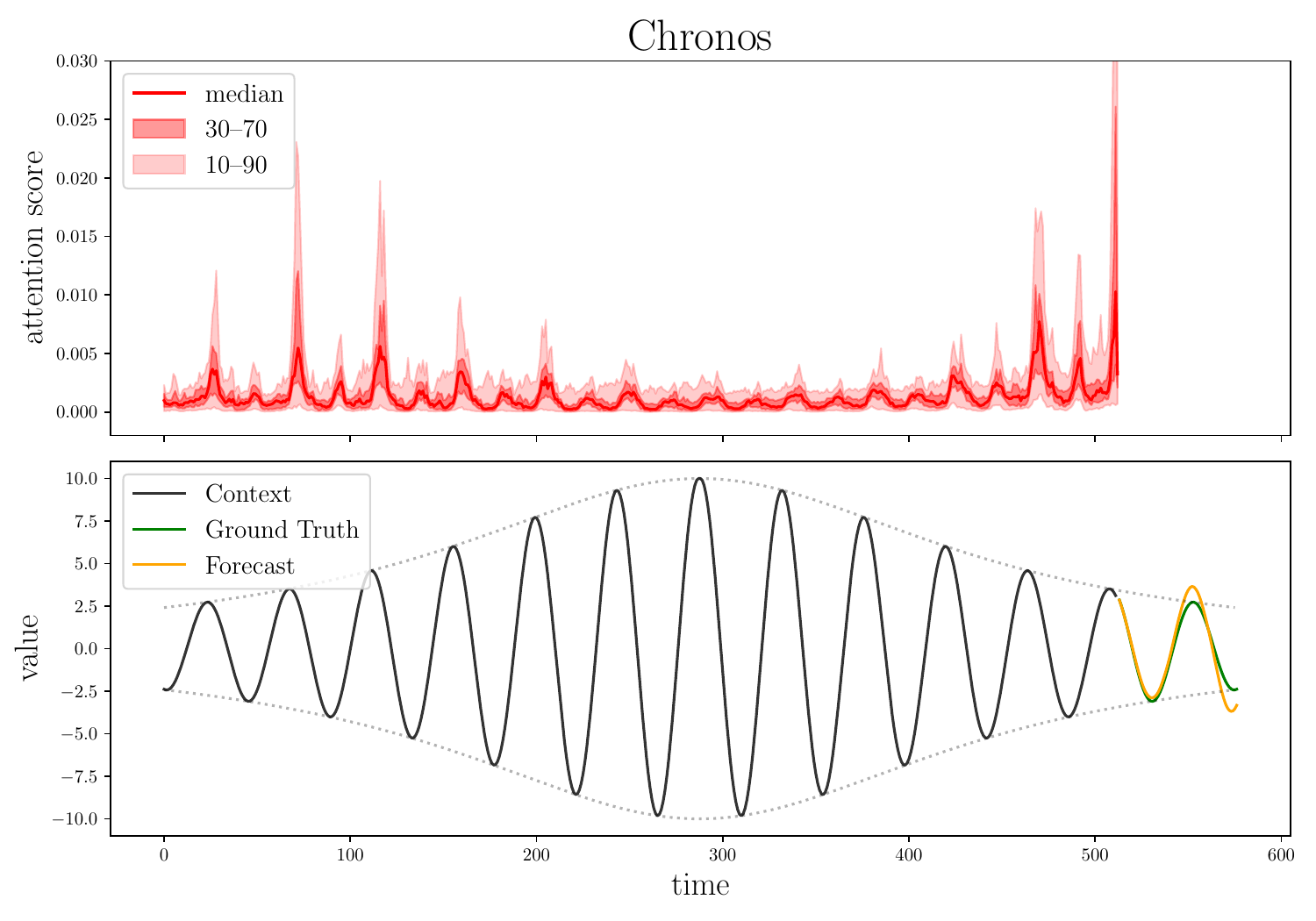}
    \end{minipage}
    \hfill
    \begin{minipage}{0.48\textwidth}
        \includegraphics[width=1\textwidth]{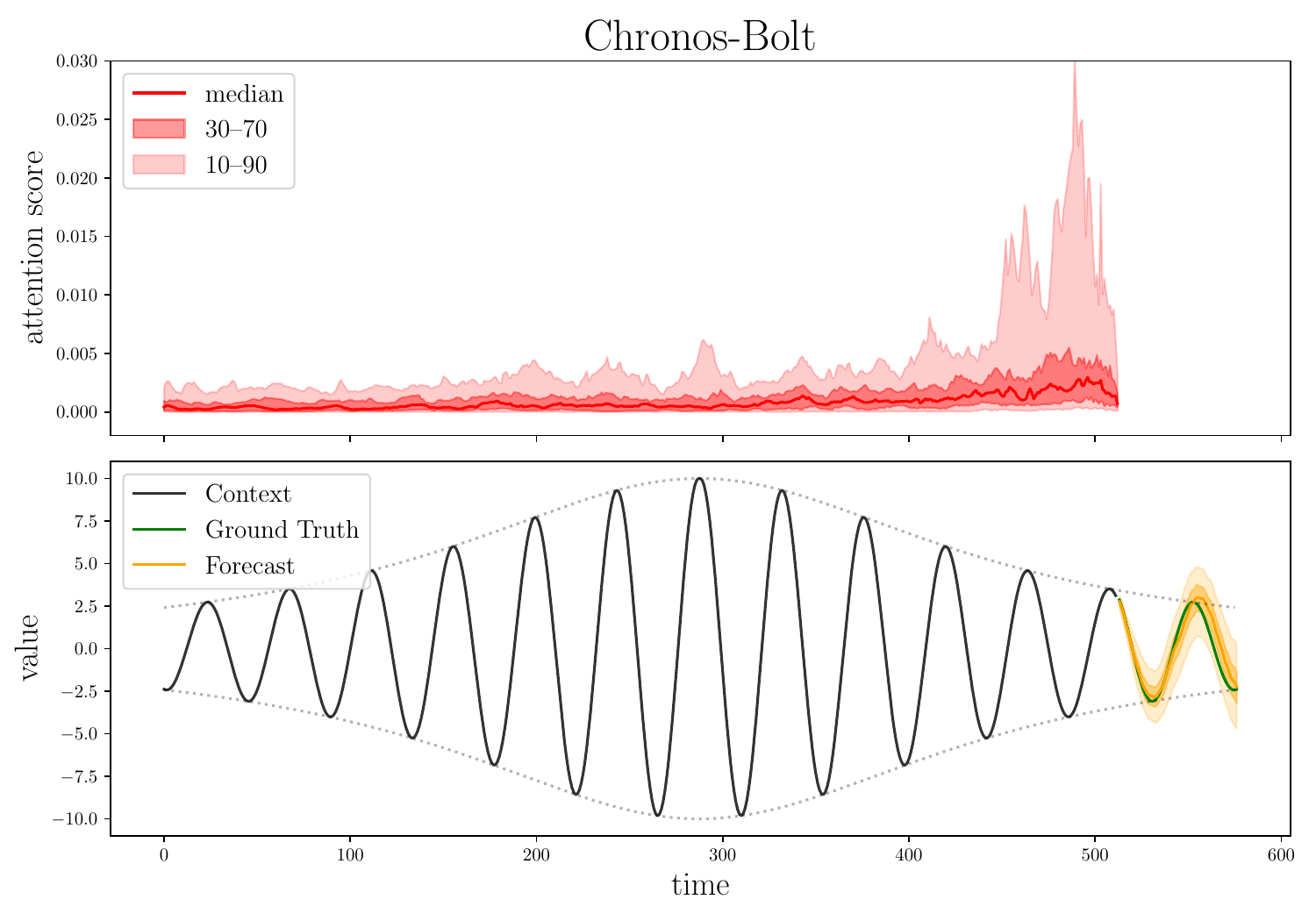}
    \end{minipage}

    \caption{We show an example where Chronos' parroting strategy harms its performance. From the cross-attention scores, we see that Chronos forecasts the future by parroting the earlier match motif, but unfortunately, the earlier motif exhibits an increasing trend, while the true trend in the forecast should clearly be decreasing. Chronos-Bolt performs much better by not parroting and mainly focusing on the recent history.}
    \label{fig:badlocality}
\end{figure}

\begin{figure}[H]
    \centering
    
    \begin{minipage}{0.48\textwidth}
        \includegraphics[width=1\textwidth]{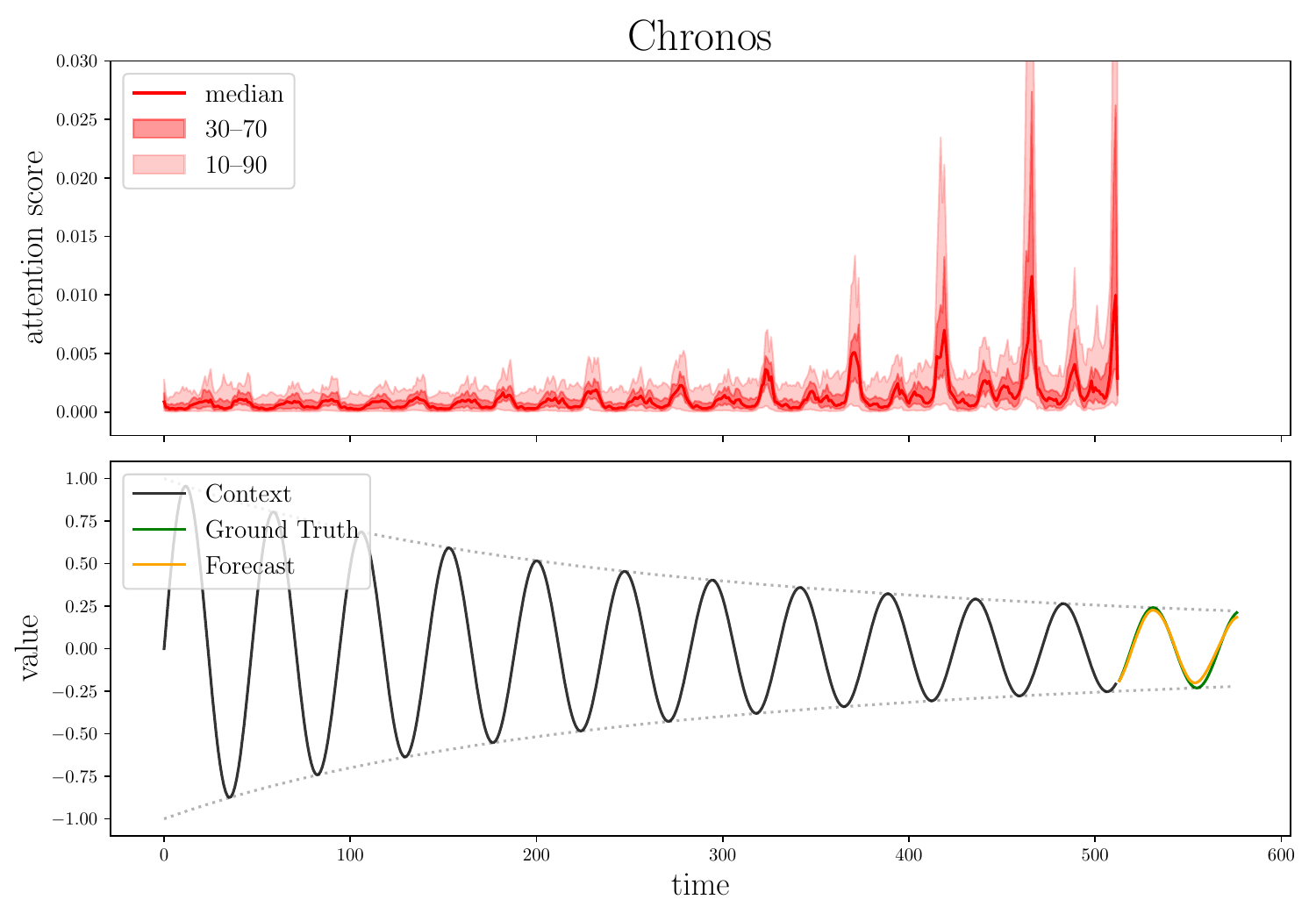}
    \end{minipage}
    \hfill
    \begin{minipage}{0.48\textwidth}
        \includegraphics[width=1\textwidth]{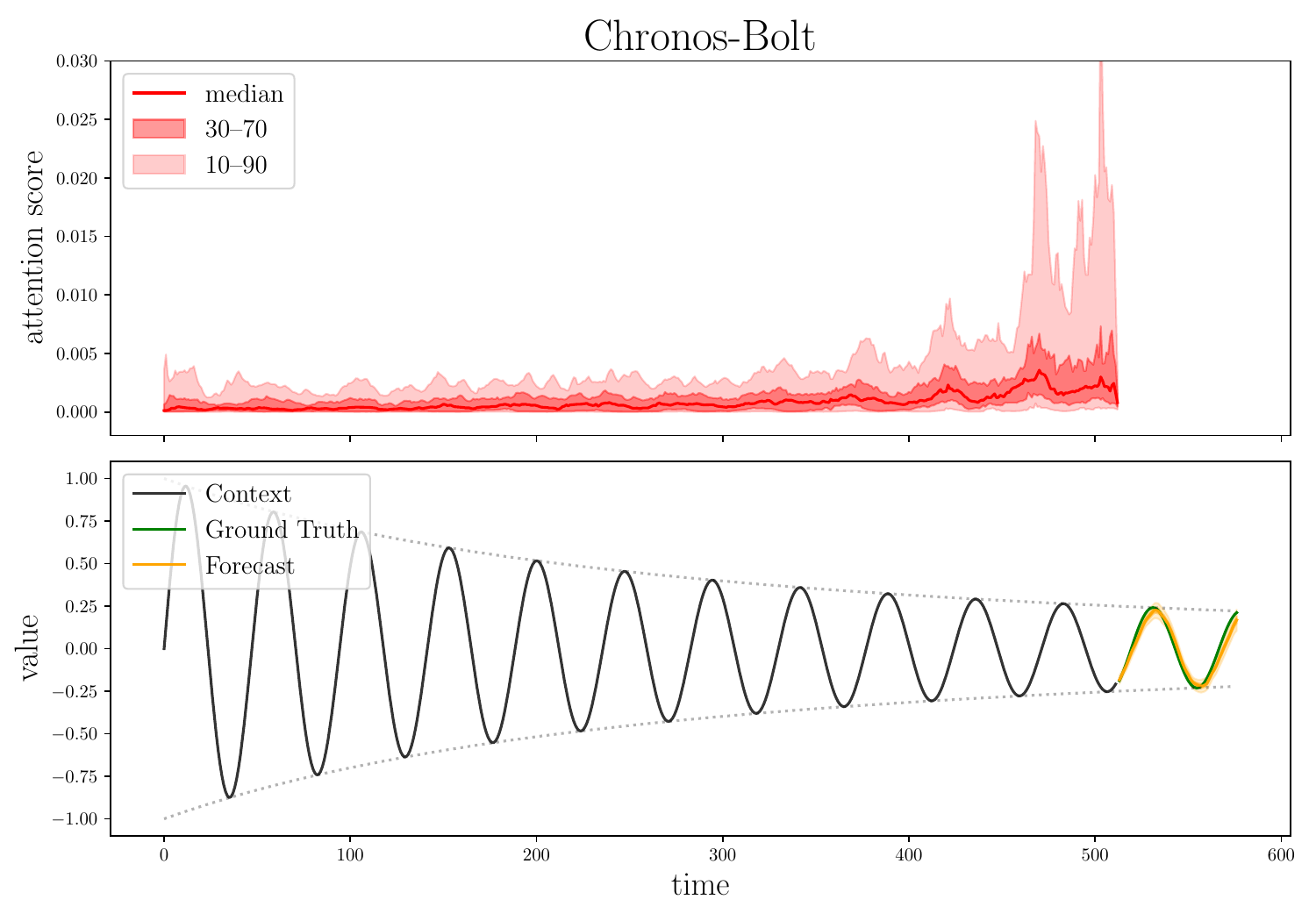}
    \end{minipage}

    \caption{We show an example where Chronos and Chronos-Bolt perform similarly. In this example, the scale of context is monotonically decreasing, making there little to parrot. Both models show some capabilities for ``reasoning'' about the decreasing scale of the signal.}
    \label{fig:samelocality}
\end{figure}

\textbf{Example III: Chronos can still ``reason'' if there is nothing to parrot.} 
Looking at the second example, one may wonder: is it that Chronos cannot perform advanced ``reasoning'' so that it gets the trend wrong? To test this, we redesign our envelope: instead of having a bidirectional envelope, we use a unidirectional one that is monotonically decreasing. The major distinction here is that we do not have an increasing period at the beginning of the context, which therefore prevents Chronos from parroting. In~\Cref{fig:samelocality}, we see that when Chronos has nothing to parrot, it attends to the recent history and gets the overall trend right; in this case, by simply ``removing'' the first decreasing part from our context, we can make Chronos perform as well as Chronos-Bolt again.

\subsection{Details of the Distance Bias}
\label{app:geomtric-distance}

\subsubsection{Design of Experiments in~\Cref{fig:scale}}

Recall~\Cref{fig:scale}, where we showed an experiment based on the real-world datasets that Chronos has a better capability of learning the fine-scale signals than Chronos-Bolt. 
We achieved this by augmenting Chronos' in-domain evaluation corpus, which contains $15$ popular real-world univariate time series benchmark datasets. Given an input context $\mathbf{x} \in \R^{L}$ and a target $\mathbf{y} \in \R^{T}$, we split $\mathbf{x} = (\mathbf{x}_1, \mathbf{x}_2)$ evenly into two parts $\mathbf{x}_1$ and $\mathbf{x}_2$. Then, we fix a scale ratio $\alpha \geq 1$, we consider two regimes:
\begin{enumerate}[leftmargin=*]
    \item Predicting the large motif: we scale $\mathbf{x}_1$ to $ \mathbf{x}_1 / \alpha$ and form the multi-scale context as $\tilde{\mathbf{x}}_{\text{large}} = (\mathbf{x}_1 / \alpha, \mathbf{x}_2)$. Let $\hat{\mathbf{y}} \in \R^{T}$ be a model's output, we compute the error by calculating the weighted quantile loss (WQL) using the ground truth $\mathbf{y}$ and forecast $\hat{\mathbf{y}}$.
    \item Predicting the small motif: we scale $\mathbf{x}_2$ to $ \mathbf{x}_2 / \alpha$ and form the multi-scale context as $\tilde{\mathbf{x}}_{\text{small}} = (\mathbf{x}_1, \mathbf{x}_2 / \alpha)$. Let $\hat{\mathbf{y}} \in \R^{T}$ be a model's output, we compute the error by calculating the WQL using the ground truth $\mathbf{y}$ and forecast $\alpha\hat{\mathbf{y}}$. Note that we need to scale the model's output by $\alpha$ to renormalize.
\end{enumerate}

To compute the overall loss, we follow Chronos' evaluation strategy~\citep{ansari2024chronos} by taking the geometric mean of the WQLs across all $15$ datasets measured by comparing a model to a baseline, which is taken to be the model's performance when $\alpha = 1$, i.e., no augmentation is done. Note also that when $\alpha = 1$, predicting the large motif collapses to predicting the small motif.

\subsubsection{How Does a Transformer Process a Multiscale Signal?}

\begin{figure}[H]
    \centering
    
    \begin{minipage}{0.48\textwidth}
        \includegraphics[width=1\textwidth]{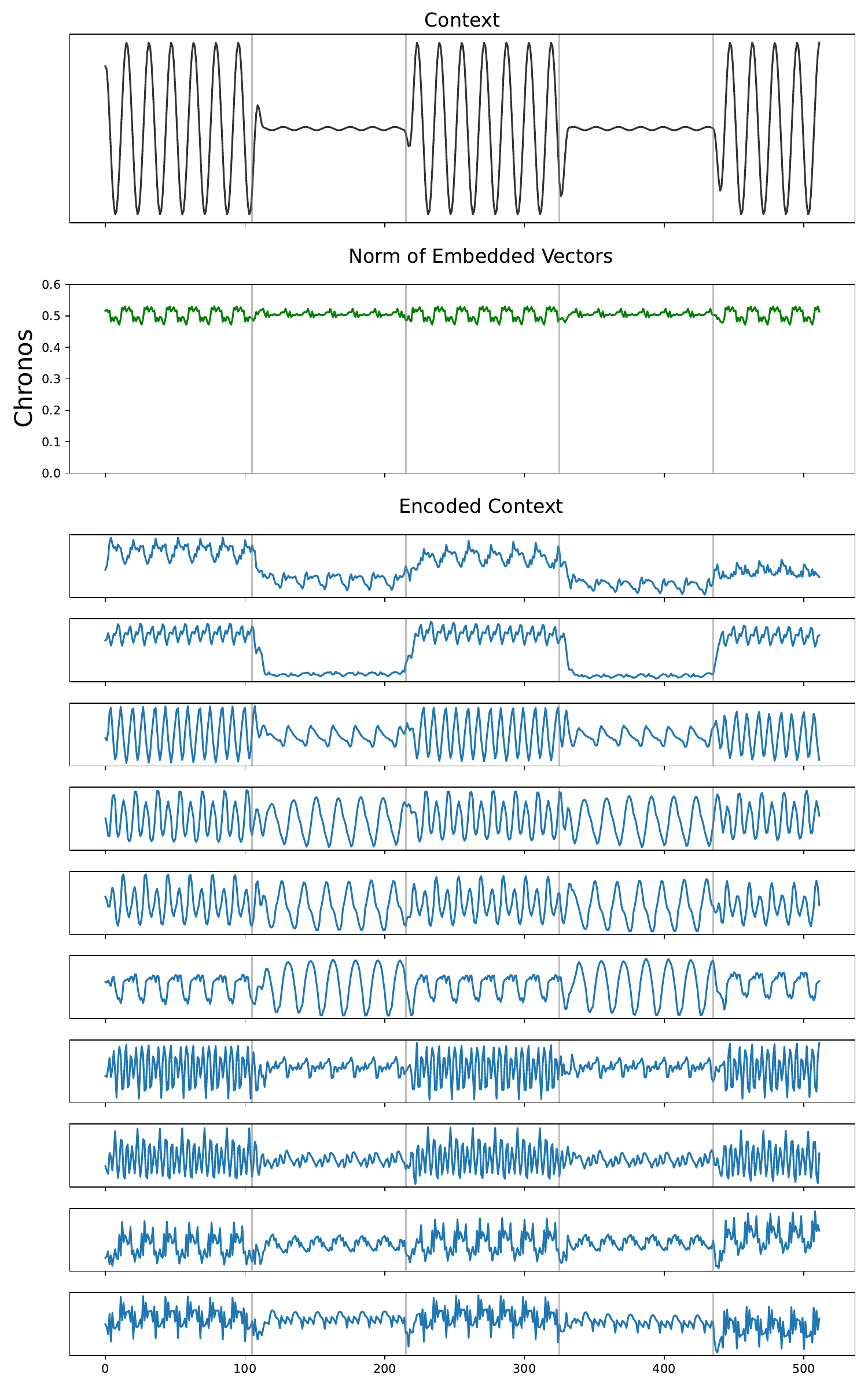}
    \end{minipage}
    \hfill
    \begin{minipage}{0.48\textwidth}
        \includegraphics[width=1\textwidth]{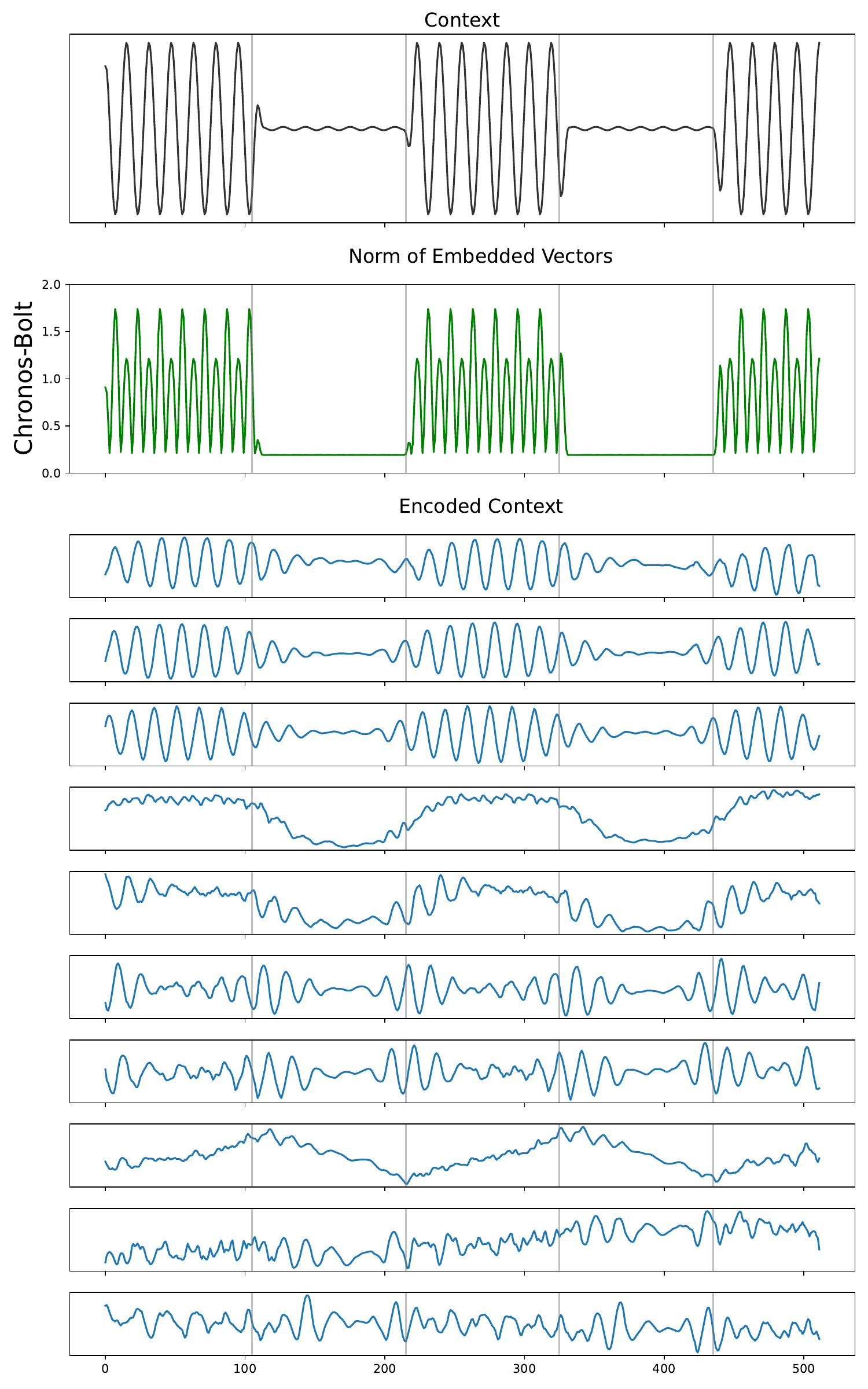}
    \end{minipage}

    \caption{We show how each of Chronos and Chronos-Bolt (patch size $k=1$) processes a multi-scale context, respectively. In the top panels, we show our input multi-scale context. The green curves on the second row of panels show the norm of the embedded vectors in each model, where we see that the fine-scale and the large-scale motifs both become large-scale in the embedded context, whereas the difference between the two scales stays distinguishable, if not more dramatic, in Chronos-Bolt's embedding. Then, we show the encoded context in both models, projected onto the top-$10$ principal components. We see that Chronos' fine-scale motifs got preserved nicely, while the fine scales are contaminated by the large scales in Chronos-Bolt.}
    \label{fig:detailedmultiscale}
\end{figure}

To illustrate the distance bias of Chronos versus Chronos-Bolt better, we perform a detailed analysis given a multiscale context shown in~\Cref{fig:detailedmultiscale}. First, given the context $\mathbf{x} \in \R^L$, we let the embedding be $\Phi(\mathbf{x}) = (\boldsymbol{\phi}_1, \ldots, \boldsymbol{\phi}_L) \in \R^{d \times L}$, where $d$ is the hidden dimension of the Transformer. In the panels on the second row of~\Cref{fig:detailedmultiscale}, we show the norm $\|\boldsymbol{\phi}_j\|_2$ of these embedded vectors, which indicates a clear multiscale structure in Chronos-Bolt's embedding, but not as much in Chronos' embedding. This is the root cause of the distance bias. To show that the embedding really matters, we further take the encoded context $\mathbf{z} \in \R^{d \times L}$, which is the sequence out of the encoder Transformer, and we plot $\mathbf{z}$ in its leading $10$ principal components along the temporal direction. From~\Cref{fig:detailedmultiscale}, we see that the fine-scale oscillation is preserved very well by Chronos' encoder, but it is severely contaminated by the large-scale oscillation in Chronos-Bolt's encoder.

\subsection{Details of the Norm Bias}
\label{app:geomtric-norm}

\subsubsection{Design of Experiments in~\Cref{fig:offset}}

Recall~\Cref{fig:offset}, where we showed an experiment based on the real-world datasets that Chronos has a better capability of learning the zero-offset signals than Chronos-Bolt. We achieved this by augmenting Chronos' in-domain evaluation corpus, which contains $15$ popular real-world univariate time series benchmark datasets. Given an input context $\mathbf{x} \in \R^{L}$ and a target $\mathbf{y} \in \R^{T}$, we split $\mathbf{x} = (\mathbf{x}_1, \mathbf{x}_2, \mathbf{x}_3)$ evenly into three parts $\mathbf{x}_1$, $\mathbf{x}_2$, and $\mathbf{x}_3$. Then, fix a offset $\beta \geq 1$, we consider two regimes:
\begin{enumerate}[leftmargin=*]
    \item Predicting the high-offset motif: we drop $\mathbf{x}_2$ to $ \mathbf{x}_2 - \beta$, lift $\mathbf{x}_3$ to $\mathbf{x}_3 + \beta$, and form the multi-scale context as $\tilde{\mathbf{x}}_{\text{high}} = (\mathbf{x}_1, \mathbf{x}_2-\beta, \mathbf{x}_3+\beta)$. Let $\hat{\mathbf{y}} \in \R^{T}$ be a model's output, we compute the error by calculating the WQL using the ground truth $\mathbf{y}$ and forecast $\hat{\mathbf{y}}-\beta$. Note that we need to drop the model's output back by $\beta$ to renormalize.
    \item Predicting the low-offset motif: we drop $\mathbf{x}_1$ to $ \mathbf{x}_1 - \beta$, lift $\mathbf{x}_2$ to $\mathbf{x}_2 + \beta$, and form the multi-scale context as $\tilde{\mathbf{x}}_{\text{low}} = (\mathbf{x}_1-\beta, \mathbf{x}_2+\beta, \mathbf{x}_3)$. Let $\hat{\mathbf{y}} \in \R^{T}$ be a model's output, we compute the error by calculating the WQL using the ground truth $\mathbf{y}$ and forecast $\hat{\mathbf{y}}$.
\end{enumerate}

To compute the overall loss, we follow Chronos' evaluation strategy~\citep{ansari2024chronos} by taking the geometric mean of the WQLs across all $15$ datasets measured by comparing a model to a baseline, which is taken to be the model's performance when $\beta = 0$, i.e., no augmentation is done. Note also that when $\beta = 0$, predicting the large motif collapses to predicting the small motif.

\subsubsection{How Does a Transformer Process a Multi-Offset Signal?}

\begin{figure}[H]
    \centering
    
    \begin{minipage}{0.48\textwidth}
        \includegraphics[width=1\textwidth]{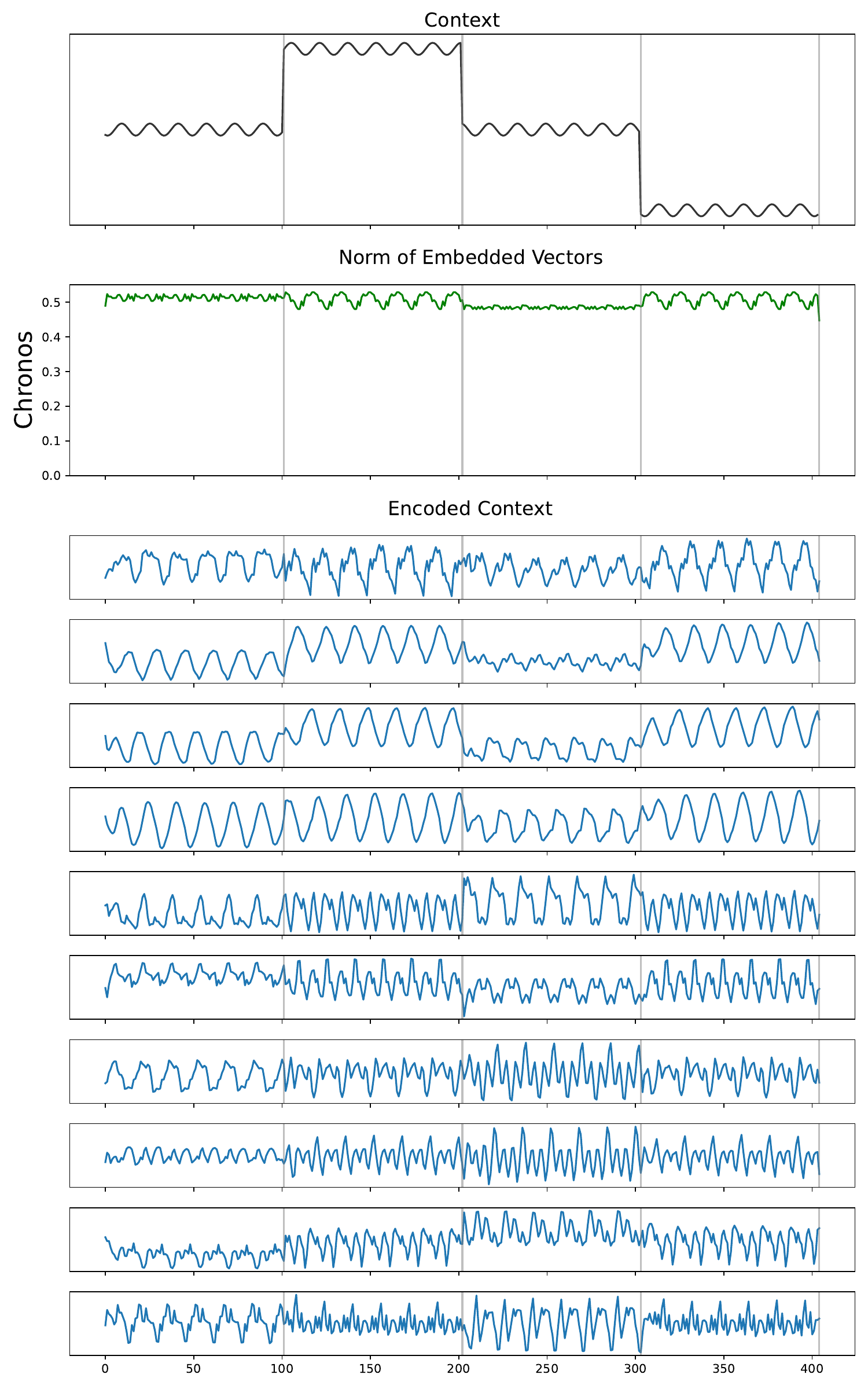}
    \end{minipage}
    \hfill
    \begin{minipage}{0.48\textwidth}
        \includegraphics[width=1\textwidth]{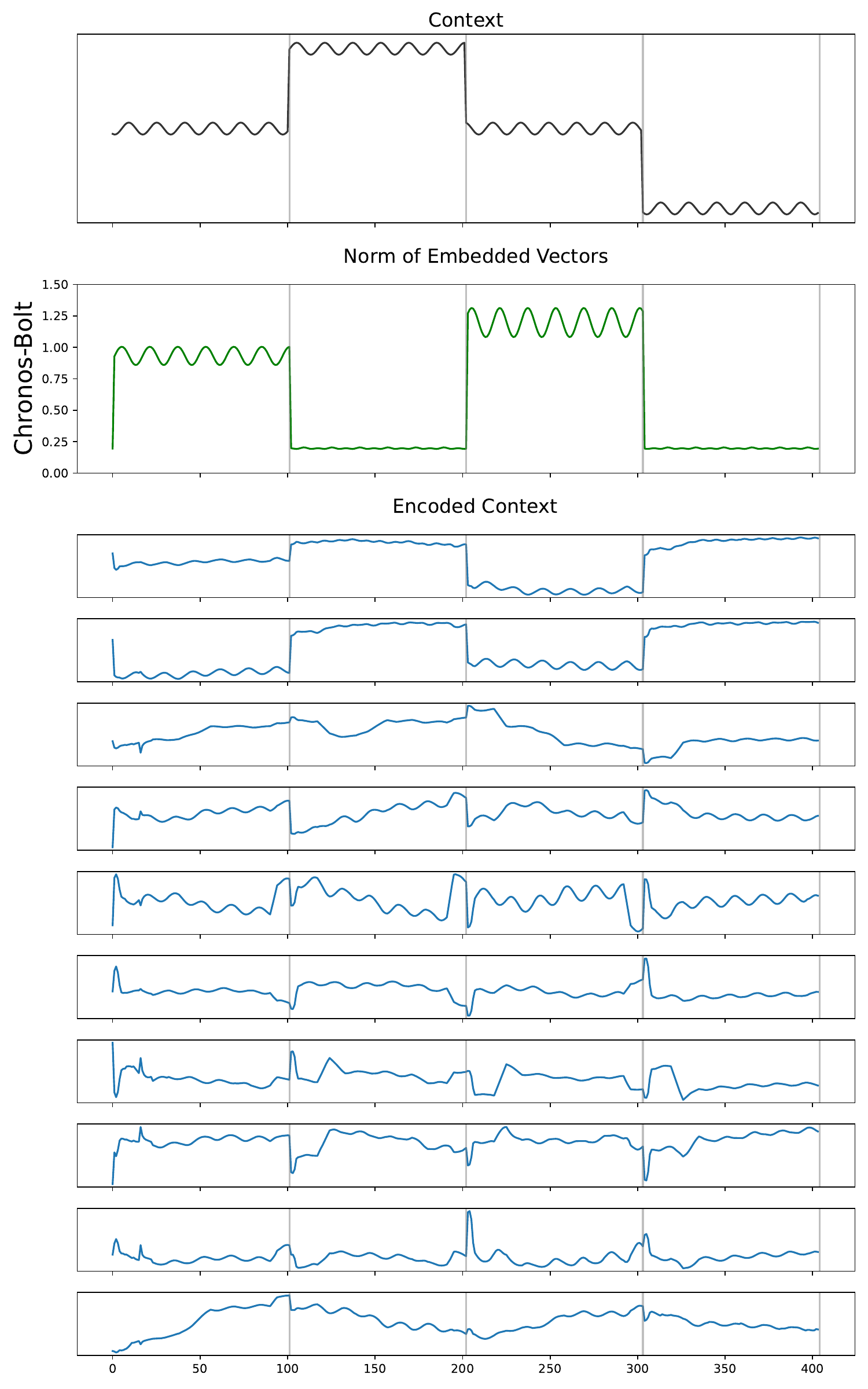}
    \end{minipage}

    \caption{We show how each of Chronos and Chronos-Bolt (patch size $k=1$) processes a multi-offset context, respectively. In the top panels, we show our input multi-offset context. The green curves on the second row of panels show the norm of the embedded vectors in each model, where we see that the high-offset and the low-offset motifs both have large norms in the embedded context, whereas in Chronos-Bolt's embedding, the low-offset motifs have much smaller norms than the high-offset ones. Then, we show the encoded context in both models, projected onto the top-$10$ principal components. We see that Chronos' low-offset motifs got preserved nicely, while they are contaminated by the high-offset motifs in Chronos-Bolt.}
    \label{fig:detailedmultioffset}
\end{figure}

To better illustrate the norm bias of Chronos versus Chronos-Bolt, we perform a detailed analysis given a multiscale context shown in~\Cref{fig:detailedmultioffset}. First, given the context $\mathbf{x} \in \R^L$, we let the embedding be $\Phi(\mathbf{x}) = (\boldsymbol{\phi}_1, \ldots, \boldsymbol{\phi}_L) \in \R^{d \times L}$, where $d$ is the hidden dimension of the Transformer. In the panels on the second row of~\Cref{fig:detailedmultioffset}, we show the norm $\|\boldsymbol{\phi}_j\|_2$ of these embedded vectors, where we see that Chronos' zero-offset motifs are embedded onto large vectors but Chronos-Bolt's are embedded onto small vectors. This is the root cause of the norm bias. To show that the embedding really matters, we further take the encoded context $\mathbf{z} \in \R^{d \times L}$, which is the sequence out of the encoder Transformer, and we plot $\mathbf{z}$ in its leading $10$ principal components along the temporal direction. From~\Cref{fig:detailedmultioffset}, we see that the zero-offset oscillation is preserved very well by Chronos' encoder, but it is severely contaminated by the large-offset oscillation in Chronos-Bolt's encoder.

\clearpage

\section{Details of the Regression-to-the-Mean Bias (from \Cref{sec:regression})}
\label{app:regression}

In this section, we provide more details on the regression-to-the-mean bias.
First, we use two examples to illustrate the loss landscape induced by the three standard loss functions: $L^1$-based, $L^2$-based, and cross-entropy (\Cref{app:regression-loss-landscape}). 
Then, we revisit the examples shown in~\Cref{fig:regression,}, and we show how Chronos learns the underlying probability distribution, instead of regressing to the mean (\Cref{app:regression-bifurcation}). 
Finally, we give the details about how we build a ``bridge'' from the deterministic case to the random case in~\Cref{fig:regression} (\Cref{app:regression-injection}).

\subsection{A Loss-Landscape Illustration of Three Kinds of Losses}
\label{app:regression-loss-landscape}

\begin{figure}[H]
    \centering
    \includegraphics[width=1\linewidth]{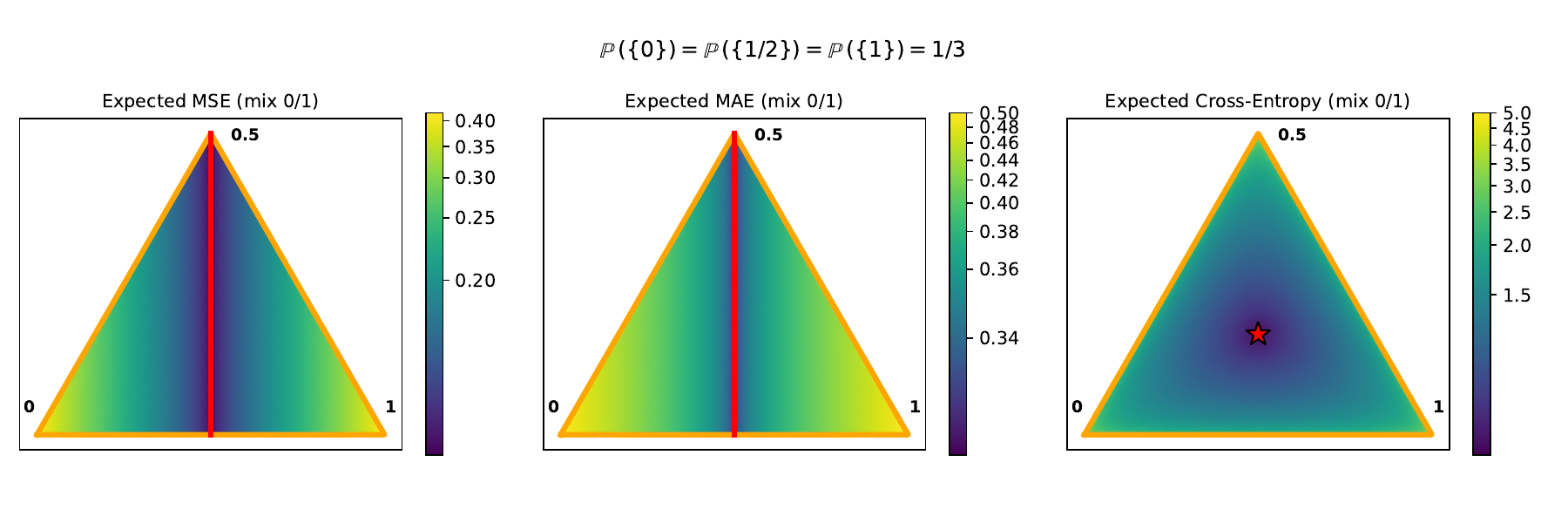}
    \caption{Assume we have a ground-truth probability distribution that equals $1/3$ on $0$, $1/2$, and $1$, respectively, and suppose our model generates a probability distribution on the set $\{0,1/2,1\}$. We show the loss landscape induced by the ground-truth probability and the model's forecast. The colors indicate the loss, and the barycentric coordinate in each triangle represents the model's predicted probability on $0$, $1/2$, and $1$, respectively. We use red colors to indicate the global minima in the loss landscape induced by each norm.}
    \label{fig:loss_trimodal}
\end{figure}

To illustrate the concept of regression-to-the-mean/median, we consider the case where we have a ground-truth probability distribution on a simple measurable set $\Omega = \{0,1/2,1\}$ equipped with the discrete $\sigma$-algebra. Assume first that the ground-truth probability distribution is the one with
\[
    \mathbb{P}_{\text{truth}}(\{0\}) = \mathbb{P}_{\text{truth}}(\{1/2\}) = \mathbb{P}_{\text{truth}}(\{1\}) = \frac{1}{3}.
\]
In each triangle in~\Cref{fig:loss_trimodal}, we show the barycentric coordinate corresponding to a model's output probability distribution on $\Omega$. The three different losses are shown using colors.

Note that the MSE and MAE are independent of the probability distribution: they only require a final prediction. In that sense, one can imagine that the entire triangle collapses to the line segment from $0$ to $1$ via orthogonal projection. That is essentially why our final forecast regresses to $1/2$, which is both the mean and median of $\mathbb{P}_{\text{truth}}$ in this example. For cross-entropy, the optimal is at the barycenter of the triangle, corresponding to exactly the ground-truth distribution.

\begin{figure}[H]
    \centering
    \includegraphics[width=1\linewidth]{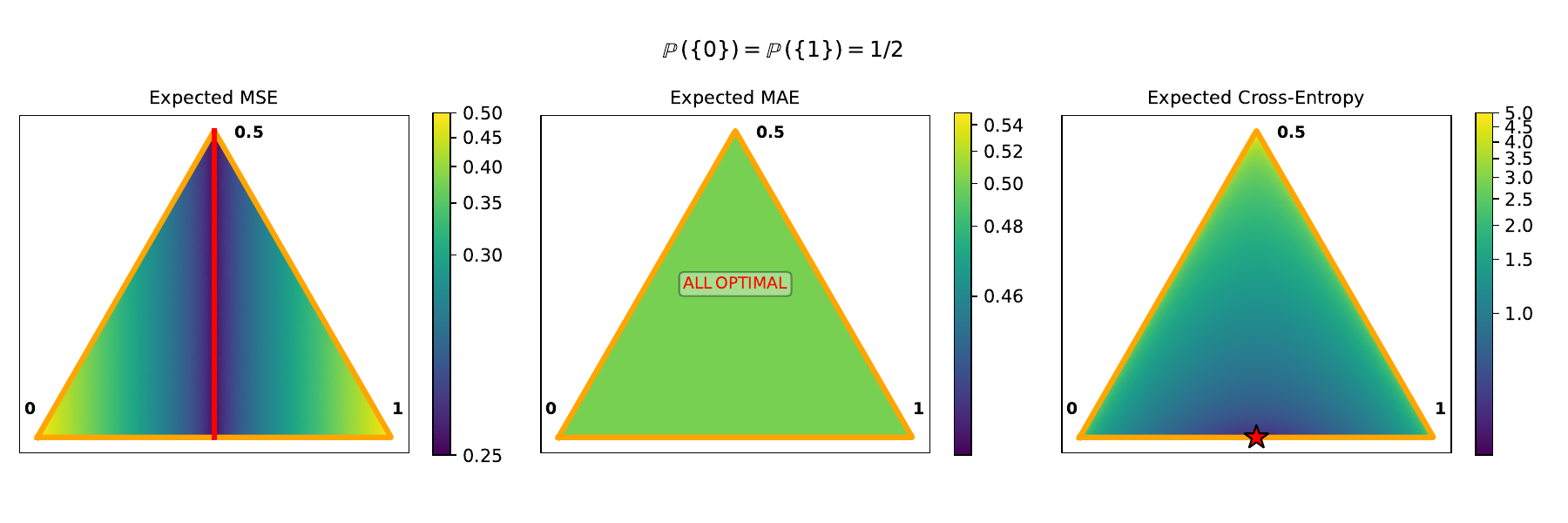}
    \caption{The plots show the same things as in~\Cref{fig:loss_trimodal}, except that we now set the ground-truth probability distribution to be a bi-modal one with a probability of $1/2$ on $0$ and $1$, respectively.}
    \label{fig:loss_bimodal}
\end{figure}

Next, we consider a slightly different ground-truth probability distribution:
\[
    \mathbb{P}_{\text{truth}}(\{0\}) = \mathbb{P}_{\text{truth}}(\{1\}) = \frac{1}{2}, \qquad \mathbb{P}_{\text{truth}}(\{1/2\}) = 0.
\]
From~\Cref{fig:loss_bimodal}, we see that MSE still regresses to the mean, which is the same as in the previous example; the cross-entropy loss now eliminates $1/2$ from its support, injecting more flexibility. The MAE loss, which tailors the median, gives us a completely flat loss landscape, where every distribution is an optimum.

\begin{figure}[H]
    \centering
    \includegraphics[width=0.66\linewidth]{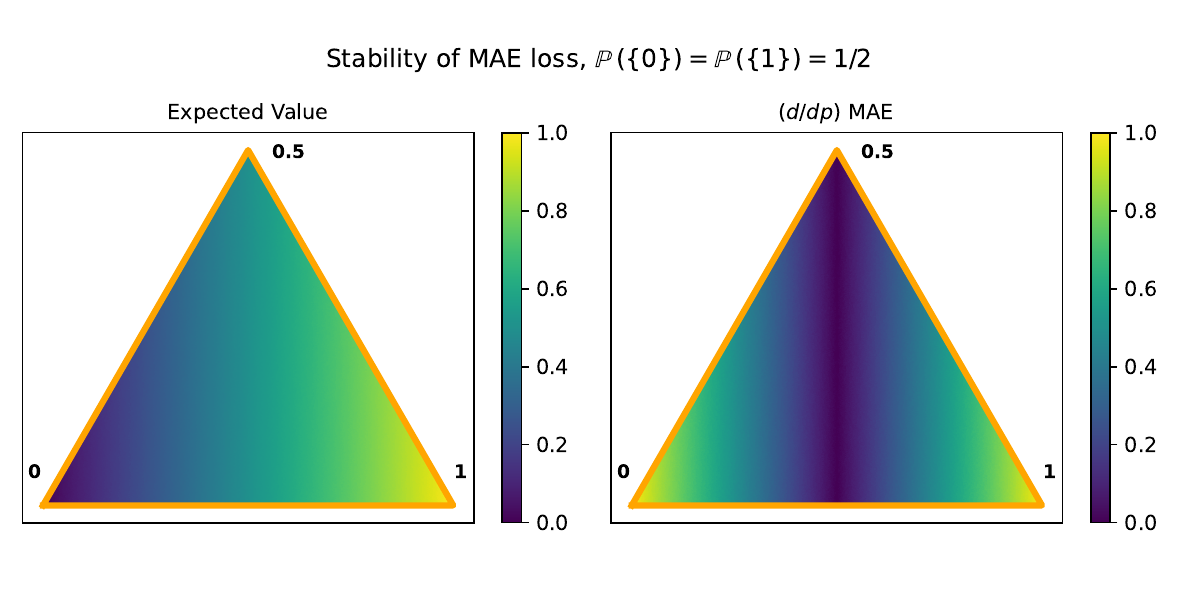}
    \caption{On the left, we show the expected value of the model's forecast in the barycentric coordinate~\citep{schumaker2022approximation}. On the right, we show the stability of the MAE loss given the ground-truth in~\Cref{fig:loss_bimodal}, where the colors indicate the magnitude of the gradient $|(d/dp)\text{MAE}(\mathbb{P}_{\text{truth}}(p), \mathbb{P}_{\text{model}})|$ at $p=1/2$, where $\mathbb{P}_{\text{truth}}(p)$ is $p$ at $0$ and $1-p$ at $1$.}
    \label{fig:loss_stability}
\end{figure}

The only question that remains is: if all points are optimal against an MAE loss, then why is it that empirically, the model usually prefers one closer to the mean (see~\Cref{fig:regression})? One answer is: although every number between $[0,1]$ is a median of this probability distribution, the medians have different stability. That is, if we compute the gradient of the MAE \textit{with respect to} the probability $p$ at $p = 2$, assuming that
\[
    \mathbb{P}_{\text{truth}}(\{0\}) = p, \qquad \mathbb{P}_{\text{truth}}(\{1\}) = 1-p, \qquad \mathbb{P}_{\text{truth}}(\{1/2\}) = 0,
\]
then this gradient is large near the two endpoints $0$ and $1$ and small in the middle (see~\Cref{fig:loss_stability}). Why does that matter? In practice, even if the probability distribution is $1/2$ on both $0$ and $1$, when we sample the data, they will sometimes ``lean'' towards $0$ and sometimes towards $1$. Given that, ``the median of the medians'' gives us a more robust prediction.

\subsection{How Confident Chronos is at Bifurcation?}
\label{app:regression-bifurcation}

In~\Cref{fig:regression}, we saw that when the context consists of only $0$'s and $1$'s, Chronos, in its forecast, only generates these two numbers. Since Chronos performs regression via classification, we can understand this better by looking into the logits of Chronos' forecast. 
Here, we look into these logits when Chronos forecasts the two contexts --- one deterministic and one random --- in~\Cref{fig:regression}. 
Recall that the Chronos generates the output via binning: it preassumes a fine grid $\mathcal{B} = \{B_1, B_2, \ldots, B_V\}$ on the real-line and computes a probability distribution over $\mathcal{B}$. 
Here, when Chronos makes a forecast, we keep track of its generated probability associated with each $B_j$. 
In particular, there are two bins $B_{j_0}$ and $B_{j_1}$ whose values correspond to (the nearest grid point from) $0$ and $1$, respectively.

\begin{figure}[H]
    \centering
    \begin{minipage}{0.48\textwidth}
        \includegraphics[width=1\textwidth]{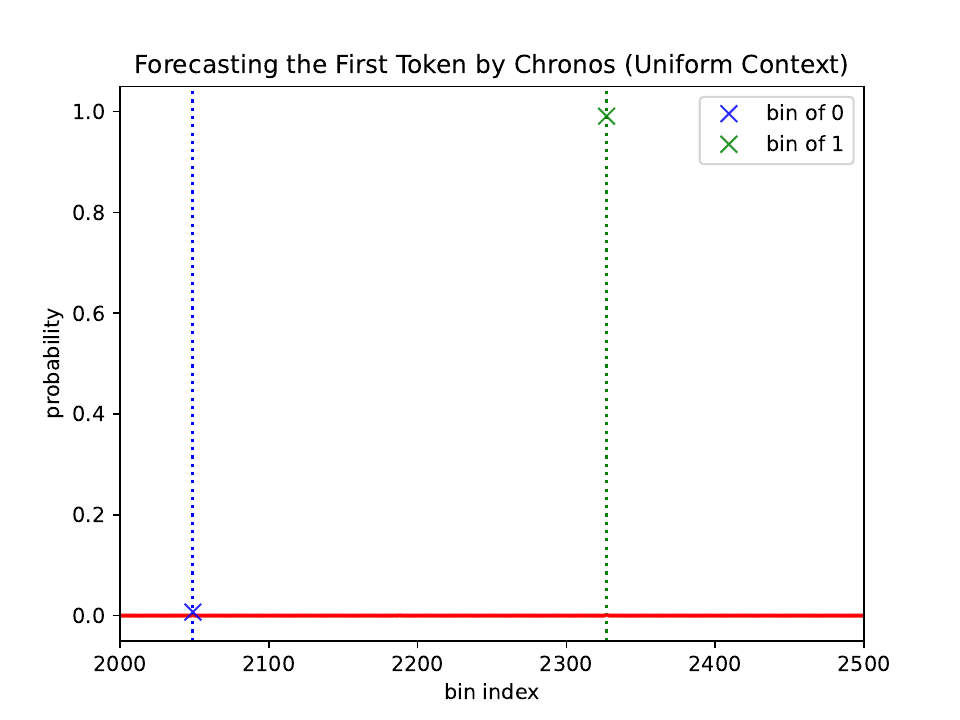}
    \end{minipage}
    \hfill
    \begin{minipage}{0.48\textwidth}
        \includegraphics[width=1\textwidth]{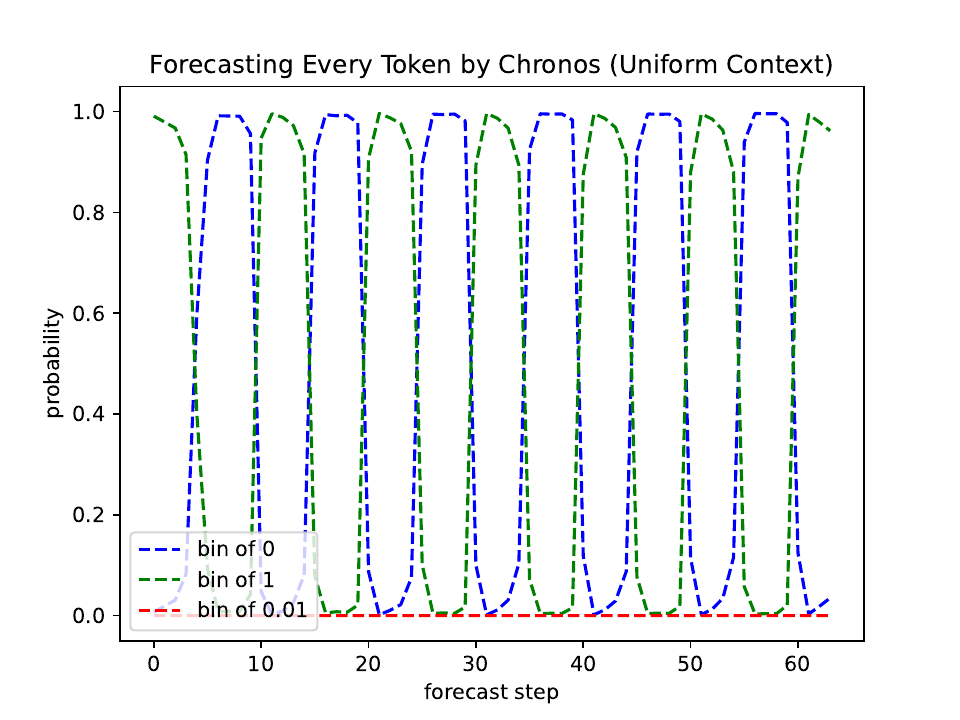}
    \end{minipage}

    \begin{minipage}{0.48\textwidth}
        \includegraphics[width=1\textwidth]{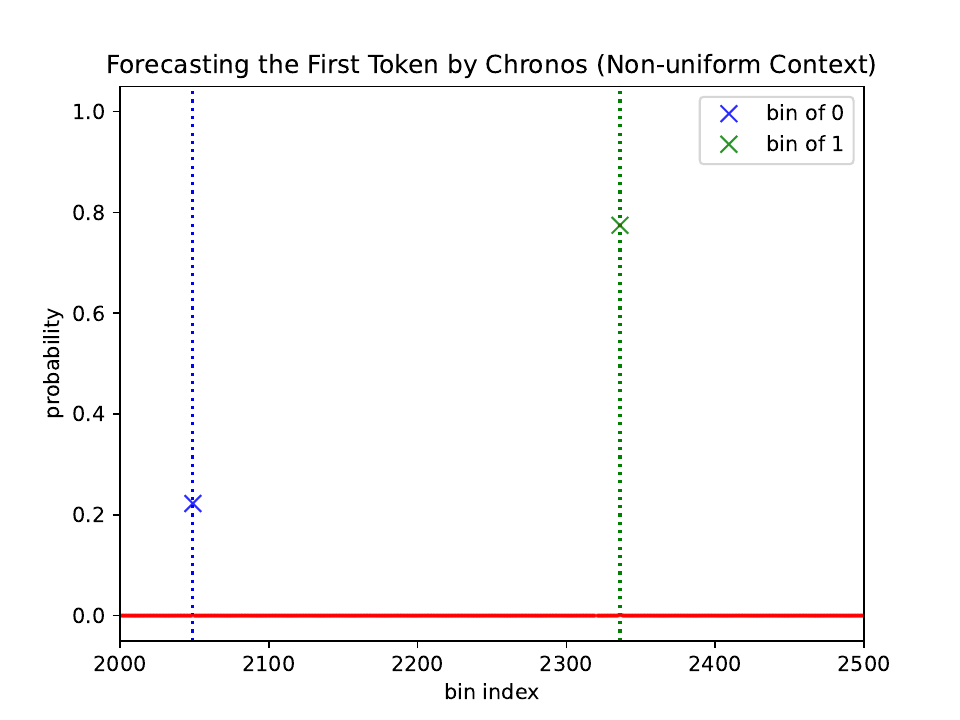}
    \end{minipage}
    \hfill
    \begin{minipage}{0.48\textwidth}
        \includegraphics[width=1\textwidth]{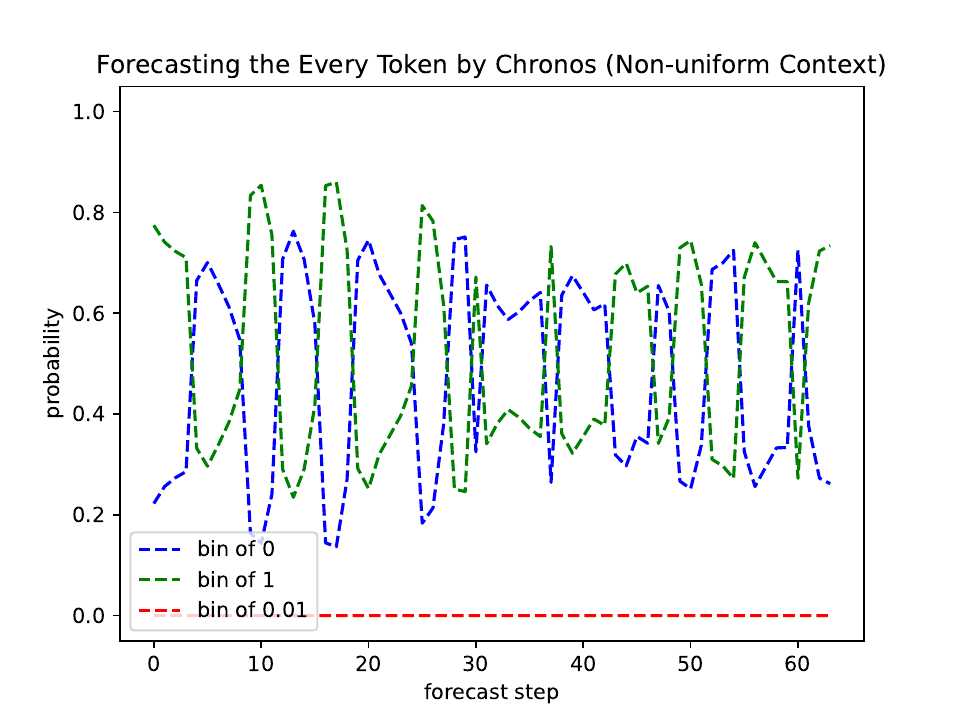}
    \end{minipage}
    
    \caption{We compute the logits when we apply a Chronos model to predict the two synthetic signals shown in~\Cref{fig:regression}. The top two panels show the results corresponding to the deterministic periodic walk between $0$ and $1$, while the bottom two panels show the results corresponding to a random walk. On the left, we show the logits associated with the first prediction. Each red dot corresponds to the logit associated with one bin, and the logits associated with the bin of ``$0$'' and the bin of ``$1$'' are highlighted with a blue and a green cross-mark, respectively. We see that the probabilities are dense on $\{0,1\}$ and tiny elsewhere. On the right, we show the logits corresponding to the bins of ``$0$'' and ``$1$'', respectively, as we make more forecast steps. As a reference, we also show the logits corresponding to the bin of ``$0.01$'', which are nearly zero.}
    \label{fig:logits_twocontexts}
\end{figure}

From~\Cref{fig:logits_twocontexts}, we see that Chronos is almost always very certain in that its forecast is either $0$ or $1$. That is, the only two non-trivial probabilities that it produces are at $0$ and $1$ --- all remaining probabilities are almost zero, even for the bins that are immediately next to $0$ or $1$. Incidentally, we should note that the fact that Chronos is capable of distinguishing $0$ and $1$ from their nearest bins is also due to the locality and distance bias (see~\cref{sec:geometry}). The only major difference between forecasting the deterministic and the random contexts is that Chronos is much more confident in which one, between $0$ and $1$, to pick when the context is deterministic, whereas given a random context, Chronos generates a close probability on $0$ and on $1$. This is not surprising, as it aligns with our discussion that the cross-entropy loss tailors a model to learn the ``ground-truth'' probability distribution.

\subsection{How to Inject Uncertainty in~\Cref{fig:regression}?}
\label{app:regression-injection}

In~\Cref{fig:regression}, we show an interesting quantification, where we start from a totally deterministic case, where we walk periodically between the two branches $0$ and $1$, and end up with the totally deterministic case, where for each time step, we randomly land at $0$ or $1$ with an even probability. Here, we show the details of how to gradually inject uncertainty and make a smooth transition from the deterministic case to the random case.

To this end, we first define the periodic context. Here, we alternately take $5$ steps of each branch. That is, our context is given by
\[
    x_t = \mathbbm{1}_{[0,5) \cup [10,15) \cup [20,25) \cup \cdots}.
\]
Given a probability of diffusion, we simply add a random ``XOR'' perturbation to this context with probability $1-p$. That is, the new context is given by
\[
    \hat{x}_t = x_t \oplus \text{Bernoulli}(1-p),
\]
where the Bernoulli distribution at every $t$ is i.i.d.. It is clear that $\mathbb{P}(\hat{x}_t)$ is $p$ on one branch of $0$ and $1$ and $1-p$ on the other, and that when we increase $p$ to $1/2$, our context becomes a random walk.

To understand how much the model regresses to $1/2$, the mean or the median of the entire sequence, we define the ``regression score'' by
\[
    \text{reg\_score}(y) = \min(|1-y|,|y|).
\]
This is shown in~\Cref{fig:regression}, where our experiments are repeated $100$ times and the error bars show the $30\%$ to $70\%$ quantiles of all regression scores.

\clearpage

\section{The Simplicity Bias}
\label{app:simplicity}

The term ``simplicity bias'' refers to the tendency of neural networks to favor simpler hypotheses, even when more complex, and potentially more robust, solutions are equally compatible with the data.  
This bias has roots in classical learning theory, which uses concepts like VC dimension and Rademacher complexity to explain how restricting effective hypothesis complexity enables generalization~\citep{wolpert2002no}.  
Empirically and theoretically, the simplicity bias has been observed in both shallow architectures, e.g., two-layer networks tending to depend on only a low-dimensional projection of the input~\citep{morwani2023simplicity}, and deep transformers, which typically learn low-order interactions first before higher-order ones~\citep{belrose2024neural,rende2024distributional}.
Some work~\citep{goldblum2023no} argues that deep models inherently favor lower-order polynomial fits when possible~\citep{wilson2025deep}, while other studies explore how parameters such as effective rank further constrain learned representations toward simpler, low-rank subspaces~\citep{huh2021low}.  
Together, these findings underscore how a simplicity bias, though still not fully explained, provides a key inductive advantage in TSFMs, helping them generalize while highlighting the trade-offs between robustness and expressivity.
In this section, we provide some discussion of this simplicity bias.

The first thing to note is that, unlike the other biases discussed in this paper (where there is typically a trade-off --- tuning a design decision ``knob'' improves performance on one end but sacrifices it on the other, in ways that depend on the specific model), the simplicity bias is different.
First, it is more of a ``design goal'' than a ``bias'' (at least in the sense we have used the word so far).
Second, and relatedly, recent work has revealed that larger neural networks often exhibit this simplicity bias, which favors simpler solutions despite their expressive power.
While the precise mechanism behind this phenomenon remains an active area of research, it is an important consideration for TSFMs.
In particular, from an expressiveness perspective, even small models are already highly capable, but larger models seem to introduce an additional simplicity bias that helps prevent overfitting. 
To maintain a simplicity bias via a larger model, however, we also pay significant costs in compute and memory. 
A promising direction, perhaps informed by our insights in this paper, is to retain this bias while reducing overhead through post-training compression techniques.

\subsection{Illustration of the simplicity bias}

\begin{figure}[!htb]
    \centering
    \includegraphics[width=1\linewidth]{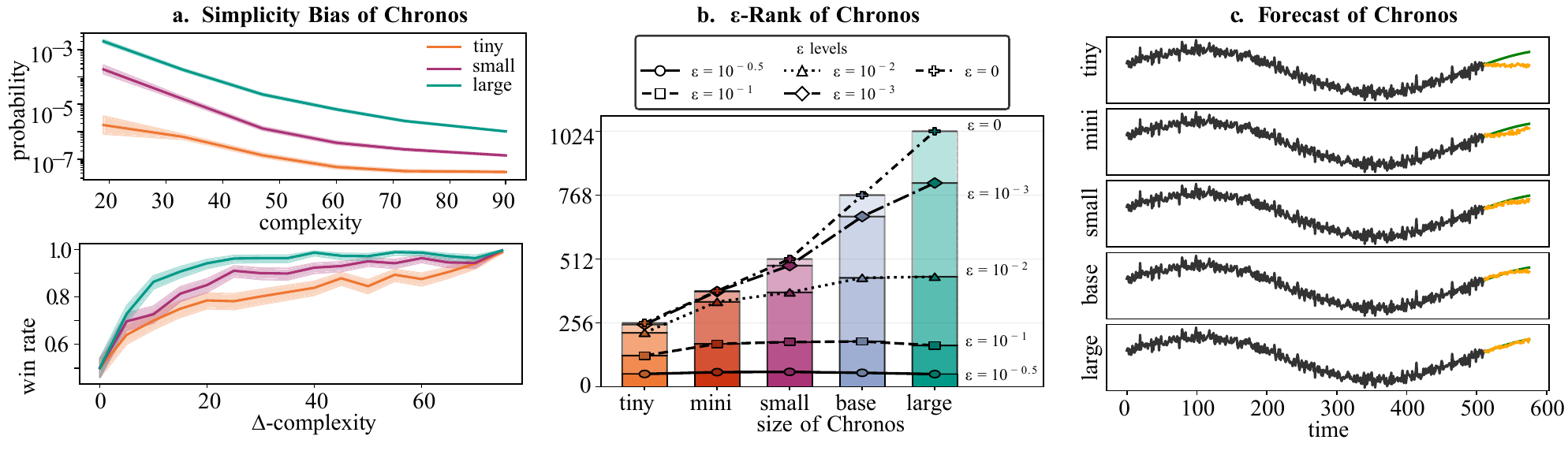}
    \caption{The simplicity bias concerns how a model prefers simple solutions over complicated ones, enhancing the generalization capability (see the right column). In the left column, we present two targeted experiments: first, we construct time series using a different number of basis elements, which we call the complexity, and we compute the model's probability of forecasting the correct future; next, we fix a context and evaluate chances that a model computes a higher probability for a simpler solution than a more complex one, which we call the ``win rate.'' The larger model consistently prefers simpler solutions more than the smaller ones. The middle panel shows the compressibility of the attention matrices in pretrained Chronos models, indicating that larger attention matrices mainly contribute to over-parameterization rather than expressiveness (see also~\citet{yu2025transformer}).}
    \label{fig:simplicity}
\end{figure}

In the experiment shown in~\Cref{fig:simplicity}, we design a context $\mathbf{x}_{1:L}$ with two continuations: a simple one $\mathbf{y}^{(s)}_{L:(L+T)}$; and a complex one $\mathbf{y}^{(c)}_{L:(L+T)}$. We compare the rate that a model prefers the simple solution over the complex one: $\mathbb{P}_{\text{model}}\left(\mathbf{y}^{(s)}_{L:(L+T)} \middle| \mathbf{x}_{1:L}\right) > \mathbb{P}_{\text{model}}\left(\mathbf{y}^{(c)}_{L:(L+T)} \middle| \mathbf{x}_{1:L}\right)$. The finesse of this experiment design is that when the $\Delta$-complexity between the two continuations is zero, then any model prefers them equally; also, when the $\Delta$-complexity gets large, eventually any model simply always prefers the simple solution. Walking from one extreme case to the other, in~\Cref{fig:simplicity}, we see that Chronos (large) consistently prefers simple solutions more than the smaller models.

\begin{tcolorbox}[highlightbox]
\textbf{Summary of the Simplicity Bias:} The \emph{simplicity bias} refers to the tendency of TSFMs to favor simpler solutions, which helps explain their robustness despite their large flexibility.  
This bias is amplified by model size, as scaling up introduces stronger preferences toward simpler hypotheses.
It is also closely tied to compressibility, suggesting that models can retain a simplicity bias while still being pruned or compressed for efficient inference.
\end{tcolorbox}

We now provide details about the two experiments shown in the first column of~\Cref{fig:simplicity}.

\subsection{Prediction Probability as a Function of Data Complexity}

First, we discuss the details of the first panel in the first column of~\Cref{fig:simplicity}.

\textbf{Data generation.}
Each synthetic series is split into a context of length $L$ and a future of length $T$, $\mathbf{y}=(y_1,\dots,y_L,y_{L+1},\dots,y_{L+T})$. The context is produced by a simple parametric mechanism that the model can plausibly capture with $L$ samples. We use two base families: a linear trend $y_t=a t + b$; and a single sinusoid $y_t=A_0\sin(2\pi f_0 t+\phi_0)$. The future is obtained by continuing the same base mechanism and then optionally adding $M\ge 0$ sinusoidal components,
\[
y_{L+t} \;=\; y^{\mathrm{base}}_{L+t} \;+\; \sum_{m=1}^M A_m\sin(2\pi f_m t+\phi_m),\qquad t=1,\ldots,T,
\]
with a short ramp-in at the beginning of the horizon to avoid discontinuities. To control the intrinsic complexity of a sample in a way that is measurable and reproducible, we use a bit-budget proxy $K(\mathbf{y})$ that counts the number of bits required to specify the added components. For each new component, we allocate $k_f$ bits to choose a frequency from a coarse-to-fine grid, $b_A$ bits to quantize amplitude in $[A_{\min},A_{\max}]$, and $b_\phi$ bits to quantize phase in $[0,2\pi)$; hence $K(\mathbf{y})\approx K_{\text{base}}+\sum_{m=1}^M (k_f+b_A+b_\phi)$. Small nonzero complexities are achievable by allowing very coarse choices (e.g., $k_f=0,1,2$) and low-bit quantization for $(A,\phi)$; larger complexities are obtained by increasing these bit budgets and/or $M$. After generation, we standardize the future with a shared scaling computed from the simple continuation (mean $\mu_S$, std $\sigma_S$) so that the magnitude of log-likelihoods is not dominated by trivial scale differences: $\tilde{y}_{L+t}=(y_{L+t}-\mu_S)/\sigma_S$.

\textbf{Result computation.}
For a model $\mathcal{M}$, we score each future under teacher forcing, using the Chronos tokenizer to encode the context as input IDs and the future as labels. The model’s total log-likelihood of the future given the context is
\[
\log p_\mathcal{M}(\mathbf{y}_{L+1:L+T}\mid \mathbf{y}_{1:L}) \;=\; \sum_{t=1}^T \log p_\mathcal{M}\!\big(y_{L+t}\,\big|\, \mathbf{y}_{1:L},\, y_{L+1:L+T-1}\big),
\]
and we normalize by $T$ to obtain an average log-probability per token $s(\mathbf{y})=\frac{1}{T}\log p_\mathcal{M}(\cdot)$. To summarize how likelihood changes with intrinsic complexity, we bin the samples by $K(\mathbf{y})$ into $B$ evenly populated\footnote{In practice, we use a helper that assigns indices by quantiles of $K$; we report the mean of $K$ within each bin as its horizontal location.} bins and compute, for each bin $b$, the sample mean $\bar{s}_b$ and a $95\%$ confidence interval using the normal approximation $\bar{s}_b \pm 1.96\,\hat{\sigma}_b/\sqrt{n_b}$, where $\hat{\sigma}_b$ is the sample standard deviation and $n_b$ the number of sequences in the bin. The baseline plot overlays the bin means for all models as a function of the corresponding bin centers in bits. For downstream plotting, we also save a compact $3\times B$ array per model containing the lower bound, mean, and upper bound in each bin, along with the vector of bin centers. Interpreting the figure is straightforward: curves that lie higher indicate that the model assigns more probability mass on average to sequences at that complexity level; a gentle downward slope with increasing bits is expected, as additional structured variation makes the future harder to predict.

\subsection{Win Rate of the Simple Solution}

Next, we discuss the details of the second panel in the first column of~\Cref{fig:simplicity}.

\textbf{Data generation.}
This Occam-pairs experiment asks whether a model prefers a {simpler} continuation over a more {complex} one, when both are plausible given the same past. We first draw a context $\mathbf{y}_{1:L}$ from the same base families as above (linear or single sinusoid) and form a {simple} future by continuing the base mechanism into the next $T$ steps. We then create a {complex} future by adding controlled extra structure to the simple one:
\[
\mathbf{y}^{(C)}_{t} \;=\; \mathbf{y}^{(S)}_{t} \;+\; \sum_{m=1}^M A_m\sin(2\pi f_m t+\phi_m),\qquad t=1,\ldots,T,
\]
with the same ramp-in convention and the same shared standardization with respect to the simple future. The key scalar we vary across trials is the {complexity gap} $\Delta K$, defined as the additional bit budget needed to describe the extra components. As above, each component pays $k_f+b_A+b_\phi$ bits; thus $\Delta K=\sum_{m=1}^M(k_f+b_A+b_\phi)$ (optionally with a combinatorial term if we explicitly encode a subset of a large frequency grid). To ensure well-behaved behavior at the endpoints, we include two calibrations. At $\Delta K=0$ we set $M=0$ and, if observational noise is added, we add the {same} noise realization to both futures. This makes the two futures exchangeable, so no model can reliably prefer one over the other. At very large $\Delta K$ we gradually turn off preview leakage from the context and increase the amplitude gain of the extra components over a window in $\Delta K$, so that the complex future departs cleanly and strongly from the pattern established in the context.

To prevent the win-rate curves from trivially saturating near $1$ (as they would if the complex future were always obviously inconsistent with the context), we apply a lightweight scheduler that couples nuisance knobs to the target gap $\Delta K$. As $\Delta K$ increases, we gradually reduce observation noise and ramp-in length, and we suppress any preview of the complex components in the context. At small and mid $\Delta K$ these choices keep the two futures similarly plausible from the same past, so wins are earned rather than guaranteed. We also regularize the construction with shared standardization (both futures are scaled by the simple future’s mean and variance), matched energy of the added components across $\Delta K$ (so higher bit budgets translate to finer specification rather than unbounded amplitude), and an explicit tie tolerance $\varepsilon$ so indistinguishable pairs contribute $0.5$ on average. Two anchors complete the calibration: at $\Delta K = 0$ we set $M = 0$ and share the same noise realization across the two futures; while at very large $\Delta K$ a soft right-end ``anchor'' smoothly turns off preview and ramp and slightly boosts gain so that all models approach $1$ without a discontinuous jump. Together, the scheduler and regularization make the mid-range of $\Delta K$ informative, avoiding the confusion that would arise if readers compared to the baseline plot (where single-sequence likelihoods often decline quickly with complexity) and wondered why pairwise win rates would otherwise sit near $1$ for almost all $\Delta K$.

\textbf{Result computation.}
For each triple $(\mathbf{y}_{1:L},\mathbf{y}^{(S)}_{1:T},\mathbf{y}^{(C)}_{1:T})$ and model $\mathcal{M}$ we compute two teacher-forced log-likelihoods,
\[
\ell_S \;=\; \log p_\mathcal{M}\!\big(\mathbf{y}^{(S)}\mid \mathbf{y}_{1:L}\big),\qquad
\ell_C \;=\; \log p_\mathcal{M}\!\big(\mathbf{y}^{(C)}\mid \mathbf{y}_{1:L}\big),
\]
and form the difference $\Delta\ell=\ell_S-\ell_C$. A trial counts as a \emph{win for simple} if $\Delta\ell$ is positive by more than a small tie tolerance $\varepsilon$; if $|\Delta\ell|\le \varepsilon$ we record a half-win so that truly indistinguishable pairs contribute $0.5$ on average. Repeating this for many independently generated pairs at the same $\Delta K$ yields an empirical win rate $W(\Delta K)=\frac{1}{n}\sum_i w_i\in[0,1]$. We summarize uncertainty with a Wilson score interval at $95\%$ confidence, which is appropriate for a proportion and behaves well even when $W(\Delta K)$ is near $0$ or $1$. Plotting $W(\Delta K)$ against the bit gap produces the Occam curve for each model. Two anchors validate the construction: at $\Delta K=0$ all models sit near $0.5$ by symmetry; at sufficiently large $\Delta K$ all models approach $1$ because the complex future contains clear superfluous structure absent from the past. The region between these anchors is informative: models whose curves rise earlier and remain higher over a broad mid-range of $\Delta K$ are better at recognizing and penalizing unnecessary structure, thus exhibiting a stronger empirical simplicity bias in this controlled setting.

\end{document}